\documentclass[10pt,journal,compsoc]{IEEEtran}

\ifCLASSOPTIONcompsoc
  \usepackage[nocompress]{cite}
\else
  \usepackage{cite}
\fi

\ifCLASSINFOpdf
\else
\fi

\usepackage{amsmath}
\usepackage{algorithmic}

\usepackage{array}

\ifCLASSOPTIONcompsoc
 \usepackage[caption=false,font=footnotesize,labelfont=sf,textfont=sf]{subfig}
\else
 \usepackage[caption=false,font=footnotesize]{subfig}
\fi

\usepackage{stfloats}

\usepackage{url}

\usepackage{mathtools}
\usepackage{accents}
\usepackage{amsfonts,amssymb}
\usepackage{algorithm}
\usepackage{array}
\usepackage{textcomp}
\usepackage{verbatim}
\usepackage{graphicx}
\usepackage{cite}

\usepackage{enumitem}
\usepackage{color, colortbl}
\usepackage[table, dvipsnames]{xcolor}
\usepackage{tabu}
\usepackage{longtable}
\usepackage{booktabs}

 \usepackage{multirow}
 \usepackage{float}
 \usepackage{diagbox}
 \usepackage{scalerel}
\usepackage{caption}
\usepackage{makecell}
\usepackage{stmaryrd}
\usepackage{trimclip}
\usepackage{relsize}
\usepackage{tikz}
\usepackage{stackengine}
\usepackage{siunitx}

\newcommand{\tocheck}[1]{\textcolor{black}{#1}}
\newcommand{\pietro}[1]{\textcolor{black}{#1}}
\newcommand{\um}[1]{\textcolor{black}{#1}}
\newcommand{\marco}[1]{\textcolor{black}{#1}}

\newcommand{\oracle}[2]{{#2}}

\newcommand{\classincr}{class incremental }

\newcommand{\ClassIncr}{Class Incremental }
\newcommand{\domincr}{domain incremental }
\newcommand{\lossname}{\marco{kd}}
\newcommand{\name}{\tocheck{LwS}}
\newcommand{\vname}{\tocheck{Learning} }
\newcommand{\tblbold}[1]{\textbf{#1}}

\newcommand{\ie}{\textit{i.e.}}
\newcommand{\eg}{\textit{e.g.}}
\newcommand{\etal}{\textit{et al.}}
\DeclareMathSymbol{\shortminus}{\mathbin}{AMSa}{"39}
\DeclareMathOperator*{\argmax}{argmax}

\newcommand{\smalless}{\raisebox{0.2mm}{\scaleto{<}{3.5pt}}}
\newcommand{\fullhat}[1]{\overset{\lower.5em\hbox{\stretchto{\scaleto{\blacktriangle}{2.7pt}}{1.5pt}}}{#1}}

\newcommand{\asteraccent}[1]{\accentset{*}{#1}}
\newcommand{\unkdP}{\asteraccent{P}}

\newcommand{\unceP}{\mathring{P}}

\makeatletter
\newcommand{\fullhatt}[2][0ex]{%
  \mathrel{
  \overset{
    \vbox to#1{\kern-2\ex@ 
    \hbox{\stretchto{\scaleto{\blacktriangle}{2.7pt}}{1.5pt}}
    \vss}
    }{#2}
    }}
\makeatother

\newcommand{\veryshortarrow}[1][3pt]{\mathrel{%
   \vcenter{\hbox{\rule[-.2pt]{#1}{.4pt}}}%
   \mkern-4mu\hbox{\usefont{U}{lasy}{m}{n}\symbol{41}}}}
 
\renewcommand{\rightarrow}{%
\parbox{.3cm}{\centering\tikz{\draw[->](0,0)--(.2cm,0);}}
}

\renewcommand{\mapsto}{%
\xmapsto{\phantom{\text{\fontsize{2}{2}\selectfont i}}}
}

\newcolumntype{P}[1]{>{\centering\arraybackslash}p{#1}}
\newcolumntype{M}[1]{>{\centering\arraybackslash}m{#1}}

\makeatletter
\DeclareRobustCommand{\shortto}{%
  \mathrel{\mathpalette\short@to\relax}%
}
\newcommand{\short@to}[2]{%
  \mkern2mu
  \clipbox{{.5\width} 0 0 0}{$\m@th#1\vphantom{+}{\shortrightarrow}$}%
  }
\makeatother

\begin{document}

\title{
\vname with Style: \tocheck{Continual} Semantic Segmentation Across Tasks and Domains}

\author{Marco Toldo,~\IEEEmembership{Student Member,~IEEE,}
\thanks{Our work was in part supported by the Italian Ministry for Education
(MIUR) under the \textit{Departments of Excellence} initiative (Law 232/2016).}%
Umberto Michieli,~\IEEEmembership{Graduate Member,~IEEE,} and
Pietro Zanuttigh,~\IEEEmembership{Member,~IEEE}
}

\markboth{Journal of \LaTeX\ Class Files,~Vol.~14, No.~8, August~2021}%
{Shell \MakeLowercase{\textit{et al.}}: A Sample Article Using IEEEtran.cls for IEEE Journals}

\IEEEtitleabstractindextext{%
\begin{abstract}
Deep learning models 
\pietro{dealing with} %
image understanding in real-world settings must be able to adapt to a wide variety of tasks across different domains. 
Domain adaptation and \classincr learning deal with domain and task variability separately, whereas their unified solution is still an open problem.
We tackle both facets of the problem together, taking into account the semantic shift within both input and label spaces. 
We start by formally introducing \um{continual learning} under task and domain shift.
Then, we address the proposed setup by using style transfer techniques to extend knowledge across domains when learning incremental tasks and a robust distillation framework to effectively recollect task knowledge under incremental domain shift. 
\tocheck{%
The devised framework (\name, \vname with Style)}
is able to generalize incrementally acquired task knowledge across all the domains encountered, proving to be robust against catastrophic forgetting.
Extensive experimental evaluation on multiple autonomous driving datasets shows how the proposed method outperforms existing approaches, which prove to be ill-equipped to deal with continual semantic segmentation under both task and domain shift. 
\end{abstract}

}

\maketitle

\section{Introduction}

With the recent rise of deep learning,  the computer vision field has witnessed remarkable advances.
Challenging tasks, such as image semantic segmentation, are nowadays successfully addressed by well-established deep learning architectures \cite{long2015fully,chen2018deeplab,romera2018erfnet}. %
Nonetheless, the fundamental problem of continuously learning and adapting to novel environments remains open and is actively investigated, 
with a long way before its \um{definitive} solution. %

\marco{Although capable of}
remarkable performance in narrow and confined tasks, deep models tend to struggle when confronted with \um{continual} learning of dynamic tasks in ever-changing environments.
A major issue stands in the tendency to \textit{catastrophically forget} previously acquired knowledge \cite{kirkpatrick2017overcoming}, %
with new information erasing that experienced so far.
\pietro{Furthermore,} 
variable input distribution between supervised training data and target data has been shown \um{to} cause  performance degradation, giving rise to the need for \textit{domain adaptation}, which targets knowledge transferability across domains.
Both constitute 
critical problems \marco{when it comes to deploying}
deep models in practical applications, as in the real world it is very likely to 
\marco{face} 
distribution variability both in terms of input data  and of target tasks. 

A thriving research endeavour has been devoted to \marco{continual} learning \marco{(also referred to as incremental \tocheck{learning, IL,} or lifelong learning \cite{delange2022continual})} in vision problems, such as image classification \cite{kirkpatrick2017overcoming,li2017learning,rebuffi2017icarl}, object detection \cite{shmelkov2017incremental,peng2020faster,joseph2021towards} and, more recently, semantic segmentation \cite{michieli2019incremental,cermelli2020modeling,douillard2021plop}.
The majority of those works, however, are limited to a 
\textit{class incremental}
perspective of the continual learning problem, where the focus is strictly posed on the variable task (\eg, class) supervision and label-space shift experienced throughout the learning process.
On the other side, a significant research effort has been directed toward the domain adaptation problem, ranging from a static learning setting \cite{Saito2018MCD,vu2019advent,yang2020fda} to, quite recently, a dynamic perspective \cite{volpi2021continual,volpi2022road,wang2022continualtest}, taking into account incremental changes \pietro{in} \um{the} data distribution.

\begin{figure}[!t]
    \centering
    \includegraphics[width=0.93\linewidth]{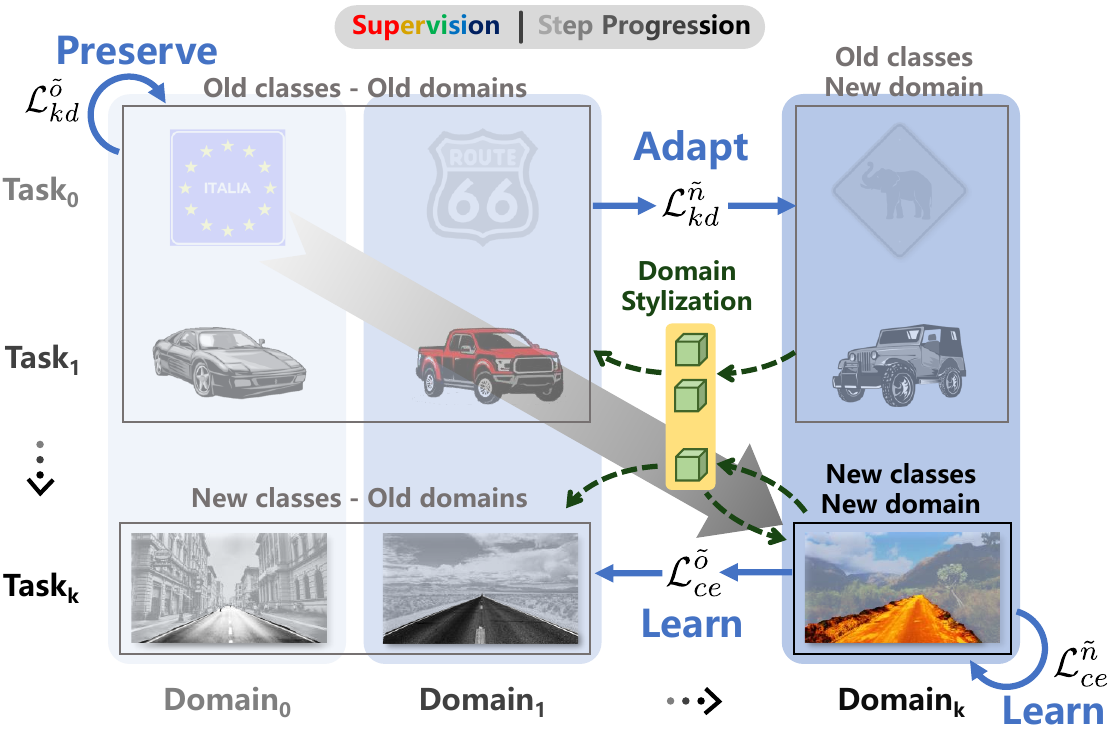}   %
    \caption{\marco{High-level view} of our approach.
    Transparency decrease 
    \tocheck{(top$\protect\rightarrow$down and left$\protect\rightarrow$right)}
    indicates progression through learning steps.
    Colored task icons denote presence of supervision within training data, %
    grayscale \pietro{ones} signal lack of supervision.
    \tocheck{
    At each step, we leverage training data to \textit{learn} new classes on the new domain. 
    Domain stylization allows to reiterate old-domain distribution, crucial to \textit{learn} new tasks and \textit{preserve} old ones on former domains, and to \textit{adapt} old-domain old-task knowledge to new domains.}
    }
    \label{fig:ga}
\end{figure}

Nonetheless, the general continual learning problem across both tasks and domains is yet unexplored for the semantic segmentation task.
Where \classincr methods usually struggle to cope with domain knowledge transferability, \domincr methods lack predisposition to address incremental task supervision. 
We instead propose to tackle \tocheck{continual} semantic segmentation with joint \tocheck{incremental} shift along class and domain directions.
The training process involves multiple steps, each of which carries a new set of classes to learn, along with a training set comprising image samples with a step-distinctive distribution, differing from those experienced in previous steps, and supervision \um{available} only on the newly introduced class set.
The overall objective is for the incremental segmentation model to deliver satisfactory performance across all the tasks (\ie, class sets) and domains encountered so far, with the class- and domain- wise joint training as the target upper bound.

In this novel problem setup \pietro{(see Fig.~\ref{fig:ga})}, both domain adaptation and recollection of past classes must be performed to achieve satisfactory performance.
Under the \domincr angle, it is required to simultaneously learn new classes over past domains and adapt old-class knowledge to the new domain.
From the \classincr perspective, recollection of past knowledge must take into account the variable input distribution characterizing the
\marco{addressed} incremental learning \pietro{problem}.

We therefore devise multiple training objectives to face underlying sub-problems.
While to rehearse knowledge of old classes we resort to the old-step segmentation model, which is a common practice among \classincr learning methods \cite{michieli2019incremental}, to replay information of past-domain input distribution we propose a stylization mechanism.
The average style (\um{\ie, a} very compact representation) of each \um{encountered domain} is computed and stored in a memory bank, to be transferred to novel domains in future steps and reproduce some domain-level information.

The overall optimization framework is made of (i) a standard task loss (\ie, cross-entropy objective) to learn new \pietro{classes} over available training data, (ii) an additional task loss instance to learn new classes in old domains  by leveraging stylization, (iii) a knowledge \pietro{distillation-like} objective to infuse adapted information of past classes in the form of hard pseudo-labels to the new domain and finally (iv) an output-level knowledge distillation objective applied \marco{on} stylized images to retain old-domain old-class performance.

To summarize, our contributions are as follows: 
\begin{itemize}
    \item We \marco{investigate} a novel %
    \marco{comprehensive}
    incremental learning \pietro{setting}  that accounts for variable distribution within both input and label spaces.
    \item We develop a framework to tackle all facets of the class and \domincr learning problem, based on a stylization mechanism to recall domain knowledge under incremental task supervision and a robust distillation framework to %
    \marco{retain} task knowledge under incremental domain shift.
    \item We devise novel experimental setups to simulate the %
    \um{proposed learning} setting and conduct an extensive evaluation campaign.
    \item We show that the proposed method outperforms existing state-of-the-art methods that address the IL problem only from a class or a \domincr perspective.
\end{itemize}

\section{Related Works}

\noindent
\textbf{Semantic Segmentation.} 
Under the impulse of deep learning, semantic segmentation has witnessed a considerable advance in recent years \cite{minaee2021image}.
Since the introduction of fully convolutional networks (FCNs) \cite{long2015fully}, which \um{introduced} 
the popular encoder-decoder architecture, \pietro{huge} research efforts have 
\um{improved} the \um{state of the art}.
Dilated convolutions \cite{yu2016multi,chen2018deeplab} allow to retain sufficiently large receptive fields 
\um{limiting the growth} in model size. 
Spatial \cite{zhao2017pyramid} and feature \cite{li2018pyramid} pyramid pooling \um{extract} and \um{aggregate} contextual information at different scales to acquire enriched representation for \um{improved} dense predictions.  
At the same time, considerable interest was devoted to the design of lightweight architectures for practical applications typically burdened by strict hardware constraints.
MobileNet architectures \cite{howard2017mobilenets,sandler2018mobilenetv2} are built upon the efficient depthwise separable convolution.
ErfNet \cite{romera2018erfnet} resorts to factorized residual layers to provide real-time \um{accurate} segmentation.
\tocheck{Recently, transformers have been applied in vision, even for dense prediction tasks such as semantic segmentation \cite{khan2021transformers}.} 

\noindent
\textbf{Class Incremental Learning} (CIL). 
Continual learning in the form of incremental classification tasks has been subject of growing research interest in the recent past \cite{delange2022continual}.
Extensive literature can be found targeting image classification \cite{kirkpatrick2017overcoming,li2017learning,rebuffi2017icarl,hou2019learning,douillard2020podnet,yan2021der,zhu2021prototype,toldo2022bring,zhu2022self,wu2022class,xie2022general,tang2022learning,douillard2022dytox} and object detection tasks \cite{shmelkov2017incremental,peng2020faster,joseph2021towards,yang2022continualobject} under the incremental learning paradigm.
\marco{Many of} these works \cite{rebuffi2017icarl,douillard2020podnet,yan2021der,wu2022class,xie2022general,tang2022learning,douillard2022dytox} rely on exemplars, \ie, a small portion of training data is stored to be replayed in future steps.
We instead place ourselves in a totally exemplar-free setup.
Among the exemplar-free methods \cite{kirkpatrick2017overcoming,li2017learning,shmelkov2017incremental,peng2020faster,joseph2021towards,yang2022continualobject,zhu2021prototype,toldo2022bring,zhu2022self} we can identify regularization-based \cite{kirkpatrick2017overcoming,joseph2021towards,yang2022continualobject}, rehearsal-based \cite{li2017learning,shmelkov2017incremental,peng2020faster,zhu2021prototype,toldo2022bring} and structure-based \cite{zhu2022self}.
Even if many works propose techniques which could in principle be generalized to various vision tasks (such as the prosperous knowledge distillation mechanism \cite{hinton2015distilling,li2017learning,shmelkov2017incremental}),
when facing the semantic segmentation \um{task, additional complexity, which is not present in case of whole-image classification or object detection, arises \cite{michieli2022domain}.}

More limited literature can be found for incremental semantic segmentation \cite{michieli2019incremental,klingner2020class,cermelli2020modeling,douillard2021plop,maracani2021recall,michieli2021continual}, even though this field has experienced a very recent rise in research consideration \cite{yang2022uncertainty,yang2022continual,cermelli2022incremental,zhang2022representation,phan2022class}.
A first direction of study has been oriented toward the adaptation of the knowledge distillation mechanism to incremental semantic segmentation \cite{michieli2019incremental,klingner2020class,cermelli2020modeling,douillard2021plop,phan2022class,yang2022continual,yang2022uncertainty}.
Michieli \etal \cite{michieli2019incremental,michieli2021knowledge} \pietro{have been} the first to introduce this technique in CIL for dense classification, proposing both feature- and output- level variants of the distillation objective.
In \cite{cermelli2020modeling} authors address the semantic shift of background regions by proposing a novel distillation formula. Furthermore, \cite{douillard2021plop} improves feature-level distillation by pooling representations to capture spatial relationships. 
Phan \etal \cite{phan2022class} introduce a measure of task similarity as a weighting factor in the distillation objective.
Yang \etal \cite{yang2022continual} resort to a structured self-attention approach for preserve relevant knowledge.
Finally, \cite{yang2022uncertainty} extends the popular contrastive learning paradigm to incremental semantic segmentation to improve class discriminability in the feature space.
Nonetheless, none of the aforementioned works address the distribution shift that could be present across tasks within the input space.
We propose to use a distillation objective which is robust to \domincr gaps, and targets the preservation of old-task knowledge both on the current domain, by distilling through robust hard pseudo-labels, and on the past domains, by leveraging domain stylization to distill knowledge when experiencing old-domain input statistics.
Targeting semantic discriminability of latent representations,  a clustering-based objective built upon class prototypes \pietro{is proposed in \cite{michieli2021continual}}.
Maracani \etal \cite{maracani2021recall} introduce a novel rehearsal approach based on the retrieval of training samples by external sources, \ie, via GAN-based generation or web-crawling.
Cermelli \etal \cite{cermelli2022incremental} further show that it is possible to perform continual training with only image-level annotations in incremental steps and reach \um{high accuracy} in some CIL experimental setups.
Nonetheless, this approach could be susceptible to the amount of dense supervision provided in the first learning step, and might not scale well \pietro{to segmentation of images containing objects of different classes.} 
Zhang \etal \cite{zhang2022representation} devise a dynamic incremental framework to decouple the representation learning of old and new tasks.
All the aforementioned works assume statistical homogeneity across learning steps in terms of input data distribution. On the other hand, we address the more realistic setup with both input and label spaces undergoing incremental shifts, and we show the superiority in this generalized setup of the proposed incremental approach \um{compared to} pure CIL competitors.

\noindent
\textbf{Domain Adaptation} (DA).
Deep models are known to suffer performance degradation when presented with varying input distribution between training and testing phases \cite{ben2006analysis}.
Domain adaptation has been extensively investigated to alleviate the aforementioned problem, by safely transferring learned knowledge from label-abundant source domains to label-scarce, \pietro{or even unsupervised, 
target ones.}
Particularly flourishing has been  unsupervised domain adaptation (UDA) for the semantic segmentation task \cite{toldo2020unsupervised,Saito2018MCD,vu2019advent,yang2020fda,zhang2021prototypical,hoyer2022daformer}, as supervision in terms of dense segmentation maps is usually very costly and time expensive to be collected for real-world data.
In its standard form, UDA entails no continual learning, being the task at hand %
the same on both source and target static domains, which are concurrently available.
We instead address a more realistic setup with dynamic task and domain evolution.

More recently, different variations of the static DA have been proposed, relaxing some of the original strict assumptions.
One research direction involves distinct tasks between source and target domains, \ie, \marco{allows} source and target classes to be different. 
Depending on the relationship between source and target class sets, partial \cite{tian2021partial}, open-set \cite{jing2021towards} and  universal \cite{saito2021ovanet,ma2021active} domain adaptation setups have been proposed, 
even though  most  research has been confined to the image classification problem \cite{jing2021towards,saito2021ovanet,ma2021active}.
Moreover, these works do not involve \classincr learning, as adaptation is performed with simultaneous access to source and target domains in a single learning phase. 

Another line of works has explored diverse setups in terms of domain availability.
Some propose to handle multiple source \cite{he2021multi,gong2021mdalu} or target \cite{liu2020open,volpi2021continual,isobe2021multi,zhao2022sourcefree,volpi2022road,wang2022continualtest,marsden2023continual} domains.
This can involve a single adaptation phase \cite{he2021multi,gong2021mdalu}, or multiple phases where different domains are experienced in different learning steps in a incremental fashion \cite{volpi2021continual,zhao2022sourcefree,volpi2022road,wang2022continualtest,marsden2023continual}, in fact, undertaking continual learning under the domain adaptation perspective.
Yet, all these works assume homogeneity of tasks across all the domains encountered, whereas the class and \domincr setup we propose deals with variable learning conditions both along task and domain \marco{progressions.} %
Garg \etal \cite{garg2022multi} develop a multi-\domincr learning (MDIL) framework that involves classification tasks shifting across multiple domains experienced in an incremental fashion. 
However, \textit{total} supervision is available on all the domains encountered, leading to overlapping incremental class sets. We instead adhere to a stricter CIL setup, with disjoint groups of semantic categories incrementally introduced.

It is possible to find a few works that address both task incremental and domain adaptation problems.
Kalb \etal \cite{kalb2021continual} discuss class and \domincr learning, but each task is tackled individually by evaluating standard CIL and DA methods. %
In \cite{shenaj2022continual} 
coarse-to-fine continual learning is explored, but the proposed  setup does not involve domain shift across learning steps, as source and target domains are kept fixed.
Recently, Simon \etal \cite{simon2022generalizing} address continual learning with tasks and domains dynamically evolving.
Still, they assume to have task supervision on all the considered domains at each \marco{task incremental} step, which may not be a realistic assumption in real-world applications.
In addition, rehearsal of training exemplars is performed, and the  method specifically targets image classification.

\section{Problem Setup}

In semantic segmentation we aim at labeling every individual spatial location of an image by associating it with a semantic class taken from a predefined collection of candidates $\mathcal{C}$.
That is, given an \um{RGB}  image $\mathbf{X} \in \mathcal{X} \subset \mathbb{R}^{H \times W \times 3}$, a segmentation network $S: \mathcal{X} \mapsto \mathcal{Y}$ is exploited to provide its segmentation map $\hat{\mathbf{Y}} \in \mathcal{Y} \subset \mathcal{C}^{H \times W}$. $\hat{\mathbf{Y}}$ should be an accurate prediction of the ground truth map $\mathbf{Y}$, which is available only at training time.

\begin{figure}[!t]
    \centering
    \includegraphics[width=1.\linewidth]{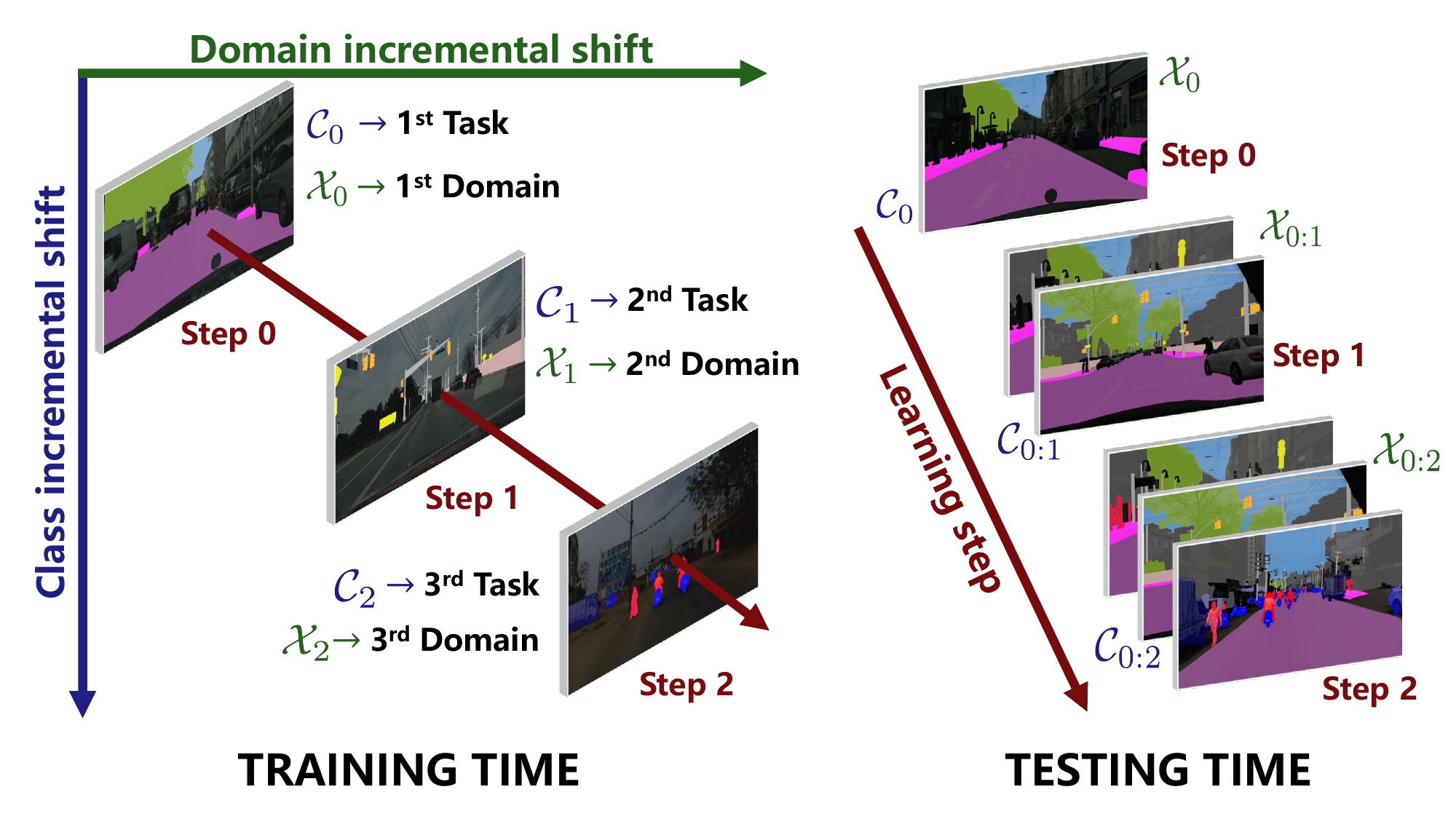}
    \caption{
    \marco{Overview of the class and domain incremental setup.}
    \pietro{At each step,  
    training data come from a new domain and is labeled on a new class set.} When testing, performance is measured on all domains and classes experienced so far.
    }
    \label{fig:setup}
    \vspace{-0mm}
\end{figure}

We follow an incremental learning protocol to optimize the segmentation network, \marco{as depicted in Fig.~\ref{fig:setup}}.
Specifically, the predictor is trained in multiple steps \pietro{$t \!=\! 0,...,T \!-\!1$} to recognize a progressively increasing set of semantic classes.
At step $t$, a new class set $\mathcal{C}_{t}$ is introduced, along with training data \marco{$\mathcal{D}_{t} \!=\! \{ (\mathbf{X}_{t}, \mathbf{Y}_{t}) \} \!\subset\! \mathcal{X}_{t} \times \marco{\mathcal{Y}}_{t}$} associated to that set, which is available on the current image domain $\mathcal{X}_{t}$. 
The supervision provided by $\mathcal{D}_{t}$ is restricted to $\mathcal{C}_{t}$, meaning that any pixel within $\mathcal{D}_{t}$ is tagged \pietro{in $\mathcal{Y}_{t}$}  with $c \in \mathcal{C}_{t}$.
At the end of the step, all the currently accessible data is discarded %
\um{and  is not} reused again. 
The  procedure is reiterated for multiple learning steps, \marco{with} a new domain $\mathcal{X}_{t}$ and class set $\mathcal{C}_{t}$ \marco{being} introduced and \um{used for training} 
\pietro{at each step}.
\\
More formally, the objective is to train $S_{t}: \mathcal{X}_{0:t} \mapsto \mathcal{Y}_{0:t}$
\begin{itemize}
    \item to recognize all the semantic classes observed up to the \pietro{current} step $t$\um{:}
    \begin{equation}
        \mathcal{Y}_{0:t} \in \mathcal{C}_{0:t}^{H \times W}, \quad \mathcal{C}_{0:t} \!=\! \bigcup_{k=0}^{t}\mathcal{C}_{k}\um{,}
    \end{equation}
    \item on all the image domains 
    \pietro{experienced} so far\um{:}
    \begin{equation}
      \mathcal{X}_{0:t} = \bigcup_{k=0}^{t} \mathcal{X}_{k}.
    \end{equation}
\end{itemize}

\marco{We remark that $\{ \mathcal{X}_{t} \}_{t=0}^{T}$ are characterized by diverse statistical properties, \ie, domain shift occurs between them, typically manifested through cross-domain variable visual appearance of scene elements that yet share semantic significance.} 
All $\mathcal{C}_{t}$ are disjoint sets, except for the \textit{unknown} ($u$) class, which belongs to each of them. Class $u$ at step $t$ contains all the past and future classes. In other words, $u$ undergoes a semantic shift across subsequent steps and, for this reason, demands special care when being handled {\cite{cermelli2020modeling}}.

\section{Overview of the Proposed Method}

We \marco{concurrently} face challenges peculiar to both \pietro{the} domain adaptation and \pietro{the \classincr learning} settings.
\\
\textbf{Domain Adaptation.} The segmentation network is trained on data from multiple domains, each \marco{holding} only a subset of the whole set of the semantic classes. 
\marco{Even so}, the model is expected to provide satisfactory prediction performance on all the observed domains and semantic classes. \marco{Hence}, it is necessary to transfer knowledge across domains to 
\begin{enumerate}
    \item[(i)] learn \textit{new-class} clues shared across the current (supervised) domain and the past ones (where new-class supervision was not available \marco{during} past steps); 
    \item[(ii)] adapt \textit{old-class} knowledge learned in \marco{former} %
    domains to the novel \um{domain}.
\end{enumerate}
\textbf{\ClassIncr Learning.} The different class supervision available on different domains leads us to a \classincr problem, where semantic categories come across in a continual fashion. Therefore, we are required to address the widely known \um{{catastrophic forgetting}} phenomenon \cite{kirkpatrick2017overcoming}, aiming at preserving knowledge from past classes when learning new ones. 
However, unlike standard CIL, knowledge preservation has to be performed differently depending on the domain in which it is applied:
\begin{itemize}
    \item[(i)] in \textit{past domains} straightforward \pietro{recollection} of previously observed classes can be imposed, 
    \pietro{as those classes were learned over past domain distributions;}
    \item[(ii)] whereas, in the \textit{novel domain} recalled memory of past classes should be adapted to account for the semantic shift happening within the input space.
\end{itemize}

\begin{table}[!t]
    \centering
    \renewcommand{\arraystretch}{1.3}
    \caption{
    Training objectives: 
    \marco{the \textit{n}/\textit{o} \pietro{superscripts} denote the use of new/old domain data, with $\tilde{\cdot}$ implying stylization.}
    }
    \begin{tabular}{c|cc}
    \toprule
    \multicolumn{1}{c|}{} & \multicolumn{1}{c}{New Domain} & Old Domains \\
    \midrule
    New Classes & $\mathcal{L}_{ce}^{\tilde{n}}$ & $\mathcal{L}_{ce}^{\tilde{o}}$ \\
    Old Classes & $\mathcal{L}_{\lossname}^{\tilde{n}}$ & $\mathcal{L}_{kd}^{\tilde{o}}$ \\
    \bottomrule
    \end{tabular}
    \label{tab:train_obj}
    \vspace{-0mm}
\end{table}

\begin{figure*}
    \centering
    \includegraphics[width=1.\linewidth]{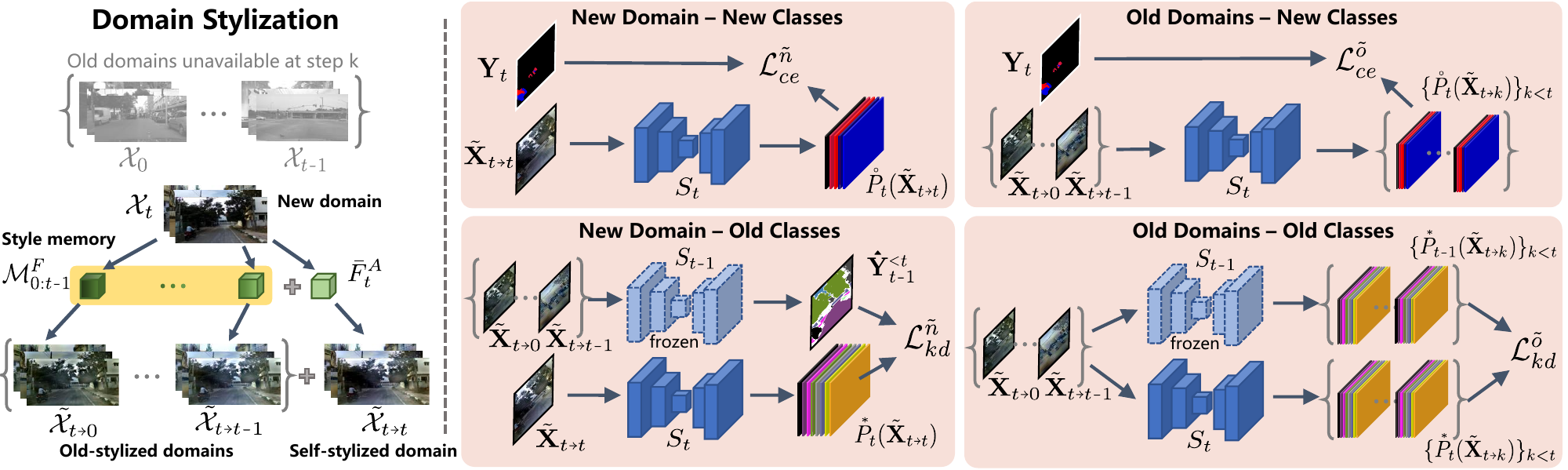}
    \caption{
    \tocheck{
    Model architecture:
    we decompose class and domain \um{IL} into simpler sub-problems, each addressed by a suitable objective (\um{4 panels in the} right side);
    to access no longer available old domain data, we resort to stylization (left side).
    }
    }
    \label{fig:arch}
\end{figure*}

We break down the domain \um{shift} and class continual learning \um{problems} into simpler underlying sub-problems, as indicated above. %
Our overall learning framework builds upon multiple individual objectives, each focusing on a specific challenge enclosed in the general setup. 
We simultaneously progress along class and \domincr directions; at each learning step\um{,} after the first one\um{,} both class\marco{es} and domains experienced so far can be arranged into \um{\textit{new}} or \um{\textit{old}} types, according to whether they are currently available or not. 
More in detail, we  propose a specific learning objective for each of the different combinations of domain and class types (see Table~\ref{tab:train_obj} and \marco{Fig.~\ref{fig:arch}}), \ie, \marco{to}\um{:}
\begin{enumerate}[topsep=0pt]
    \item[(i)] learn new classes on new domains (Sec.~\ref{sec:newD_newC})\um{;}
    \item[(ii)] \marco{learn new classes on old domains} (Sec.~\ref{sec:oldD_newC})\um{;}
    \item[(iii)] adapt old-class information to new domains  (Sec.~\ref{sec:newD_oldC})\um{;}
    \item[(iv)] preserve old-class information in old domains  (Sec.~\ref{sec:oldD_oldC})\um{.}
\end{enumerate}

\subsection*{Domain Stylization}
\label{sec:domain_style}

We resort to a style transfer mechanism to recreate image data with statistical properties resembling those of past domains. 
More specifically, starting from the available image data originating from the input domain accessible at the current step, we transfer the styles extracted from all the previously encountered domains.
By doing so, a stylized version of each of the former domains is produced, with image content derived from the novel dataset.

The benefits that \marco{originate} from domain stylization are manifold:
(i) We force the prediction model to experience past input distributions under supervision or 
\tocheck{pseudo-supervision}, 
tackling domain-level catastrophic forgetting.
(ii) We aim at learning new classes on old domains, where supervision was not available when they were directly observed.
At the same time, we propose to preserve old-class knowledge on old domains, counteracting class-level catastrophic forgetting.
(iii) By encountering \marco{a} variegate input distribution, the predictor is encouraged to develop the ability to generalize to unseen domains, which is crucial in a continual learning paradigm \marco{that involves} \pietro{domain shift}. 

\marco{The style transfer mechanism we adopt is inspired by \cite{yang2020fda} and involves low computational cost and memory requirements.}
\pietro{ The original algorithm works in the Fourier transform domain:} the low frequency portion of the amplitude of the spectral representation from a target image (\um{\ie, the} style) is extracted and applied to replace that of a source image (\um{\ie, the} content), whose phase \marco{component} %
is kept unchanged.
The outcome %
\marco{is} image data with source semantic information, and target-like low-level appearance.

\marco{We enhance the original method to accommodate for the further complexity brought \um{in} by the class and domain incremental setting.}
From each image of the currently available dataset, we extract its style tensor (\ie, \um{the} amplitude central window), and we average \um{it} over all the samples\um{:}
\begin{equation}
    \bar{F}^{A}_{t} = 
    \frac{1}{|\mathcal{D}_{t}|}
    \sum_{\mathbf{X} \in \mathcal{D}_{t}} 
    \mathcal{F}^{A} (\mathbf{X})[{W_{\beta}}],
\end{equation}
where $\mathcal{F}^{A}(\mathbf{X})$ is the amplitude obtained by the FFT applied to image $\mathbf{X}$, and $W_{\beta}$ is the style window.
By doing so, we are extracting significant knowledge of \tocheck{\textit{domain-dependent}} statistical properties, condensed in a compact representation.
The domain-specific style $\bar{F}^{A}_{t}$ of step $t$ is stored in an incrementally\um{-}filled memory bank 
$\mathcal{M}^{F}_{0:t\shortminus 1} \!=\! \{ \bar{F}^{A}_{k} \, | \, k \!<\! t \}$
and preserved across steps.
\marco{By leveraging the proposed storage mechanism, at each incremental step we can access crucial information of past domain low-level properties (yet minimal if compared to that contained in whole training sets), without requiring direct access to raw image data, which would violate the exemplar-free assumption.
We stress that domain shift affects low-level details, while high-level semantic content is mostly shared across domains (\eg, the road serves the same purpose regardless of the dataset, while its appearance in terms of texture or pavement material might vary considerably).
}
To create an oldly-stylized dataset at step $t$ looking back at step $k \!<\! t$ (\ie, $\tilde{\mathcal{X}}_{k}^{t}$), 
\marco{for each image of the current domain we replace its amplitude window with that of the selected former domain as follows:} 
\begin{equation}
    \tilde{\mathcal{X}}_{t\shortto k} = 
    \{ \mathcal{F}^{\shortminus1}([\bar{F}^{A}_{k} + \mathcal{F}^{A} (\mathbf{X})[{W^{c}_{\beta}}], \mathcal{F}^{P}(\mathbf{X})]) \, | \, \mathbf{X} \in {\mathcal{X}}_{t} \},
\end{equation}
\pietro{where $\mathcal{F}^{\shortminus1}$ is the inverse FFT operator and $\mathcal{F}^{P}\marco{(\mathbf{X})}$ is the Fourier phase component of $\mathbf{X}$.}
\marco{In addition, we devise a self-stylization mechanism by self-applying domain style to improve generalization toward future steps, promoting forward transfer.}
As for the dimension of the style window, we experimentally found that the $\beta$ parameter as defined in \cite{yang2020fda} (\ie, the parameter controlling the window size) provides satisfactory \um{and robust} results when set to $1\mathrm{e}\!\shortminus\!2$.

\um{Finally, \marco{we stress}} that our approach is independent of the style transfer technique used, 
\marco{provided that style information and content can be extracted in two distinct} \um{steps}.

\section{Learning Across Tasks and Domains}
\label{sec:method}

\subsection{Learning New Classes over New Domains}
\label{sec:newD_newC}

In the proposed class and domain continual learning framework, direct supervision comes uniquely for the newly introduced class set $\mathcal{C}_{t}$ and image domain $\mathcal{X}_{t}$ in the form of the training dataset $\mathcal{D}_{t} \subset \mathcal{X}_{t} \times \marco{\mathcal{Y}_{t}}$.
As mentioned before, image pixels not belonging to $\mathcal{C}_{t}$, \ie, of past or never seen classes, are assigned to a special class \textit{unknown}, whose semantic \marco{statistical properties} \um{are} highly dynamic.

To account for the semantic shift suffered by the \textit{unknown} class at the current step $t>0$ w.r.t.\ previous steps, we group the past and unknown class probability channels as follows:
\begin{equation}
    \unceP_{t}(\mathbf{X})[x,y,c] = 
    \begin{cases}
        P_{t}(\mathbf{X})[x,y,c],& \text{if } c\neq u \\
        \sum_{c' \in \mathcal{C}_{0:t\shortminus1}} P_{t}(\mathbf{X})[x,y,c'],& \text{if } c = u \\
    \end{cases}
\end{equation}
where $P_{t}(\mathbf{X}) \in \mathbb{R}^{H \times W \times |\mathcal{C}_{0:t}|}$ is the output of $S_{t}$ prior to the $\argmax$ when a generic image $\mathbf{X} \in \mathcal{X}$ is given as input.
\\
We additionally define 
$\tilde{\mathcal{D}}_{t\shortto t} \subset \tilde{\mathcal{X}}_{t\shortto t} \times \mathcal{Y}_{t}$ 
as the \textit{self-stylized} training dataset at step $t$, where the average style (defined above in \marco{Sec.~\ref{sec:domain_style}}) 
of the current image domain has been applied on top of the $\mathcal{X}_{t}$ domain itself. 

To learn the newly introduced classes over the new domain we optimize\um{:}
\begin{equation}
    \mathcal{L}_{{ce}}^{\tilde{n}}(\mathcal{C}_{t},\mathcal{X}_{t}) = 
    - \frac{1}{
    |\tilde{\mathcal{D}}_{t\shortto t}|
    } 
    \sum_{\tilde{\mathbf{X}}, \mathbf{Y} \in 
    \tilde{\mathcal{D}}_{t\shortto t}
    }
    \!\!\!\!  \mathbf{Y} \cdot \log \unceP_{t}(\tilde{\mathbf{X}}) \um{,}
\label{eq:nc_nd}
\end{equation}
where we leverage input data \marco{with} current style and supervision over the new class set.
The $\tilde{n}$ superscript indicates the use of self-stylized data \marco{on the \textit{new} domain}.
The purpose of self-stylization is twofold;
first, it provides additional robustness and generalization capability to the prediction model, since input data is supplied with more homogeneous low-level statistic across individual samples. 
Second, it forces the prediction model to experience domain statistics that will be stored and replayed in the future, acting as proxies for the no longer available previous domain statistics.

\subsection{Learning New Classes over Past Domains}
\label{sec:oldD_newC}

To \marco{compensate} for the lack of \marco{available input data} for past domains, we generate proxy datasets retaining low-level statistics resembling those of past domains.
More precisely, for each style $\bar{F}^{A}_{k} \in \mathcal{M}_{0:t}$ of step $k<t$ we \pietro{build}
$\tilde{\mathcal{D}}_{t\shortto k} \subset \tilde{\mathcal{X}}_{t\shortto k} \times \mathcal{Y}_{t}$
(as detailed in Sec.\ \ref{sec:domain_style}), \ie, an \textit{oldly-stylized} training dataset at step $t$, for which 
\marco{domain-specific visual attributes}
of step $k<t$ has been applied on %
domain $\mathcal{X}_{t}$. 

Supervision on the newly introduced classes over the old domains is exploited by optimizing\marco{:}
\begin{equation}
    \mathcal{L}_{ce}^{\tilde{o}}(\mathcal{C}_{t},\mathcal{X}_{0:t\shortminus1}) =  
    - \frac{1}{t}
    \sum_{k=0}^{t\shortminus1}
    \frac{1}{
    |\tilde{\mathcal{D}}_{t\shortto k}|
    } 
    \sum_{\tilde{\mathbf{X}}, \mathbf{Y} \in 
    \tilde{\mathcal{D}}_{t\shortto k}
    } \!\!\!\! \mathbf{Y} \cdot \log \unceP_{t}(\tilde{\mathbf{X}}),
\label{eq:nc_od}
\end{equation}
\um{where we leverage input data with past styles (\ie, with distributions supposedly close\footnote{\tocheck{The \textit{closeness} depends on what the style transfer mechanism is able to transfer in terms of statistical properties. The distribution gap is reduced in terms of low-level properties, while semantic high-level distribution should already be similar across domains.}} to those of no longer available former domains) and the supervision over the new class set.}
The superscript $\tilde{o}$ indicates the use of \textit{oldly}-stylized data.

By concurrently learning the segmentation task at the present step over an augmented pool of input data distributions \marco{from the past}, the prediction model should learn more general and shareable clues, overcoming \marco{the} domain shift %
\marco{inherent in}
the domain continual learning paradigm.

\subsection{Adapting Old Classes to New Domains}
\label{sec:newD_oldC}

In the addressed \classincr learning scenario, at each new learning step all past class sets are assumed to lack any direct supervision. 
To recall previously acquired knowledge, we resort to the 
\pietro{well-known}
knowledge distillation objective \cite{hinton2015distilling}.
\marco{Yet}, differently from the standard \classincr learning problem as traditionally \pietro{formalized} in \um{the} literature \cite{rebuffi2017icarl}, we expect to encounter additional challenges: 
\\
(i) the input data of past domains %
(\ie, experienced by the segmentation model when previous class sets were learned) 
are no longer available;  \\
(ii) a distribution shift separates the current image data to that available at \marco{former} steps.
\marco{Thus}, we no longer have access to %
\tocheck{data distributed as that experienced}
by the segmentation model saved from \marco{the} past step, 
\tocheck{which, in principle, should be}
leveraged to distill knowledge of old \marco{classes}. 

To replicate the image distribution of %
data of \marco{past} steps, we resort to the stylization {mechanism} (Sec.~\ref{sec:domain_style}).
Specifically, for each \marco{old} domain $\mathcal{X}_{k}$, $k\!<\!t$, we build an oldly-stylized dataset 
$\mathcal{D}_{t\shortto k}$
starting from that of the current step $t$. 

To access a form of supervision over the past classes we make use of pseudo-labeling via the prediction model from the previous step, which should retain profitable knowledge on the semantic categories learned so far.
However, said model might not distill knowledge effectively when fed with input data of an unseen distribution, \ie, originating from the newly introduced domain.
Therefore, we exploit  oldly-stylized data to enhance pseudo-labeling by mitigating domain shift.
\pietro{\um{We} denote with $P_{t\shortminus1}^{k}(\tilde{\mathbf{X}}) \subset \mathbb{R}^{H \times W \times |\mathcal{C}_{t\shortminus1}|}$,
$\tilde{\mathbf{X}} \in \mathcal{X}_{t\shortto k}$, 
 the classification} probability map from model $S_{t\shortminus1}$ over new domain images with \marco{the style of step $k$.} %
We then compute pseudo-labels following\marco{:}
\begin{equation}
    \hat{\mathbf{Y}}_{t\shortminus1}^{
    \mathcal{K}
    }[x,y] = \argmax_{c \in \mathcal{C}_{0:t\shortminus1}} \max_{
    \marco{k \in \mathcal{K}}
    } P_{t\shortminus1}^{k}(\tilde{\mathbf{X}})[x,y],
    \label{eq:pseudo}
\end{equation}
\pietro{where we leverage old model predictions over past styles, \ie, we set $\mathcal{K} \!=\! \{0,...,t \!-\! 1\}$, %
 while}
$\max_{
\marco{k \in \mathcal{K}}
} P_{t\shortminus1}^{k}(\tilde{\mathbf{X}})[x,y]$ indicates that for each spatial location $(x,y)$ we take the probability vector \marco{associated to the style} with maximum peak value.
We then refine the generated pseudo-labels at each spatial location (we  \pietro{will shorten 
$\hat{\mathbf{Y}}_{t\shortminus1}^{\mathcal{K}=\{0,...,t \shortminus 1\}}$ as
$\hat{\mathbf{Y}}_{t\shortminus1}^{\smalless t}$ and drop \marco{the term} $[x,y]$ for ease of notation) }as\marco{:}
\begin{equation}
    \fullhatt[-0.3ex]{\mathbf{Y}}^{
    \lower.75em\hbox{\smalless \scriptsize $t$}
    }_{t\shortminus1} = 
    \begin{cases}
        \hat{\mathbf{Y}}_{t\shortminus1}^{\smalless t},& \text{if } \hat{\mathbf{Y}}_{t\shortminus1}^{\smalless t} \marco{\text{ confident }} \wedge \mathbf{Y}_{t} = u \\
        {u,}& {
        \text{if } {\mathbf{Y}}_{t} \neq u
        }\\
        \mathrm{ignore},& \text{elsewhere} \\
    \end{cases}
\end{equation}
where $\mathbf{Y}_{t} \in \mathcal{Y}_{t}$. 
The hard pseudo-label $\hat{\mathbf{Y}}_{t\shortminus1}^{\smalless t}[x,y]$ (\ie, after the $\argmax$ operation in Eq.~\eqref{eq:pseudo}) is considered to provide a confident prediction if the peak probability value (of the probability map prior to the $\argmax$) is bigger than a threshold $\tau$, or if that value is among the top-$K$ fraction of highest peaks for class $c \!=\! \hat{\mathbf{Y}}_{t\shortminus1}[x,y]$. We set $\tau \!=\! 0.9$ and $K \!=\! 0.66$ \marco{as advised in} \cite{yang2020fda}.
In addition, we leverage the ground-truth supervision on new classes to correct noisy estimations in pseudo-labels, by marking as \textit{unknown} \marco{(\ie, \textit{u})} all the pixels of newly introduced categories. We remark that the employed knowledge distillation is designed to provide insight on previous tasks (where current new classes were assigned to the \marco{\textit{u}} class), whereas we entrust Eq.~\eqref{eq:nc_nd} to instill understanding of the novel task. 
We experimentally \marco{verify} that \marco{using} separate objectives to train on new and old classes leads to improved results, 
as it forces the model to learn to better discriminate 
between different incremental class sets, part of which might coexist under the same \textit{unknown} group for one or more learning steps. This is especially true for autonomous driving datasets, where each image can contain several semantically diverse elements, for all of which we may not have supervision from the start of the training.

To infuse adapted information about past classes at the current step without direct access to ground-truth information, we resort to the following objective:
\begin{equation}
    \mathcal{L}^{\tilde{n}}_{\lossname}(\mathcal{C}_{0:t\shortminus1},\mathcal{X}_{t}) = 
    - \frac{1}{|\mathcal{D}_{t}|} \sum_{{\tilde{\mathbf{X}} \in \mathcal{D}_{t}}} 
    \fullhatt[-0.3ex]{\mathbf{Y}}^{\lower.75em\hbox{\smalless \scriptsize $t$}}_{t\shortminus1} 
    \cdot \log {\unkdP_{t}}({\tilde{\mathbf{X}}}),
\end{equation}
by which we distill knowledge of past tasks (\ie, recognition of classes in $\mathcal{C}_{0:t\shortminus1}$) over the new domain $\mathcal{X}_{t}$ via the pseudo-labels derived from the old model $S_{t}$.

To account for the semantic shift suffered by the \textit{unknown} class of step $t - 1$ when moving to a \marco{new} step $t \!>\! 0$, we group \textit{new} and \textit{unknown} class probability channels as follows\marco{:}
\begin{equation}
    \unkdP_{t}(\mathbf{X})[x,y,c] = 
    \begin{cases}
        P_{t}(\mathbf{X})[x,y,c],& \text{if } c\neq u \\
        \sum_{c' \in \mathcal{C}_{t}} P_{t}(\mathbf{X})[x,y,c'],& \text{if } c = u \\
    \end{cases}
\label{eq:unb_ce}
\end{equation}
where $\unkdP_{t}(\mathbf{X}) \!\in\! \mathbb{R}^{H \times W \times |\mathcal{C}_{0:t\shortminus1}|}$.
We opt for the use of hard-labels in place of the more common soft-labels in the distillation-like loss in order to prevent enforcing an uncertain behavior to $S_{t}$. 
This behaviour could be originated by the mismatch between training and inference input distribution undergone by the old model $S_{t-1}$, which has been trained over past domains and now is fed with new domain data (the \marco{oldly-stylizing} operation reduces domain shift but 
\um{has no guarantees on its complete removal}).
\pietro{Experimental data on the pseudo-labeling strategy is provided in Sec.~\ref{sec:abl_pseudo}.}

\subsection{Preserving Old Classes on Old Domains}
\label{sec:oldD_oldC}

In Sec.~\ref{sec:newD_oldC} we focused on distilling old-task knowledge on the current novel domain.
Nonetheless, our ultimate target is to end up with a segmentation network capable to recognize all the observed \um{classes} over all the experienced domains, that is a prediction model robust to both domain and label distribution shifts. 
For this reason, at every novel incremental step it is required to preserve the task knowledge acquired in the past, that is, on past classes over past domains. 
To do so, we leverage the %
output\marco{-level} knowledge distillation objective in its standard formulation \cite{hinton2015distilling}, where we force a student model (\ie, the current model) to mimic the predicted classification probability distribution of a teacher model (\ie, the model saved and kept frozen \marco{since} the end of the previous step).
We opted for the objective in its standard fashion \cite{hinton2015distilling}, \marco{as} both image and label distributions ideally \marco{originate} from previous steps, so no domain shift should, in principle, affect the distillation process. 
In practice, we can not access former incremental datasets. 
Therefore, to retrieve the missing old-domain data, we resort \marco{once more} to stylization (Sec.~\ref{sec:domain_style}), \marco{so that} %
we \marco{can} leverage oldly-stylized data as proxy for the missing original images.
The final objective is of the following form:
\begin{equation}
    \mathcal{L}^{\tilde{o}}_{kd}(\mathcal{C}_{0:t\shortminus1},\mathcal{X}_{0:t\shortminus1}) =  
    - \frac{1}{t}
    \sum_{k=0}^{t\shortminus1}
    \frac{1}{
    |\tilde{\mathcal{D}}_{t\shortto k}|
    } 
    \!\!\!\!\!
    \sum_{
    \;\;\;
    \tilde{\mathbf{X}} \in 
    \tilde{\mathcal{D}}_{t\shortto k}
    } \!\!\!
    \!\!\!
    P_{t\shortminus1}(\tilde{\mathbf{X}}) \cdot \log \unkdP_{t}(\tilde{\mathbf{X}}),
\label{eq:kd}
\end{equation}
where $\unkdP_{t}(\tilde{\mathbf{X}}) \in \mathbb{R}^{H \times W \times |\mathcal{C}_{0:t\shortminus1}|}$ refers to the modified probability distribution \pietro{from Eq.~\eqref{eq:unb_ce}}, for which \textit{new} and \textit{unknown} categories are incorporated into a single output channel to address the label shift within the $u$ class.

The overall objective \marco{is given by}:
\begin{equation}
    \mathcal{L}_{tot} = \mathcal{L}_{{ce}}^{\tilde{n}}
    + \lambda_{{ce}}^{\tilde{o}} \cdot \mathcal{L}_{{ce}}^{\tilde{o}}
    + \lambda_{{\lossname}}^{\tilde{n}} \cdot \mathcal{L}_{{\lossname}}^{\tilde{n}}
    + \lambda_{{kd}}^{\tilde{o}} \cdot \mathcal{L}_{{kd}}^{\tilde{o}}.
    \label{eq:complete}
\end{equation}

\section{Experimental Setup}

In this section we provide a detailed description of the experimental setup utilized to validate the proposed framework against multiple competing methods. 
\marco{In Sec.~\ref{sec:exp_res} and \ref{sec:ablation} we will report the results of the evaluation campaign and extensive ablation studies as additional support.}

\subsection{Datasets}

To simulate the distribution shift at the input (image) level, we make use of multiple driving data sets, each limited to a specific geographic region \marco{or environmental factors}, and thus characterized by its distinctive low-level appearance (\eg, road pavement material, type of vehicles, %
\marco{light conditions}). On the contrary, the high-level semantic content is mostly consistent across image sets, that is, 
\marco{the road-related or other categories, moving and static obstacles}
can be found everywhere, and follow similar inter-class structural relations (\eg, the sky will always appear above the road).

\noindent
\textbf{Cityscapes. } The Cityscapes \cite{Cordts2016} dataset (CS) is a popular benchmark for autonomous driving applications. 
Images are collected across 50 cities, all located in Central Europe.

\noindent
\textbf{BDD100K. } The Berkeley DeepDrive dataset (BDD) \cite{yu2020bdd100k} is a more diverse collection of road scenes,  \pietro{captured with variable weather conditions at different times of the day}. 
Still, all samples are from 4 restricted localities in the US. %

\noindent
\textbf{IDD. } The Indian Driving Dataset (IDD) \cite{varma2029idd} includes driving scenes from Indian cities and their outskirts. 
It offers a diversified set of moving and static road obstacles, as well as a wilder and more natural environment, which breaks away from the typical European or American urban scenarios.

\noindent
\textbf{Mapillary Vistas. } The Mapillary Vistas dataset \cite{neuhold2017mapillary} contains images collected worldwide, with highly \pietro{diverse} acquisition settings and locations. 
Unlike previously introduced benchmarks, 
samples are not limited to a few cities located within quite uniform \marco{geographic} regions. 
We leverage the Mapillary \pietro{dataset} to generate continent-wise \marco{data} splits, as well as to test the domain generalization potential of the proposed class and domain incremental approach.

\noindent
\textbf{Shift. } The Shift benchmark \cite{sun2022shift} is a synthetic dataset for autonomous driving, designed to provide a plethora of distribution shifts, simulating the highly variable environmental conditions faced in real-world applications. 
We exploit it to mimic domain shift due to environmental diversity.

For BDD, IDD and Mapillary datasets, only the 19 classes available on Cityscapes were used.
For Shift, we considered the available 22 semantic categories.

\subsection{Incremental Learning Setup}
\label{sec:incr_setup}

\noindent
\textbf{Domain Incremental Setup. }
The first domain incremental setup is \pietro{created} by experiencing in succession the CS, BDD and IDD datasets (in different orders) during 3 separate learning steps. 
Additionally, we propose a 
further setup,
where domain shift across learning steps is achieved by splitting the entire Mapillary dataset into incremental sets based on \marco{geographic} proximity of samples, \ie, 6 separate data subsets are generated, grouping together pictures taken on the same continent. 
Finally, we leverage Shift to simulate incrementally variable environmental conditions, by partitioning the whole dataset into 3 groups of samples according to light conditions (\ie, \textit{daytime}, \textit{twilight} and \textit{night}).

\noindent
\textbf{Class Incremental Setup. } 
We start by following \cite{klingner2020class} to identify 3 separate groups within the 19 Cityscapes' classes, \ie, (i) \textit{background regions}, (ii) \textit{moving elements}, (iii) \textit{static elements}, which are observed incrementally under various arrangements.
Then, we extend the aforementioned 3-way class splitting to Shift in a similar fashion to \cite{klingner2020class}, this time on the 22 classes offered by the synthetic benchmark. 
\marco{All the class incremental sets are detailed in Table~\ref{tab:train_cil}.}

By merging class and domain individual settings, we devise each
\pietro{class and domain incremental setup} reported in Table \ref{tab:train_incr_setup}.
The first (\ie, \textit{urban}) is generated using CS, BDD and IDD datasets, together with the 3-way class split from \cite{klingner2020class}. 
Formally, we set the total number of \marco{learning} steps $T=3$, and at each step $0 \le t < T$:
\begin{equation}
    \mathcal{D}_{t} \subset (\mathcal{X}_{t}, \mathcal{C}_{t}) \in \{ 
    \text{CS}, \text{BDD}, \text{IDD} 
    \} 
    \times \{ \mathcal{C}_{bgr}, \mathcal{C}_{stat}, \mathcal{C}_{mov} \},
\end{equation}
where each dataset and class split is observed once.
\\
We further propose an incremental setup (\ie, \textit{\tocheck{worldwide}}) based on continent-wise splitting of the Mapillary dataset.
To match the increase in domain set size %
to 6 elements, we divide each class group \cite{klingner2020class} in half, for a total of 6 class splits (Table~\ref{tab:train_cil}).
We set $T=6$, and at each step $0 \le t < T$: 
\begin{equation}
\begin{split}
    \mathcal{D}_{t} \subset (\mathcal{X}_{t}, \mathcal{C}_{t}) \in & \{ 
    \text{EU}, \text{NA}, \text{AS}, \text{OC}, \text{AF}, \text{SA} 
    \} 
    \times \\
    &\{ \mathcal{C}^{0}_{bgr}, \mathcal{C}^{1}_{bgr}, \mathcal{C}^{0}_{stat}, \mathcal{C}^{1}_{stat}, \mathcal{C}^{0}_{mov}, \mathcal{C}^{1}_{mov} \},
\end{split}
\end{equation}
where each class set and \marco{each} domain appears only in a single step. 
Among the large number of possible incremental sequences, 
we perform the experimental evaluation in the $\text{EU} \rightarrow\allowbreak \text{NA} \rightarrow\allowbreak \text{AS} \rightarrow\allowbreak \text{OC} \rightarrow\allowbreak \text{AF} \rightarrow\allowbreak \text{SA}$
and 
{$\mathcal{C}^{0}_{bgr}\rightarrow\allowbreak  \mathcal{C}^{1}_{bgr}\rightarrow\allowbreak \mathcal{C}^{0}_{stat}\rightarrow\allowbreak \mathcal{C}^{1}_{stat}\rightarrow\allowbreak \mathcal{C}^{0}_{mov}\rightarrow\allowbreak \mathcal{C}^{1}_{mov}$}
\pietro{setups}.
\\
Finally, the last setup (\ie, \textit{\tocheck{environmental}}) combines the environmental partitioning chosen for Shift with the 3-way class splitting from \cite{klingner2020class}.

\begin{table}[!t]
    \centering
    \setlength{\tabcolsep}{0.3pt}
    \renewcommand{\arraystretch}{1.}
    \caption{\marco{Split of Cityscapes's (CS) and Shift's class sets following the criterion proposed by \cite{klingner2020class}.}}
    \begin{tabular}{c@{\hspace{1mm}}c|cc @{\hspace{-1.4mm}}c}
    \toprule
    
    \multicolumn{2}{c|}{} & \multicolumn{1}{c}{$\mathcal{C}_{bgr}$} & $\mathcal{C}_{stat}$ & $\mathcal{C}_{mov}$ \\
    
    \midrule
    \multirow{3}{*}{\vspace{-1.mm} \rotatebox[origin=c]{90}{CS}} &
    \multirow{2}{*}{$\mathcal{C}^{0}$}
    & \multirow{2}{*}{\{road, sidewalk\}} 
    & \multirow{2}{*}{\{build., wall, fence\}} 
    & \{person, rider, \\
    & & & & motorcycle, bicycle\} \\
    
    \rule{0pt}{7.5pt} %
    &
    \multirow{1}{*}{$\mathcal{C}^{1}$} 
    & \{veg., terr., sky\} & \{pole, t.\ light, t.\ sign\}  & \{car, truck, bus, train\} \\
    
    \midrule
    
    \multirow{3}{*}{\rotatebox[origin=c]{90}{Shift}} &
    \multirow{3}{*}{$\mathcal{C}^{s}$}
    & \{r.line, road, veg.,
    & \{build., wall, fence, pole,
    & \{pedestrian, \\
    
    &
    & ground,  water, 
    & t.\ light, bridge, r.track,  
    & vehicles, \\
    
    &
    &   s.walk, terr., sky\} 
    &  g.rail, t.\ sign, static\}
    & dynamic\} \\

    \bottomrule
    \end{tabular}
    \label{tab:train_cil}
\end{table}

\begin{table}[!t]
    \centering
    \setlength{\tabcolsep}{5.pt}
    \caption{Class and domain incremental sets.}
    \begin{tabular}{c|cc}
    \toprule
    \multicolumn{1}{c|}{} & \multicolumn{1}{c}{Class sets} & Domains \\
    
        \midrule
    \multirow{1}{*}{Urban} 
    & $\{ \mathcal{C}_{bgr}, \mathcal{C}_{stat}, \mathcal{C}_{mov} \}$ & $ \{ \text{CS}, \text{BDD}, \text{IDD} \}$\\
    
    \midrule
    \multirow{2}{*}{\tocheck{Worldwide}} 
    & $\{ \mathcal{C}^{0}_{bgr}, \mathcal{C}^{1}_{bgr}, \mathcal{C}^{0}_{stat},$
    & $\{ \text{EU}, \text{NA}, \text{AS},$ \\
    
    & $\mathcal{C}^{1}_{stat}, \mathcal{C}^{0}_{mov}, \mathcal{C}^{1}_{mov} \}$
    & $\text{OC}, \text{AF}, \text{SA} \}$ \\

    \midrule
    \multirow{1}{*}{\tocheck{Environmental}} 
    & $\{ \mathcal{C}^{s}_{bgr}, \mathcal{C}^{s}_{stat}, \mathcal{C}^{s}_{mov} \}$ & $ \{ \text{Daytime}, \text{Twilight}, \text{Night} \}$ \\
    
    \bottomrule
    \end{tabular}
    \vspace{1mm}
    \\ \marco{$\mathcal{C}^{s}$ indicates that the class subset is derived from Shift's original set.}
    \label{tab:train_incr_setup}
\end{table}

\subsection{Implementation Details}

We \marco{built} %
our framework in PyTorch.
Due to the complexity of the investigated problem, 
\marco{in most experiments}
we use a lightweight segmentation model, \ie,  ErfNet \cite{romera2018erfnet}.
We argue that a smaller network complies more realistically to deployment-related constraints in real-word applications, 
\eg,
in terms of memory occupation and inference speed. 
Yet, for comparison purposes we report additional results with the heavier and better performing DeeplabV3 architecture \cite{chen2017rethinking} with ResNet101 backbone \cite{he2016deep}. 
In all experiments, the segmentation model is pre-trained on ImageNet \cite{deng2009imagenet}.

With ErfNet, we use the Adam optimizer \cite{kingma2015adam} and learning rate set to $5\mathrm{e}{\shortminus4}$.
With DeeplabV3, we use the SGD optimizer and learning rate set to $1\mathrm{e}{\shortminus3}$.
Weight decay is fixed to $1\mathrm{e}{\shortminus4}$, and we employ a polynomial decay of power $0.9$ for learning rate scheduling.
We train for 100 and 50 epochs at each learning step, with ErfNet and DeeplabV3 respectively \marco{(except in Shift, where we set the number of epochs to 10)}.
With ErfNet we use a batch size of 6, with DeeplabV3 we reduce its value to 2 \pietro{due to} GPU memory constraints.

When experimentally evaluating on Cityscapes-BDD-IDD and Shift setups, images are resized to $512\times1024$ resolution.
When using Mapillary for training, inputs are first resized to $1024$ width (fixed aspect ratio), and then cropped to $512\times1024$.
This pre-processing is done to accommodate for the highly variable aspect ratios of Mapillary's samples.

The $\beta$ parameter controlling the size of the style window is empirically set to $1\mathrm{e}{\shortminus2}$ and fixed in all experiments.
Plus, we 
\marco{experimentally fix}
$\lambda_{{ce}}^{\tilde{o}} \!=\! \lambda_{{\lossname}}^{\tilde{n}} \!=\! \lambda_{{kd}}^{\tilde{o}} \!=\! 10$, and keep them unchanged in every incremental setup.
This shows that our approach is robust to change of experimental setting, and requires minimal hyper-parameter tuning.
 Ablation studies \pietro{on the impact of $\beta$ and loss weights are in Sec.\ \ref{sec:ablation}.}

\subsection{Competitors}
To the best of our knowledge, this is the first work explicitly modeling and addressing class and domain incremental learning in semantic segmentation.
For this reason, we compare with other methods targeting class (CIL) or domain (DIL) incremental learning as individual problems.

Among class-incremental methods, we consider ILT \cite{michieli2019incremental} and MiB \cite{cermelli2020modeling}, along with state-of-the-art PLOP \cite{douillard2021plop} and UCD \cite{yang2022uncertainty}.
When using PLOP with ErfNet, we apply the \textit{LocalPOD} loss \cite{douillard2021plop} on embeddings extracted at the end of the first and second blocks, as well as at the output of the encoder. %
For UCD, we modify the contrastive distillation loss so that the maximum number of positives and negatives is set to 3000 each (which are randomly selected among the whole sets as defined in the original work).
We perform this adjustment to meet GPU memory limitations. 
All experiments were performed on a RTX Titan GPU with 24GB of memory.
We believe that a fair comparison should involve comparable GPU resources for all the competitors.

On the domain-incremental side, we compare with \cite{garg2022multi}.
Differently from our setup, the\um{y} assume to have full task supervision on all the domains incrementally encountered.
We adapt their framework to a class-incremental setup by replacing the standard cross-entropy loss with the unbiased version from \cite{cermelli2020modeling}, to prevent the background shift from erasing the task-knowledge learned in past steps.

\subsection{Metrics}
\label{sec:metrics}

\marco{Inspired by \cite{garg2022multi}, to provide a valuable measure of prediction performance across multiple tasks and domains, we resort to a}
domain average relative performance w.r.t.\ a fully-supervised \marco{\textit{oracle}} reference \um{(the smaller the better)} \marco{defined at any step $t$ as}:
\begin{equation}
    \bar{\Delta}_{t} = 
    \underbrace{\frac{1}{t+1} \, \,
    \sum_{k=0}^{t}}_{\text{domain avg}}
    \underbrace{ \, \,
    \frac{
    A_{\mathcal{X}_{k}|S_{t}}^{\mathcal{C}_{0:t}} -
    A_{\mathcal{X}_{k}| \marco{S^{*}} }^{\mathcal{C}_{0:t}}
    }{
    A_{\mathcal{X}_{k}| \marco{S^{*}} }^{\mathcal{C}_{0:t}}
    }
    }_{\marco{\substack{
    \Delta^k_{t} \text{: relative acc.\ gap w.r.t.} \\ \text{oracle on step-$k$ domain}
    }}
    },
    \label{metric:rel}
\end{equation}
where $A_{\mathcal{X}|S}^{\mathcal{C}}$ is the class-average accuracy (we make use of the commonly employed mIoU metric \cite{minaee2021image}) attained by segmentation network $S$ on domain $\mathcal{X}$ and class set $\mathcal{C}$. 
\marco{$S^{*}$ is the oracle segmentation model, \ie, trained with full supervision on the entire pool of classes and domains (even classes and domains that will be observed after step $t$).
}

\marco{We further provide a measure of generalization aptitude \um{(the higher the better)},
expressed as the accuracy (\ie, in terms of mIoU) achieved over the entire class set observed so far on a novel dataset never experienced before.
At step $t$, the metric follows:}
\begin{equation}
    \Gamma^{gen}_{t} = A_{\mathcal{X}_{ext}|S_{t}}^{\mathcal{C}_{0:t}} = 
    \frac{1}{|\mathcal{C}_{0:t}|} \sum_{c \in \mathcal{C}_{0:t}} A^{c}_{\mathcal{X}_{ext}|S_{t}},
\label{eq:gen_metric}
\end{equation}
\marco{where $\mathcal{X}_{ext}$ is the unseen domain.}

\section{Experimental Results}
\label{sec:exp_res}

\subsection{Evaluation on Urban Scenes}

\begin{table*}[t!]
 \scriptsize
 \centering
  \setlength{\tabcolsep}{4.3pt}
  \renewcommand{\arraystretch}{1.}
 \caption{Experimental results on $\text{CS} \protect\rightarrow  \text{BDD} \protect\rightarrow  \text{IDD}$ domain setup and $\mathcal{C}_{bgr} \protect\rightarrow  \mathcal{C}_{stat} \protect\rightarrow  \mathcal{C}_{mov}$ class setup. 
 }
  \begin{tabu}{l |cc|c| cc|cc|c| cc|cc|cc|c}
  \toprule
 \multicolumn{1}{c|}{\multirow{3}{*}{\vspace{-2mm}\shortstack{ $\text{CS} \veryshortarrow \text{BDD} \veryshortarrow \text{IDD}$ \\
 $\mathcal{C}_{bgr} \veryshortarrow \mathcal{C}_{stat} \veryshortarrow \mathcal{C}_{mov}$}}}
 & \multicolumn{3}{c|}{Step 0} 
  & \multicolumn{5}{c|}{Step 1} 
  & \multicolumn{7}{c}{Step 2} \\
  
 \rule{0pt}{7pt} %
  & \multicolumn{2}{c}{\textbf{CS ($\mathcal{X}_0$)}} &
   & \multicolumn{2}{c}{\textbf{BDD ($\mathcal{X}_1$)}} & \multicolumn{2}{c}{CS ($\mathcal{X}_0$)}&
   & \multicolumn{2}{c}{\textbf{IDD ($\mathcal{X}_2$)}} & \multicolumn{2}{c}{BDD ($\mathcal{X}_1$)} & \multicolumn{2}{c}{CS ($\mathcal{X}_0$)} & \\
   
  \rule{0pt}{7pt} %
  & 
  $\text{mIoU}_{0}^{0}$ $ \! \uparrow$ & $\Delta^0_{0} \! \downarrow$ & $\bar{\Delta}_{0} \! \downarrow$ &
  $\text{mIoU}_{1}^{1}$ $ \! \uparrow$ & $\Delta^1_{1} \! \downarrow$ & $\text{mIoU}_{1}^{0}$ $ \! \uparrow$ & $\Delta^0_{1} \! \downarrow$ & $\bar{\Delta}_{1} \! \downarrow$ &
  $\text{mIoU}_{2}^{2}$ $ \! \uparrow$ & $\Delta^2_{2} \! \downarrow$ & $\text{mIoU}_{2}^{1}$ $ \! \uparrow$ & $\Delta^1_{2} \! \downarrow$ & $\text{mIoU}_{2}^{0}$ $ \! \uparrow$ & $\Delta^0_{2} \! \downarrow$ & $\bar{\Delta}_{2} \! \downarrow$ \\
  
  \midrule
  FT ($\mathcal{L}_{ce}^{n}$)          & 79.67 & \oracle{6.44}{5.32} & \oracle{6.44}{5.32}   
                                       & 24.38 & \oracle{61.51}{61.35} & 18.11 & \oracle{75.18}{74.06} & \oracle{68.35}{67.71}   
                                       & 26.27 & \oracle{61.45}{61.48} & 10.47 & \oracle{78.90}{81.72} & 12.10 & \oracle{82.18}{81.18} & \oracle{74.18}{74.79} \\
                                     
  FT w/ self-style ($\mathcal{L}_{ce}^{\tilde{n}}$)  & 79.19 & \oracle{7.0}{5.89}  & \oracle{7.0}{5.89}   
                                       & 20.41 & \oracle{67.78}{67.65} & 19.08 & \oracle{73.86}{72.67} & \oracle{70.82}{70.16}   
                                       & 27.12 & \oracle{60.20}{60.24} & 11.51 & \oracle{76.80}{79.91} & 13.68 & \oracle{79.86}{78.72} & \oracle{72.29}{72.95} \\
                                       
  MDIL \cite{garg2022multi}         & 80.35 & \oracle{5.64}{4.51} & \oracle{5.64}{\tblbold{4.51}}   
                                    & 26.12 & \oracle{58.76}{58.59} & 23.65 & \oracle{67.59}{66.13} & \oracle{63.18}{62.36}   
                                    & 28.10 & \oracle{58.76}{58.80} & 12.46 & \oracle{74.89}{78.25} & 13.22 & \oracle{80.53}{79.44} & \oracle{71.39}{72.16} \\
                                    
  ILT \cite{michieli2019incremental}   & 79.67 & \oracle{6.44}{5.32} & \oracle{6.44}{5.32}    
                                       & 22.21 & \oracle{}{64.80} & 44.70 & \oracle{}{35.99} & \oracle{}{50.39}   
                                       & 26.69 & \oracle{}{60.87} & 16.70 & \oracle{}{70.85} & 29.76 & \oracle{}{53.71} & \oracle{}{61.81} \\
                                    
  MiB \cite{cermelli2020modeling} %
  & 79.67 & \oracle{6.44}{5.32} & \oracle{6.44}{5.32}   
  & 34.35 & \oracle{45.77}{45.55} & 49.24 & \oracle{32.53}{29.48} & \oracle{39.15}{37.51}   
  & 42.58 & \oracle{37.51}{37.57} & 26.36 & \oracle{46.88}{53.98} & 36.58 & \oracle{46.14}{43.10} & \oracle{43.51}{44.88} \\ 
  
  PLOP \cite{douillard2021plop}     & 79.67 & \oracle{6.44}{5.32} & \oracle{6.44}{5.32}    
                                    & 36.78 & \oracle{41.93}{41.70} & 50.05 & \oracle{31.42}{28.32} & \oracle{36.68}{35.01}   
                                    & 43.15 & \oracle{36.67}{36.73} & 27.24 & \oracle{45.10}{52.44} & 36.84 & \oracle{45.75}{42.70} & \oracle{42.51}{43.96} \\
                                    
  UCD \cite{yang2022uncertainty}    & 79.67 & \oracle{6.44}{5.32} & \oracle{6.44}{5.32}   
                                    & 35.45 & \oracle{44.03}{43.80} & 50.38 & \oracle{30.97}{27.85} & \oracle{37.50}{35.83}   
                                    & 43.19 & \oracle{36.61}{36.67} & 27.38 & \oracle{44.81}{52.19} & 37.34 & \oracle{45.01}{41.91} & \oracle{42.15}{43.59} \\
                                    
  \name \; w/o $\mathcal{L}_{kd}^{\tilde{o}}$
                            & 79.19 & \oracle{7.0}{5.89}  & \oracle{7.0}{5.89}  
                            & 44.41 & \oracle{29.89}{29.60} & 50.77 & \oracle{30.43}{27.29} & \oracle{30.16}{28.44}   
                            & 50.70 & \oracle{25.59}{25.66} & 34.86 & \oracle{29.75}{39.14} & 43.04 & \oracle{36.62}{33.05} & \oracle{30.65}{32.62} \\
   \name
                            & 79.19 & \oracle{7.0}{5.89}  & \oracle{7.0}{5.89}   
                            & 44.47 & \oracle{29.79}{29.51} & 53.31 & \oracle{26.95}{23.65} & \oracle{28.37}{\tblbold{26.58}}   
                            & 51.20 & \oracle{24.86}{24.93} & 35.73 & \oracle{27.99}{37.62} & 44.17 & \oracle{34.96}{31.29} & \oracle{29.27}{\tblbold{31.28}} \\
  \midrule
  Oracle                      & \oracle{85.15}{84.15} & - & -   & \oracle{63.34}{63.08} & - & \oracle{72.98}{69.82} & - & -   & \oracle{68.14}{68.20} & - & \oracle{49.62}{57.28} & - & \oracle{67.91}{64.29} & - & -\\
  \bottomrule
  
  \end{tabu}
 \label{tab:CBI}
\end{table*}

\begin{table*}[t!]
 \scriptsize
 \centering
  \setlength{\tabcolsep}{4.3pt}
  \renewcommand{\arraystretch}{1.}
 \caption{Experimental results on $\text{BDD} \protect\rightarrow  \text{IDD} \protect\rightarrow  \text{CS}$ domain setup and $\mathcal{C}_{bgr} \protect\rightarrow  \mathcal{C}_{stat} \protect\rightarrow  \mathcal{C}_{mov}$ class setup.}
  \begin{tabu}{l |cc|c| cc|cc|c| cc|cc|cc|c}
  \toprule
  
   \multicolumn{1}{c|}{\multirow{3}{*}{\vspace{-2mm}\shortstack{ $\text{BDD} \veryshortarrow \text{IDD} \veryshortarrow \text{CS}$ \\
 $\mathcal{C}_{bgr} \veryshortarrow \mathcal{C}_{stat} \veryshortarrow \mathcal{C}_{mov}$}}}
 
  & \multicolumn{3}{c|}{Step 0} 
  & \multicolumn{5}{c|}{Step 1} 
  & \multicolumn{7}{c}{Step 2} \\
  
  \rule{0pt}{7pt} %
  & \multicolumn{2}{c}{\textbf{BDD ($\mathcal{X}_0$)}} &
   & \multicolumn{2}{c}{\textbf{IDD ($\mathcal{X}_1$)}} & \multicolumn{2}{c}{BDD ($\mathcal{X}_0$)}&
   & \multicolumn{2}{c}{\textbf{CS ($\mathcal{X}_2$)}} & \multicolumn{2}{c}{IDD ($\mathcal{X}_1$)} & \multicolumn{2}{c}{BDD ($\mathcal{X}_0$)} & \\
  
  \rule{0pt}{7pt} %
  & 
  $\text{mIoU}_{0}^{0}$ $ \! \uparrow$ & $\Delta^0_{0} \! \downarrow$ & $\bar{\Delta}_{0} \! \downarrow$ &
  $\text{mIoU}_{1}^{1}$ $ \! \uparrow$ & $\Delta^1_{1} \! \downarrow$ & $\text{mIoU}_{1}^{0}$ $ \! \uparrow$ & $\Delta^0_{1} \! \downarrow$ & $\bar{\Delta}_{1} \! \downarrow$ &
  $\text{mIoU}_{2}^{2}$ $ \! \uparrow$ & $\Delta^2_{2} \! \downarrow$ & $\text{mIoU}_{2}^{1}$ $ \! \uparrow$ & $\Delta^1_{2} \! \downarrow$ & $\text{mIoU}_{2}^{0}$ $ \! \uparrow$ & $\Delta^0_{2} \! \downarrow$ & $\bar{\Delta}_{2} \! \downarrow$ \\

  \midrule
  FT ($\mathcal{L}_{ce}^{n}$)      & 72.22 & \oracle{5.10}{6.61} & \oracle{5.10}{6.61}   
                                  & 33.37 & \oracle{51.54}{52.43} & 20.49 & \oracle{67.65}{67.52} & \oracle{59.60}{59.98}   
                                  & 23.36 & \oracle{65.60}{65.60} & 7.09 & \oracle{89.6}{89.60} & 5.45 & \oracle{89.02}{89.02} & \oracle{81.40}{81.41} \\
  FT w/ self-style ($\mathcal{L}_{ce}^{\tilde{n}}$)  & 72.12 & \oracle{5.12}{6.74} & \oracle{5.12}{6.74}   
                                       & 33.27 & \oracle{51.68}{52.58} & 21.12 & \oracle{66.66}{66.52} & \oracle{59.17}{59.55}   
                                       & 28.52 & \oracle{58.00}{55.64} & 15.44 & \oracle{77.34}{77.36} & 14.24 & \oracle{71.30}{75.14} & \oracle{68.88}{69.38} \\
                                       
  MDIL \cite{garg2022multi} & 72.44 & \oracle{4.81}{6.33} & \oracle{4.81}{\tblbold{6.33}}   
                            & 26.78 & \oracle{61.10}{61.83} & 15.44 & \oracle{75.62}{75.52} & \oracle{68.36}{68.68}   
                            & 25.52 & \oracle{62.42}{60.30} & 11.61 & \oracle{82.96}{82.98} & 10.77 & \oracle{78.30}{81.20} & \oracle{74.56}{74.83} \\
                            
  ILT \cite{michieli2019incremental}   & 72.22 & \oracle{5.10}{6.61} & \oracle{5.10}{6.61}    
                                       & 42.10 & \oracle{}{39.98} & 43.00 & \oracle{}{31.84} & \oracle{}{35.91}   
                                       & 33.33 & \oracle{}{48.15} & 26.93 & \oracle{}{60.52} & 29.68 & \oracle{}{48.19} & \oracle{}{52.29} \\

  MiB \cite{cermelli2020modeling} %
  & 72.22 & \oracle{5.10}{6.61} & \oracle{5.10}{6.61}   
  & 52.18 & \oracle{24.21}{25.62} & 45.28 & \oracle{28.51}{28.22} & \oracle{26.36}{26.92}   
  & 48.22 & \oracle{29.00}{25.00} & 33.57 & \oracle{50.73}{50.77} & 30.94 & \oracle{37.64}{45.98} & \oracle{39.12}{40.58} \\
  
  PLOP \cite{douillard2021plop}                     & 72.22 & \oracle{5.10}{6.61} & \oracle{5.10}{6.61}  
                                                    & 53.15 & \oracle{22.80}{24.24} & 44.25 & \oracle{30.14}{29.85} & \oracle{26.47}{27.05}   
                                                    & 47.21 & \oracle{30.48}{26.56} & 35.36 & \oracle{48.10}{48.15} & 32.02 & \oracle{35.47}{44.10} & \oracle{38.02}{39.60} \\
                                                    
  UCD \cite{yang2022uncertainty}                      & 72.22 & \oracle{5.10}{6.61} & \oracle{5.10}{6.61}   
                                                      & 52.42 & \oracle{23.86}{25.28} & 45.20 & \oracle{28.64}{28.35} & \oracle{26.25}{26.81}   
                                                      & 48.40 & \oracle{28.74}{24.72} & 32.60 & \oracle{52.15}{52.19} & 28.95 & \oracle{41.66}{49.47} & \oracle{40.85}{42.13} \\
                                                      
  \name \; w/o $\mathcal{L}_{kd}^{\tilde{o}}$
                            & 72.12 & \oracle{5.12}{6.74} & \oracle{5.12}{6.74}    
                            & 54.34 & \oracle{21.07}{21.07} & 41.36 & \oracle{34.70}{34.44} & \oracle{27.89}{27.75}   
                            & 52.56 & \oracle{22.61}{18.24} & 36.70 & \oracle{46.14}{46.19} & 32.33 & \oracle{34.84}{43.56} & \oracle{34.53}{36.00} \\
  \name
                            & 72.12 & \oracle{5.12}{6.74} & \oracle{5.12}{6.74}    
                            & 54.53 & \oracle{20.80}{20.80} & 43.98 & \oracle{30.57}{30.28} & \oracle{25.68}{\tblbold{25.54}}   
                            & 52.63 & \oracle{22.50}{18.14} & 38.14 & \oracle{44.03}{44.08} & 34.03 & \oracle{31.42}{40.59} & \oracle{32.65}{\tblbold{34.27}} \\
  \midrule
  Oracle                      & \oracle{76.10}{77.33} & - & -   & \oracle{68.85}{70.16} & - & \oracle{63.34}{63.08} & - & -   & \oracle{68.14}{68.20} & - & \oracle{49.62}{57.28} & - & \oracle{67.91}{64.29} & - & -\\
  \bottomrule
 \end{tabu}
 \label{tab:BIC}
\end{table*}

The first experimental setup we explore entails incrementally transitioning between urban and suburban areas of different regions around the world.
High- and low- level image contents undergo distribution shifts of different extent: although it might be reasonable to assume that the basic semantic structure of road images is invariant to \marco{geographic} location, scene elements are likely to change appearance significantly when travelling around the world.

\subsubsection{
\marco{Study on Domain Ordering}
}
To reproduce class and domain distribution shifts, we train on the Cityscapes, BDD and IDD datasets in an incremental fashion.
The class incremental protocol is instead the one \marco{proposed} in \cite{klingner2020class} (\ie, $\mathcal{C}_{bgr}\rightarrow \mathcal{C}_{stat}\rightarrow \mathcal{C}_{mov}$).
As detailed in Sec.~\ref{sec:incr_setup}, we define a total of 3 learning steps.
In Tables \ref{tab:CBI}, \ref{tab:BIC} and \ref{tab:ICB} we report experimental results following 3 different dataset orders, so that each dataset is viewed at all the 3 possible learning steps, considering all experiments performed.

We report results in terms of %
\um{mIoU} computed over all 
classes excluding the \textit{unknown} one, as typically done in the literature.  
\tocheck{The mIoU is computed for each domain $\mathcal{X}_{k}$ (\ie, dataset) experienced up to a current step 
$t$ (\ie, $\text{mIoU}^{k}_{t}$, $k \!\leq\! t$), $ \forall t \!<\! T$}.
\tocheck{In addition, we provide a measure of relative performance w.r.t.\ a supervised reference, %
both for individual domains $\Delta_{t}^k$, %
and as a global quantity $\bar\Delta_{t}$ (Eq.~\eqref{metric:rel}).
The supervised reference, denoted as \textit{Oracle}, corresponds to the joint training over both class sets and domains.}

We compare with methods %
addressing \classincr learning (ILT \cite{michieli2019incremental}, MiB \cite{cermelli2020modeling}, PLOP \cite{douillard2021plop} and UCD \cite{yang2022uncertainty}) and with a recent \domincr method (MDIL \cite{garg2022multi}).
We also include a simple baseline, activating only the \marco{task} loss on the new \marco{classes} and new domain (Eq.~\eqref{eq:nc_nd}).
This approach is usually referred to as \textit{fine-tuning}, as the focus is just posed on learning the new task.
Two variants are reported for this baseline, \ie, with or without self-stylization applied on input images, %
indicated \pietro{respectively as $\mathcal{L}_{ce}^{\tilde{n}}$ and $\mathcal{L}_{ce}^{n}$.}
As for \marco{our} approach, we evaluate its final form (Eq.~\eqref{eq:complete}), complete of all the training objectives detailed in Sec.\ \ref{sec:method}, as well as a simpler configuration without the $\marco{\mathcal{L}_{kd}^{\tilde{o}}}$ loss (Eq.~\eqref{eq:kd}).  %

By inspecting results in Tables \ref{tab:CBI}, \ref{tab:BIC} and \ref{tab:ICB}, %
we notice that the performance achieved by different methods at the end of the \textbf{initial learning step} are comparable.
This is due to the similar objectives employed so far, to learn just the first class set ($\mathcal{C}_{bgr}$) on the first domain, regardless of the domain order.  %
We remark that the proposed self-stylization is not detrimental when learning the current task. %
\tocheck{We will provide some ablation studies on the impact of stylization in Sec.~\ref{sec:ablation}}

When progressing to the \textbf{first incremental step}, catastrophic forgetting has to be addressed to retain good performance.
We observe that the $\mathcal{L}_{ce}^{n}$ and $\mathcal{L}_{ce}^{\tilde{n}}$ losses alone are not sufficient to achieve satisfactory results, being focused on the new task and providing no  constraints \marco{to preserve} past knowledge.
MDIL \cite{garg2022multi} performs poorly as well, since the proposed dynamic architecture is not suitable to address partial \classincr supervision, which in our setup is present along with domain incremental shift.
\\
By analyzing \classincr learning methods, we note that they are able to preserve previously acquired knowledge to some extent, while allowing some plasticity for learning the new task.
Still, the domain shift between previous and current datasets has a negative impact on the prediction accuracy of the incrementally trained predictor.
All the considered CIL methods, in fact, rely on the ability of a segmentation model frozen from the previous step to preserve knowledge of the past.
Yet, because of the domain discrepancy between past and new data, this distillation mechanism could introduce unreliable guidance on former tasks, as the frozen model is subject to a shift in the experienced distribution at the input level when fed with new domain data.
At the same time, the distribution gap may hinder the transferability of new-class knowledge to old domains, which are no longer available as training data.
\\
These drawbacks are revealed \marco{by} results of
Table \ref{tab:ICB} ($\text{IDD} \rightarrow\allowbreak \text{CS} \rightarrow\allowbreak \text{BDD}$): the significant domain shift between the Cityscapes and IDD datasets prevents CIL methods from effectively preserving and learning \tocheck{task-related clues} %
on IDD, which was experienced at step 0. 
On the contrary, our approach addresses domain shift by leveraging the stylization scheme and applying carefully designed objectives to suitably tackle the \marco{general} class and domain incremental learning.
In particular, the proposed objectives $\mathcal{L}^{\tilde{o}}_{ce}$ (Eq.~\eqref{eq:nc_od}) and $\mathcal{L}^{\tilde{o}}_{kd}$ (Eq.~\eqref{eq:kd}) are specifically designed to address the aforementioned problems affecting CIL methods and \pietro{allow to achieve} superior accuracy on former domains.
As a result,  \name\ \um{improves accuracy by more than 17 mIoU points on  IDD}  at step 1 w.r.t.\ the best competitor \um{(\ie,} \um{UCD} \cite{yang2022uncertainty}).
\\
We also remark that, even with alternative domain orders (Tables \ref{tab:CBI} and \ref{tab:BIC}), %
\name\ shows the best stability-plasticity trade-off, retaining the best overall accuracy in terms of $\bar\Delta_{1}$.
Furthermore, we can see that, for both $\text{CS} \rightarrow \text{BDD} \rightarrow \text{IDD}$ and $\text{BDD} \rightarrow \text{IDD} \rightarrow \text{CS}$ orders, the addition of the $\marco{\mathcal{L}_{kd}^{\tilde{o}}}$ objective in \um{ \name\ leads to a boost} in performance on the past domain, which coincides with the design purpose of the objective.

\begin{table*}[!t]
 \scriptsize
 \centering
  \setlength{\tabcolsep}{4.3pt}
  \renewcommand{\arraystretch}{1.}
 \caption{Experimental results on $\text{IDD} \protect\rightarrow  \text{CS} \protect\rightarrow  \text{BDD}$ domain setup and $\mathcal{C}_{bgr} \protect\rightarrow  \mathcal{C}_{stat} \protect\rightarrow  \mathcal{C}_{mov}$ class setup.}
  \begin{tabu}{l |cc|c| cc|cc|c| cc|cc|cc|c}
  \toprule
  
  \multicolumn{1}{c|}{\multirow{3}{*}{\vspace{-2mm}\shortstack{ $\text{IDD} \veryshortarrow \text{CS} \veryshortarrow \text{BDD}$ \\
 $\mathcal{C}_{bgr} \veryshortarrow \mathcal{C}_{stat} \veryshortarrow \mathcal{C}_{mov}$}}}
  
  & \multicolumn{3}{c|}{Step 0} 
  & \multicolumn{5}{c|}{Step 1} 
  & \multicolumn{7}{c}{Step 2} \\
  
  \rule{0pt}{7pt} %
    & \multicolumn{2}{c}{\textbf{IDD ($\mathcal{X}_0$)}} &
   & \multicolumn{2}{c}{\textbf{CS ($\mathcal{X}_1$)}} & \multicolumn{2}{c}{IDD ($\mathcal{X}_0$)}&
   & \multicolumn{2}{c}{\textbf{BDD ($\mathcal{X}_2$)}} & \multicolumn{2}{c}{CS ($\mathcal{X}_1$)} & \multicolumn{2}{c}{IDD ($\mathcal{X}_0$)} & \\
   
   \rule{0pt}{7pt} %
  & 
  $\text{mIoU}_{0}^{0}$ $ \! \uparrow$ & $\Delta^0_{0} \! \downarrow$ & $\bar{\Delta}_{0} \! \downarrow$ &
  $\text{mIoU}_{1}^{1}$ $ \! \uparrow$ & $\Delta^1_{1} \! \downarrow$ & $\text{mIoU}_{1}^{0}$ $ \! \uparrow$ & $\Delta^0_{1} \! \downarrow$ & $\bar{\Delta}_{1} \! \downarrow$ &
  $\text{mIoU}_{2}^{2}$ $ \! \uparrow$ & $\Delta^2_{2} \! \downarrow$ & $\text{mIoU}_{2}^{1}$ $ \! \uparrow$ & $\Delta^1_{2} \! \downarrow$ & $\text{mIoU}_{2}^{0}$ $ \! \uparrow$ & $\Delta^0_{2} \! \downarrow$ & $\bar{\Delta}_{2} \! \downarrow$ \\
  
  \midrule
   FT ($\mathcal{L}_{ce}^{n}$)   & 78.80 & \oracle{9.54}{{8.52}} & \oracle{9.54}{\tblbold{8.52}}   
                                & 9.66 & \oracle{49.64}{47.37} & 8.64 & \oracle{92.94}{93.07} & \oracle{71.29}{70.22}   
                                & 9.66 & \oracle{80.53}{83.14} & 8.64 & \oracle{87.28}{86.56} & 7.09 & \oracle{89.59}{89.60} & \oracle{85.80}{86.43}\\
  FT w/ self-style ($\mathcal{L}_{ce}^{\tilde{n}}$)   & 78.78 & \oracle{9.56}{8.55} & \oracle{9.56}{8.55}   
                                        & 42.11 & \oracle{42.30}{39.69} & 19.81 & \oracle{71.23}{71.76} & \oracle{56.76}{55.73}   
                                        & 14.05 & \oracle{71.68}{75.47} & 12.63 & \oracle{81.40}{80.35} & 11.01 & \oracle{83.84}{83.86} & \oracle{78.98}{79.89} \\
  MDIL \cite{garg2022multi} & 78.72 & \oracle{9.63}{8.62} & \oracle{9.63}{8.62}   
                            & 34.87 & \oracle{52.22}{50.06} & 11.70 & \oracle{83.01}{83.32} & \oracle{67.61}{66.69}   
                            &  8.90 & \oracle{82.06}{84.46} & 8.22 & \oracle{87.90}{87.21} & 6.70 & \oracle{90.17}{90.18} & \oracle{86.71}{87.28} \\
                            
  ILT \cite{michieli2019incremental}   & 78.80 & \oracle{9.54}{{8.52}} & \oracle{9.54}{\tblbold{8.52}}     
                                       & 44.44 & \oracle{}{36.35} & 43.32 & \oracle{}{38.26} & \oracle{}{37.30}   
                                       & 24.48 & \oracle{}{57.26} & 30.00 & \oracle{}{53.34} & 27.88 & \oracle{}{59.12} & \oracle{}{56.57} \\
                            
  MiB \cite{cermelli2020modeling} %
  & 78.80 & \oracle{9.54}{8.52} & \oracle{9.54}{\tblbold{8.52}}   
  & 56.23 & \oracle{22.95}{19.47} & 23.59 & \oracle{65.74}{66.37} & \oracle{44.34}{42.92}   
  & 23.62 & \oracle{52.40}{58.76} & 33.24 & \oracle{51.05}{48.30} & 20.57 & \oracle{69.81}{69.84} & \oracle{57.75}{58.97} \\
  PLOP \cite{douillard2021plop}                      & 78.80 & \oracle{9.54}{8.52} & \oracle{9.54}{\tblbold{8.52}}   
                                                     & 57.05 & \oracle{21.83}{18.29} & 24.74 & \oracle{64.07}{64.74} & \oracle{42.95}{41.51}   
                                                     & 24.18 & \oracle{51.27}{57.79} & 34.23 & \oracle{49.60}{46.76} & 21.42 & \oracle{68.56}{68.59} & \oracle{56.48}{57.71} \\
  UCD \cite{yang2022uncertainty}                       & 78.80 & \oracle{9.54}{8.52} & \oracle{9.54}{\tblbold{8.52}}   
                                                       & 56.29 & \oracle{22.86}{19.38} & 26.45 & \oracle{61.58}{62.29} & \oracle{42.22}{40.84}   
                                                       & 24.88 & \oracle{49.87}{56.57} & 34.72 & \oracle{48.87}{45.99} & 22.35 & \oracle{67.21}{67.24} & \oracle{55.31}{56.60} \\
                                                       
  \name \; w/o $\mathcal{L}_{kd}^{\tilde{o}}$
                            & 78.78 & \oracle{9.56}{8.55} & \oracle{9.56}{8.55}   
                            & 59.61 & \oracle{18.32}{14.63} & 43.30 & \oracle{37.11}{38.28} & \oracle{27.71}{26.45}   
                            & 34.84 & \oracle{29.79}{39.18} & 39.11 & \oracle{42.41}{39.17} & 36.13 & \oracle{46.98}{47.03} & \oracle{39.72}{41.79}\\
  \name
                            & 78.78 & \oracle{9.56}{8.55} & \oracle{9.56}{8.55}  
                            & 59.26 & \oracle{18.80}{15.13} & 43.95 & \oracle{36.17}{37.35} & \oracle{27.48}{\tblbold{26.24}}   
                            & 37.94 & \oracle{23.54}{33.76} & 42.10 & \oracle{38.01}{34.51} & 36.60 & \oracle{46.29}{46.34} & \oracle{35.94}{\tblbold{38.21}}\\
  \midrule
  Oracle                    & \oracle{87.11}{86.14} & - & -   & \oracle{72.98}{69.82} & - & \oracle{68.85}{70.16} & - & -   & \oracle{68.14}{68.20} & - & \oracle{49.62}{57.28} & - & \oracle{67.91}{64.29} & - & -\\
  \bottomrule
 \end{tabu}
 \label{tab:ICB}
\end{table*}

In the \textbf{final learning step}, the struggle to handle the class and domain incremental training is exacerbated for all the competitors.
Baselines and MDIL still provide inferior results, with the latter performing even worse than na\"ive fine-tuning with self-stylization in some setups.
\\
As for CIL methods, PLOP \cite{douillard2021plop} and UCD \cite{yang2022uncertainty} are the best performing.
Both combines output and feature level %
objectives, which prove to be somewhat robust to domain shift.
Even so, the simpler MiB \cite{cermelli2020modeling} approach shows very competitive results, suggesting that strategies 
taking into account only a \classincr perspective
may not be so effective when incremental domain shift is also occurring.
Our method in its complete form greatly outperforms all CIL competitors by a large margin regardless of domain order, going from $5\%$ ($\text{BDD} \rightarrow \text{IDD} \rightarrow \text{CS}$) to $12\%$ ($\text{CS} \rightarrow \text{BDD} \rightarrow \text{IDD}$) and even $16\%$ ($\text{IDD} \rightarrow \text{CS} \rightarrow \text{BDD}$) in terms of $\bar\Delta_{2}$ \marco{gap}.

\tocheck{Furthermore, in Table~\ref{tab:map_gen} we investigate the generalization performance (\ie, $\Gamma^{gen}_{t}$ from Eq.~\eqref{eq:gen_metric}) achieved by 
 the considered methods.
To do so, we compute the  accuracy at each incremental step on the \textit{unseen} Mapillary dataset for the sets of classes observed so far.
We notice that simple fine-tuning and MDIL offer poor generalization results, which is expected due to the low accuracy they already provide on datasets directly observed. %
On the other hand, CIL methods reach more competitive results, 
even if none of them proves to be superior in all setups.
Still, our approach outperforms all competitors, getting significantly closer to the \textit{Oracle} upper-bound (\ie, the supervised training on the entire Mapillary), specially in the $\text{IDD} \rightarrow\allowbreak \text{CS} \rightarrow\allowbreak \text{BDD}$ setup.
Also, we remark how we get similar generalization results with different domain incremental orders, demonstrating how our approach is able to learn and preserve generalizable task-related clues regardless of the training environment.}

\begin{table}[t!]
 \scriptsize
 \centering
  \setlength{\tabcolsep}{2.6pt}
 \caption{Generalization performance \marco{($\Gamma^{gen}_{t}$)} as mIoU computed on Mapillary's test set ($\mathcal{C}_{bgr} \protect\rightarrow \mathcal{C}_{stat} \protect\rightarrow \mathcal{C}_{mov}$ setup).}
  \begin{tabu}{l |ccc|ccc|ccc}
  \toprule
  & \multicolumn{3}{c}{$\text{CS} \veryshortarrow \text{BDD} \veryshortarrow \text{IDD}$} 
  & \multicolumn{3}{c}{$\text{BDD} \veryshortarrow \text{IDD} \veryshortarrow \text{CS}$} 
  & \multicolumn{3}{c}{$\text{IDD} \veryshortarrow \text{CS} \veryshortarrow \text{BDD}$}\\
 & Step 0 & Step 1 & Step 2  
 & Step 0 & Step 1 & Step 2 
 & Step 0 & Step 1 & Step 2 \\
  \midrule
  FT ($\mathcal{L}_{ce}^{n}$)         & 36.27 & 22.03 & 13.71   & 66.74 & 25.27 &  6.60   & \tblbold{59.56} & 7.81 & 8.52 \\
  FT\textsuperscript{\textdagger} ($\mathcal{L}_{ce}^{\tilde{n}}$)     & \tblbold{58.09} & 19.83 & 14.99   & \tblbold{66.83} & 25.77 & 16.34   & 59.19 & 23.40 & 11.97 \\
  MDIL               & 44.60 & 24.77 & 16.05   & 66.36 & 18.40 & 11.01   & 56.41 & 14.86 &  8.55 \\

  ILT & 36.27 & 26.80 & 20.69   & 66.74 & 41.32 & 28.97   & \tblbold{59.56} & 39.27 & 27.96 \\

  MiB %
  & 36.27 & 37.68 & 32.36   & 66.74 & 45.99 & 33.01   & \tblbold{59.56} & 23.61 & 24.23 \\
  PLOP                                   & 36.27 & 39.62 & 33.69   & 66.74 & 45.45 & 34.01   & \tblbold{59.56} & 25.29 & 25.04 \\
  UCD                                    & 36.27 & 38.46 & 34.07   & 66.74 & \tblbold{46.22} & 29.92   & \tblbold{59.56} & 27.08 & 25.89 \\
  \name                                    & \tblbold{58.09} & \tblbold{46.36} & \tblbold{40.43}   & \tblbold{66.83} & 44.99 & \tblbold{37.33}   & 59.19 & \tblbold{43.15} & \tblbold{39.16} \\
  \midrule
  Oracle                                  & 83.96 & 73.77 & 65.42   & 83.96 & 73.77 & 65.42   & 83.96 & 73.77 & 65.42 \\ 
  \bottomrule
 \end{tabu}
 \\
 \vspace{0.7mm}
 \textsuperscript{\textdagger} indicates the presence of self-stylization.
 \label{tab:map_gen}
 \vspace{-0mm}
\end{table}

\begin{table}[t!]
 \scriptsize
 \centering
  \setlength{\tabcolsep}{3.pt}
  \renewcommand{\arraystretch}{1.}
 \caption{Experimental results on $\text{CS} \protect\rightarrow  \text{BDD} \protect\rightarrow  \text{IDD}$ domain setup and $\boldsymbol{\mathcal{C}_{bgr} \protect\rightarrow  \mathcal{C}_{mov} \protect\rightarrow  \mathcal{C}_{stat}}$ class setup.}

  \begin{tabu*}{
  cccccccccc}
  \toprule
  \multicolumn{3}{c}{$\text{CS} \veryshortarrow \text{BDD} \veryshortarrow \text{IDD}$}  & \multicolumn{7}{c}{Method} \\
  \multicolumn{3}{c}{$\mathcal{C}_{bgr} \veryshortarrow \mathcal{C}_{stat} \veryshortarrow \mathcal{C}_{mov}$} 
  & $\mathcal{L}_{\scaleto{ce\mathstrut}{4pt}}^{\scaleto{n\mathstrut}{4pt}}$ 
  & $\mathcal{L}_{\scaleto{ce\mathstrut}{4pt}}^{\scaleto{\tilde{n}\mathstrut}{4pt}}$ 
  & MiB %
  & PLOP
  & UCD
  & \name 
  & Oracle\\
  \midrule
  \multirow{2}{*}{Step 0} & \multirow{1}{*}{mIoU\textsubscript{0} $ \! \uparrow$} & \textbf{CS}  & 79.83 & 79.39 & 79.83 & 79.83 & 79.83 & 79.39 & \oracle{85.15}{84.15} \\
  
                          \arrayrulecolor{black!50}
                         \cmidrule{2-10}
                         \arrayrulecolor{black!100}
                         
                          & \multicolumn{2}{c}{$\bar{\Delta}_{0} \! \downarrow$}               & \oracle{6.25}{\tblbold{5.13}} & \oracle{6.77}{5.65} & \oracle{6.25}{\tblbold{5.13}} & \oracle{6.25}{\tblbold{5.13}} & \oracle{6.25}{\tblbold{5.13}} & \oracle{6.77}{5.65} & - \\
 \midrule                          
 \multirow{3}{*}{Step 1} & \multirow{2}{*}{mIoU\textsubscript{1} $ \! \uparrow$} & \textbf{BDD}   & 15.79 & 19.43 & 26.15 & 27.69 & 23.73 & 40.92 & \oracle{49.39}{63.08} \\
                         & & CS  & 14.26 & 17.22 & 40.38 & 42.44 & 39.76 & 49.70 & \oracle{70.26}{69.82} \\
                         
                       \arrayrulecolor{black!50}
                        \cmidrule{2-10}
                        \arrayrulecolor{black!100}
                         
                         & \multicolumn{2}{c}{$\bar{\Delta}_{1} \! \downarrow$}      & \oracle{73.87}{76.26} & \oracle{68.08}{71.03} & \oracle{44.79}{48.21} & \oracle{41.77}{45.40} & \oracle{47.68}{50.68} & \oracle{23.21}{\tblbold{28.99}} & - \\
 \midrule 
 \multirow{4}{*}{Step 2} & \multirow{3}{*}{mIoU\textsubscript{2} $ \! \uparrow$} & \textbf{IDD}  & 13.47 & 14.82 & 31.01 & 31.89 & 30.72 & 43.54 & \oracle{68.14}{68.20} \\
                         & & BDD  & 6.94 & 8.21 & 23.40 & 25.21 & 22.83 & 32.34 & \oracle{49.62}{57.28} \\
                         & & CS  & 7.45 & 10.49 & 33.60 & 33.81 & 32.85 & 39.76 & \oracle{67.91}{64.29} \\
                         
                       \arrayrulecolor{black!50}
                        \cmidrule{2-10}
                        \arrayrulecolor{black!100}
                         
                         & \multicolumn{2}{c}{$\bar{\Delta}_{2} \! \downarrow$}      & \oracle{85.09}{85.52} & \oracle{82.09}{82.54} & \oracle{52.62}{53.81} & \oracle{50.87}{52.21} & \oracle{53.51}{54.67} & \oracle{37.46}{\tblbold{39.29}} & - \\

  \bottomrule
 \end{tabu*}
 \label{tab:CBI_rCIL}
\end{table}

\begin{table*}[h!]
  \centering
  \setlength{\tabcolsep}{1.2pt}
  \renewcommand{\arraystretch}{0.2}
  \small
  \begin{minipage}{0.99\linewidth}\centering
  \begin{tabular}{ cccccccc }
  & Input & GT & FT w/ self-style& MiB & PLOP & \name & \\[-2ex]

  \raisebox{1.8\normalbaselineskip}[0pt][0pt]{\rotatebox[origin=c]{90}{Step 0}} &
  
  \subfloat{\includegraphics[width=0.15\textwidth]{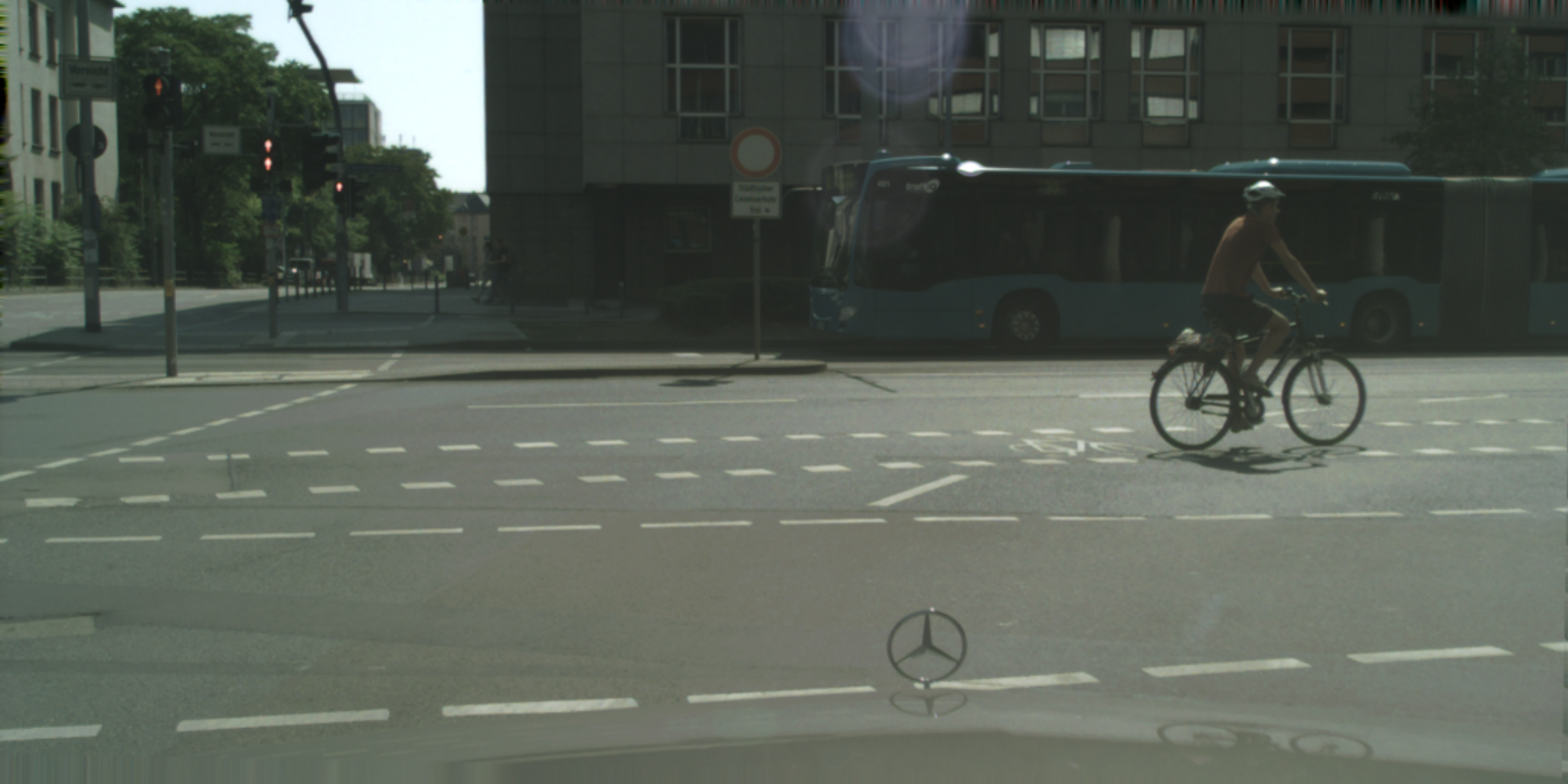}} &
  \subfloat{\includegraphics[width=0.15\textwidth]{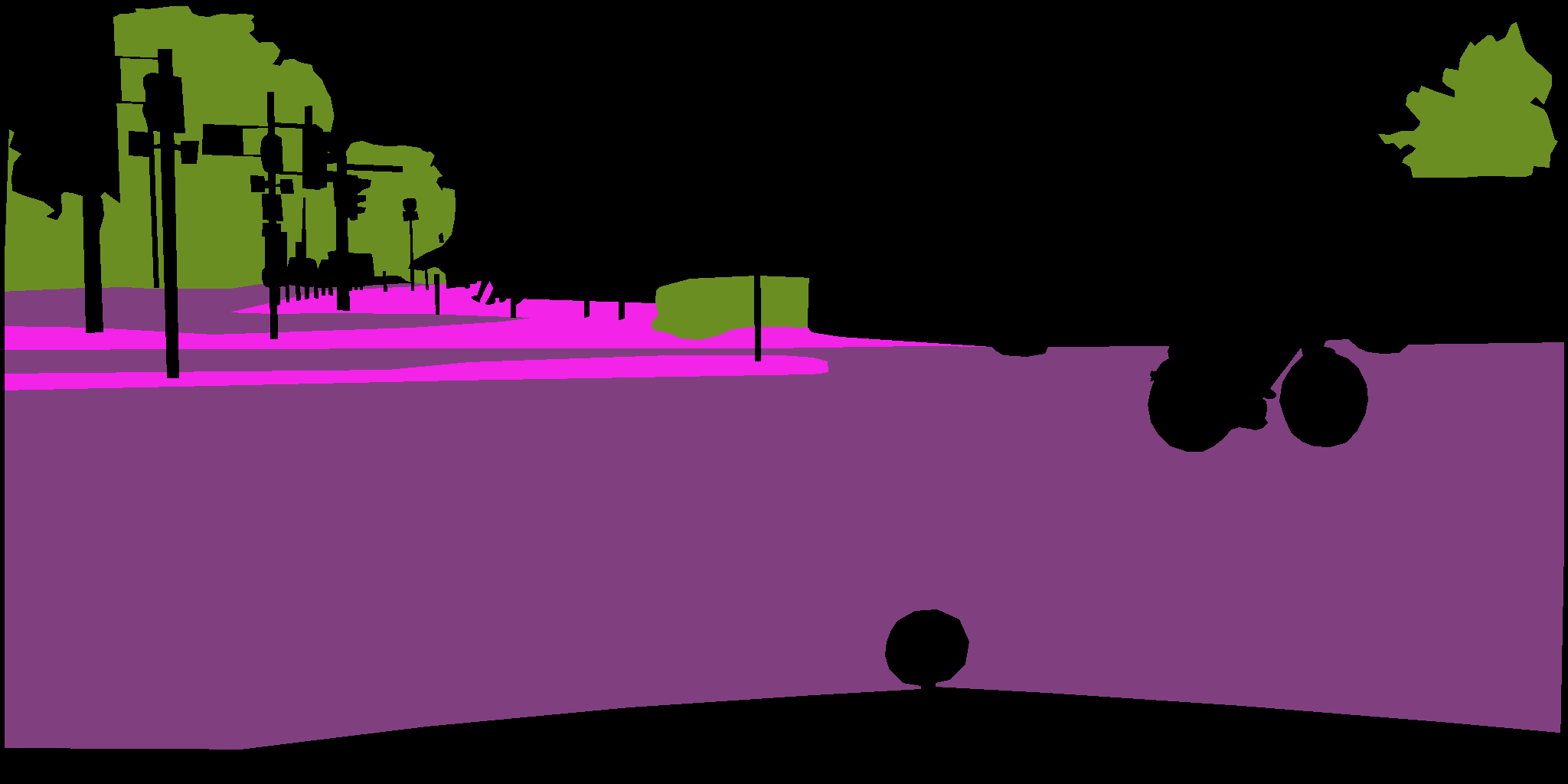}} &
  \subfloat{\includegraphics[width=0.15\textwidth]{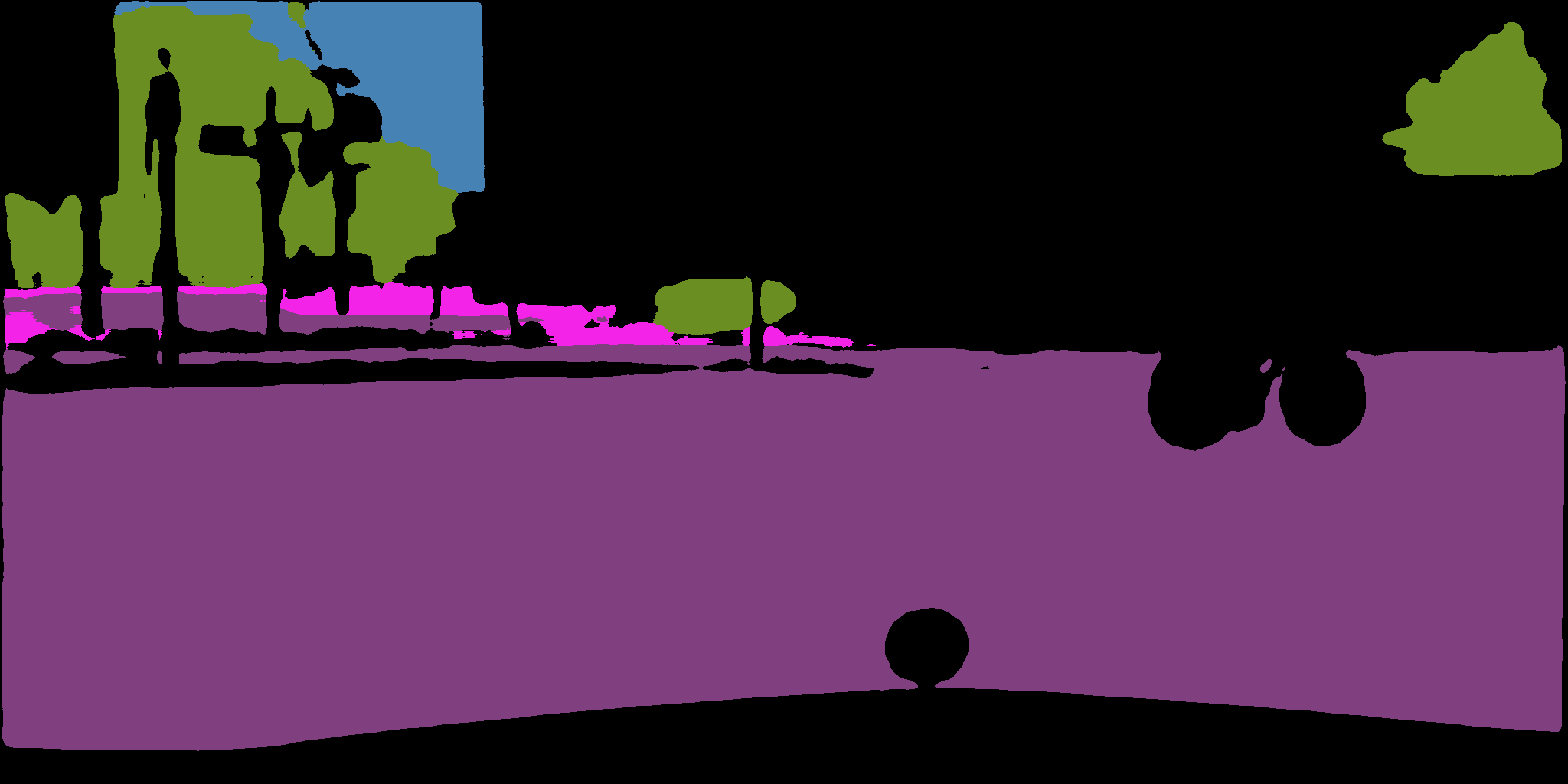}} &
  \subfloat{\includegraphics[width=0.15\textwidth]{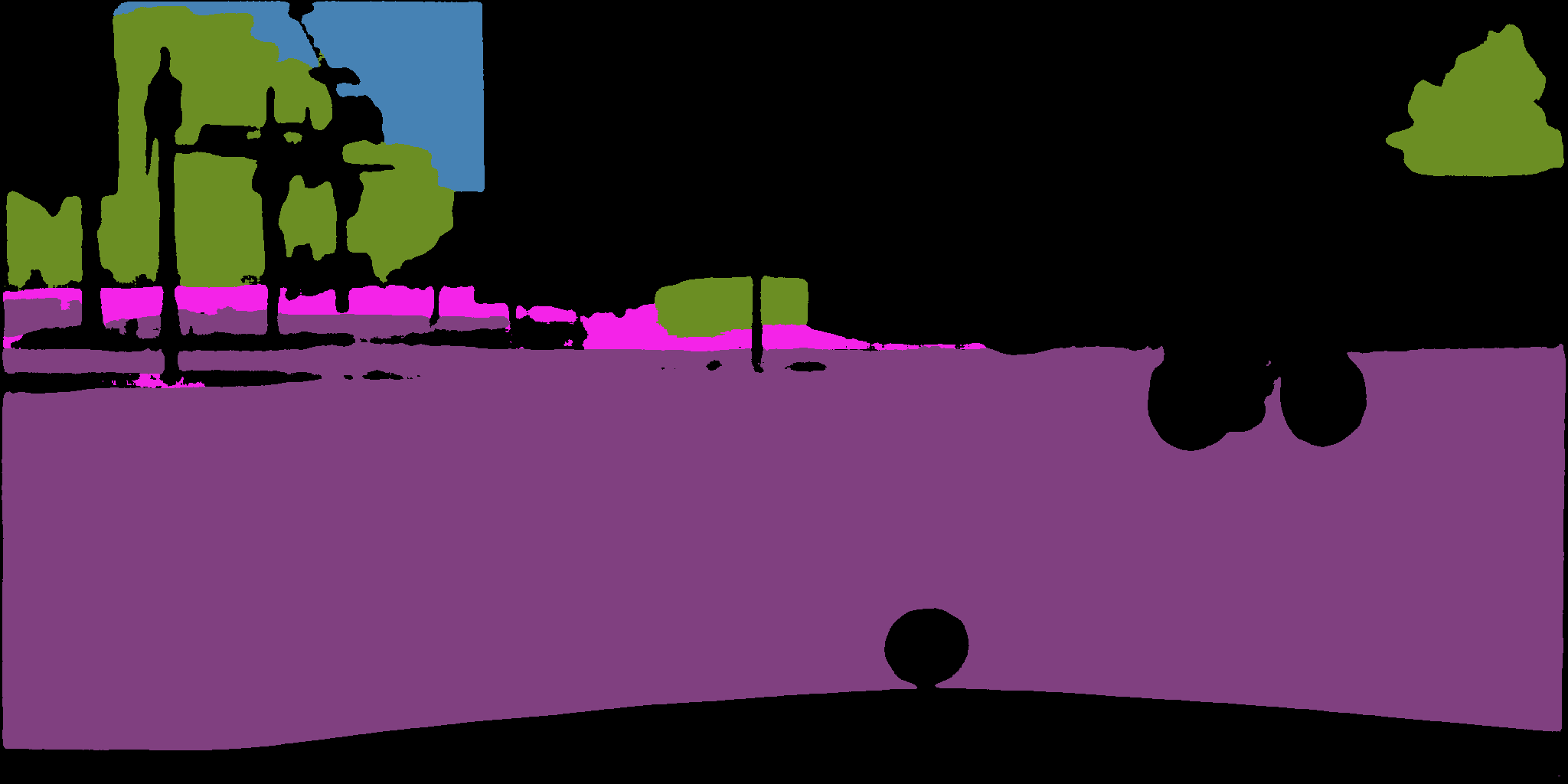}} &
  \subfloat{\includegraphics[width=0.15\textwidth]{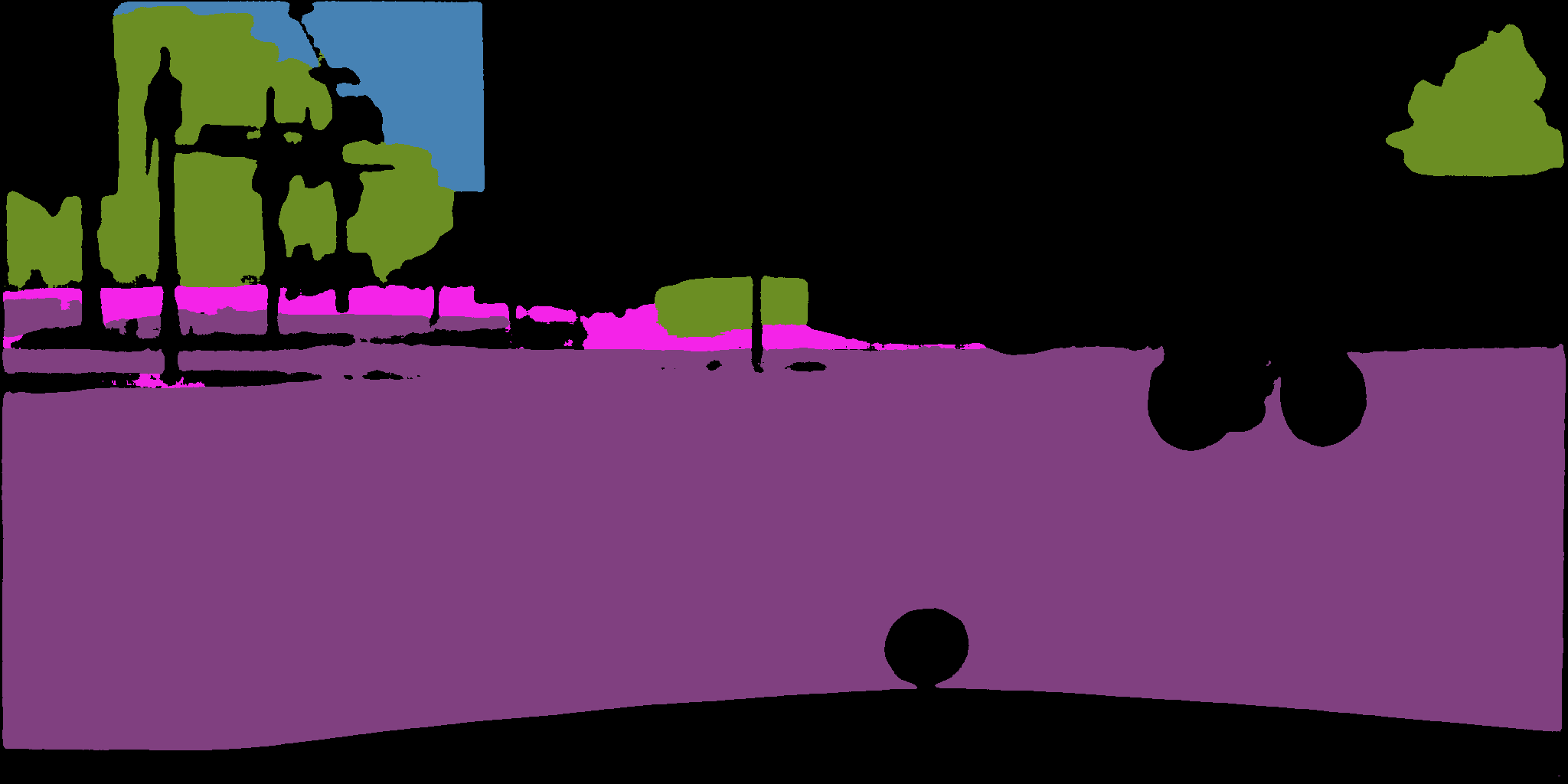}} &
  \subfloat{\includegraphics[width=0.15\textwidth]{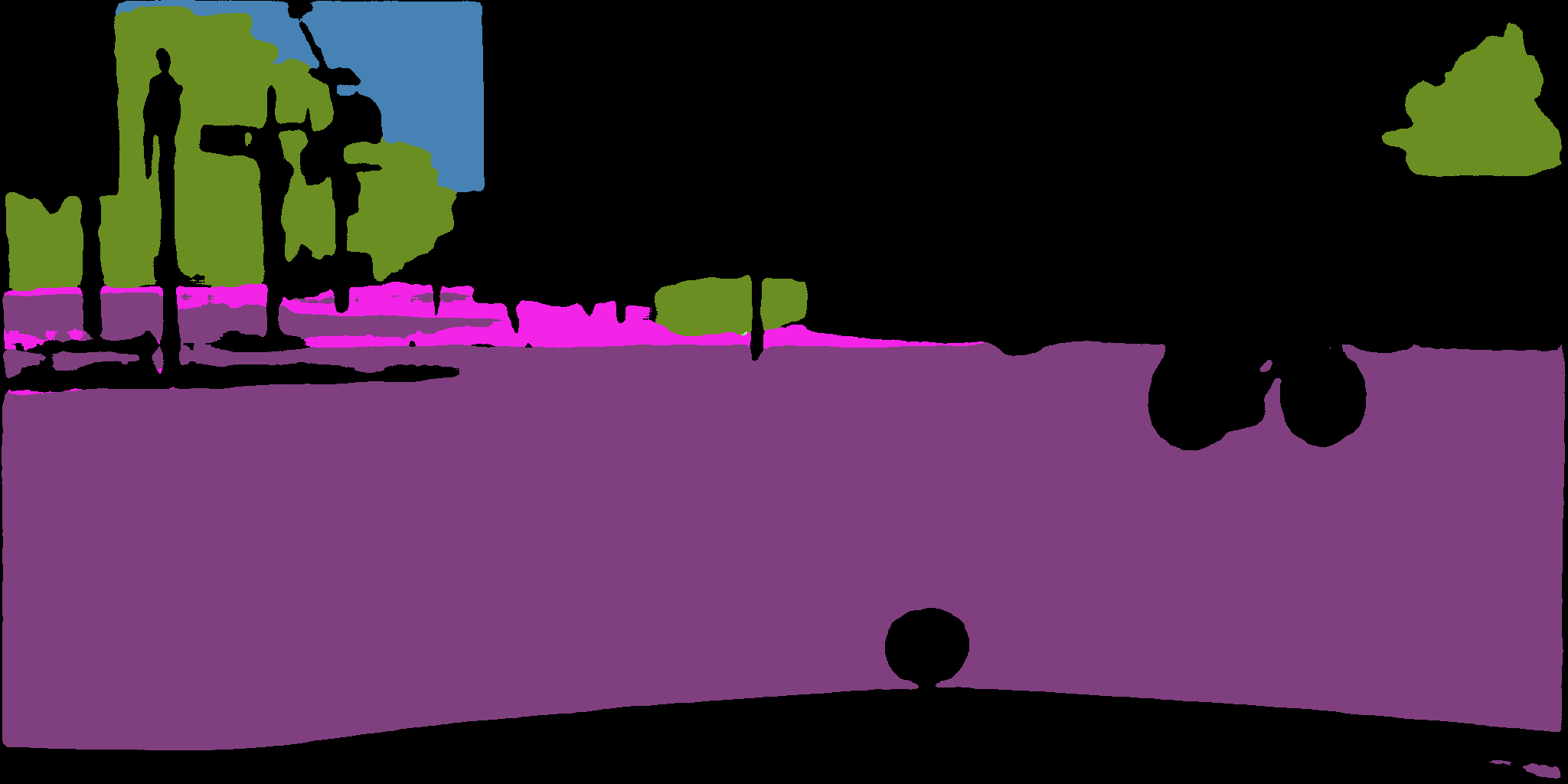}} &
  \raisebox{1.8\normalbaselineskip}[0pt][0pt]{\rotatebox[origin=c]{270}{\textbf{CS}}} \\[0.5ex]

  \hline & \\[-2.25ex]

  \raisebox{0\normalbaselineskip}[0pt][0pt]{\rotatebox[origin=c]{90}{Step 1}} &

  \subfloat{\includegraphics[width=0.15\textwidth]{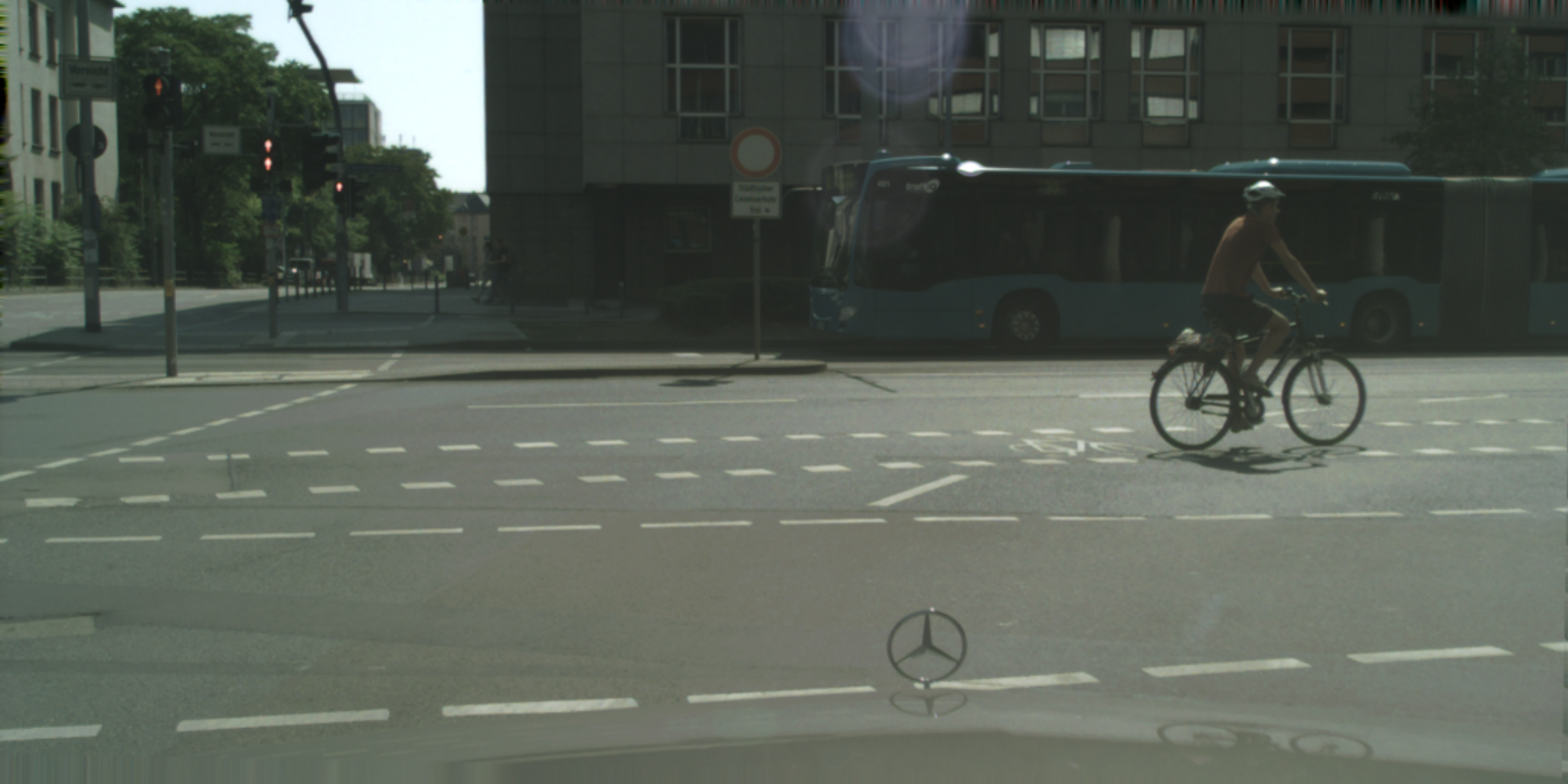}} &
  \subfloat{\includegraphics[width=0.15\textwidth]{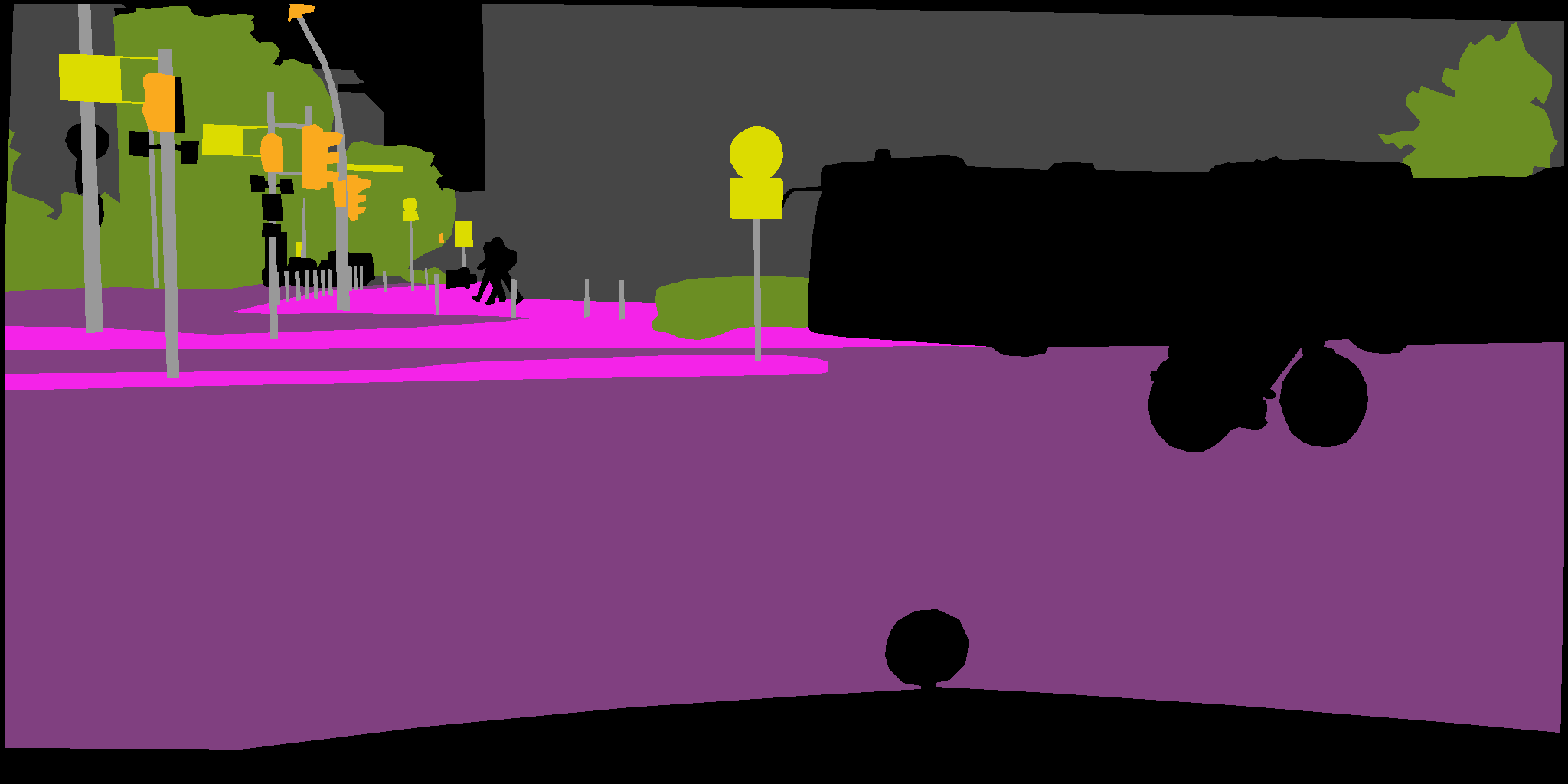}} &
  \subfloat{\includegraphics[width=0.15\textwidth]{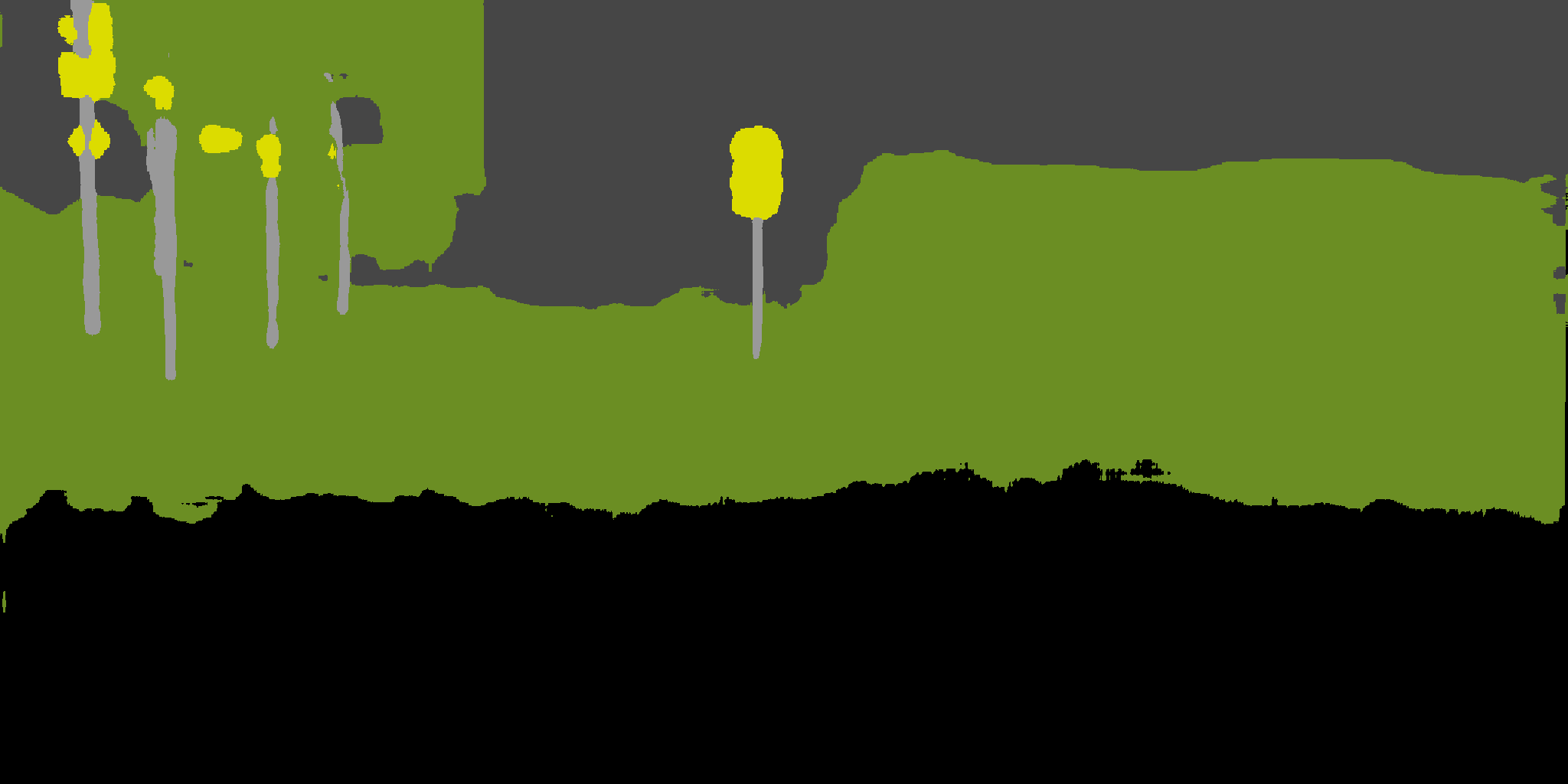}} &
  \subfloat{\includegraphics[width=0.15\textwidth]{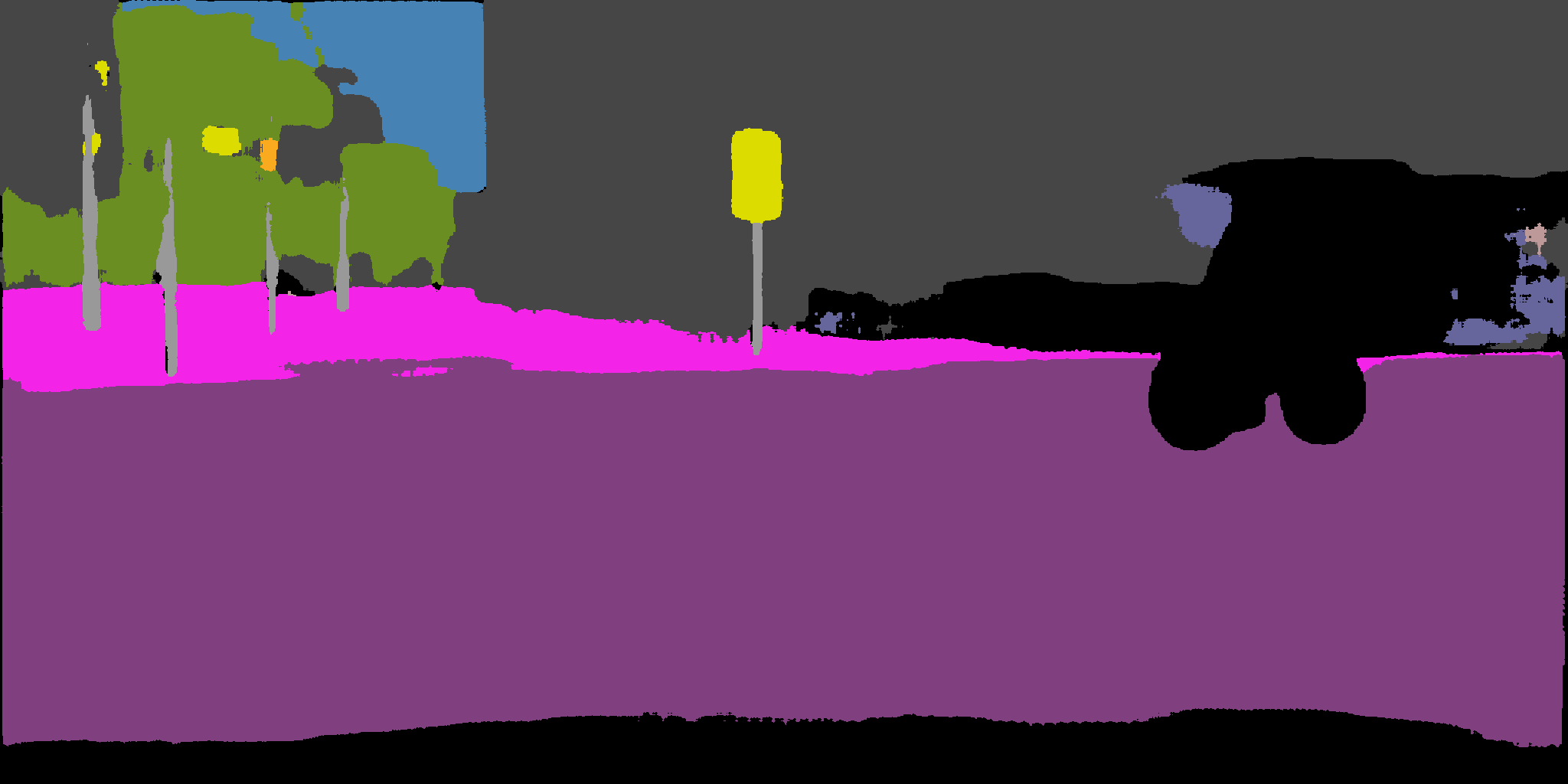}} &
  \subfloat{\includegraphics[width=0.15\textwidth]{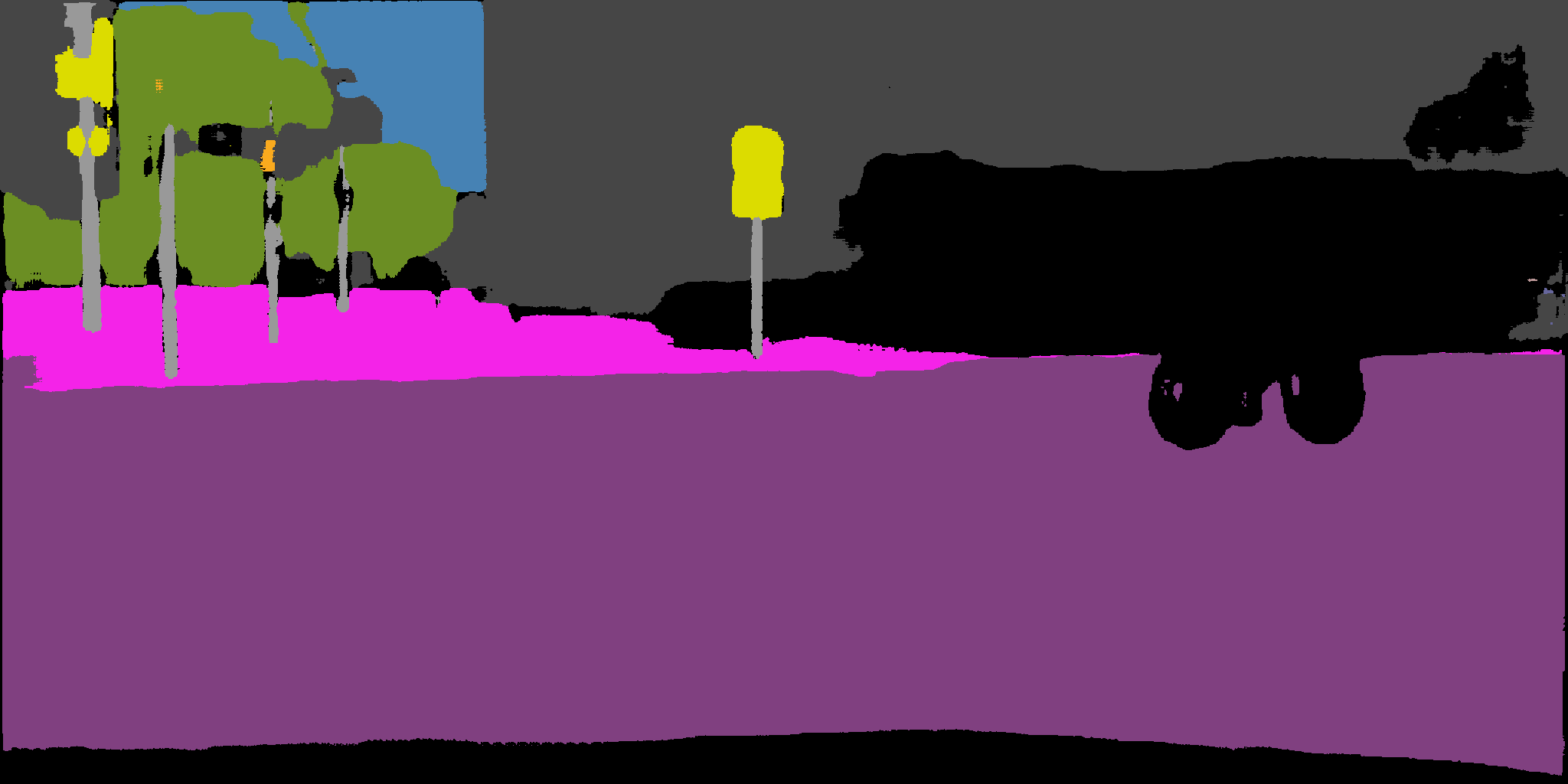}} &
  \subfloat{\includegraphics[width=0.15\textwidth]{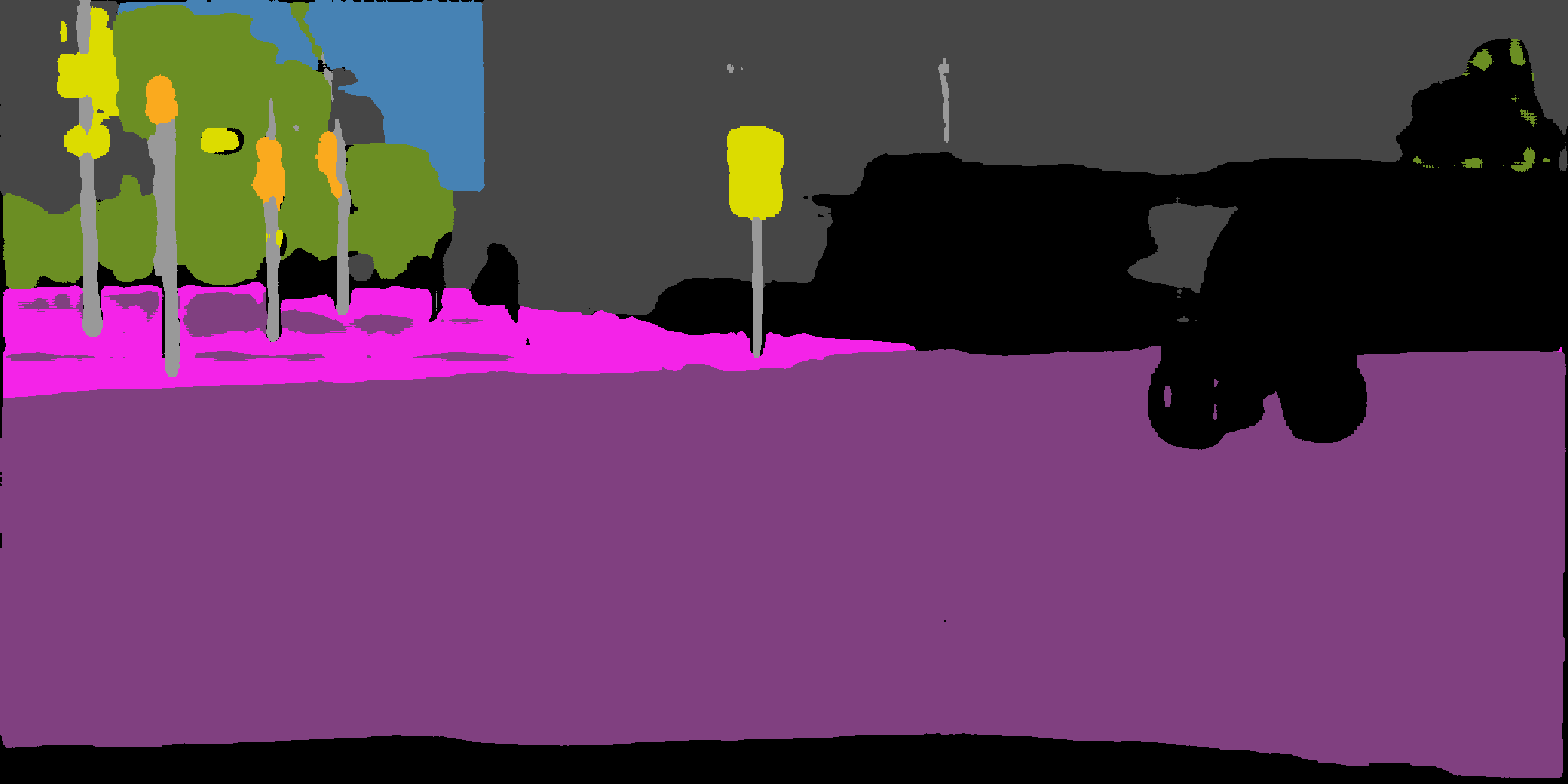}} &
  \raisebox{1.8\normalbaselineskip}[0pt][0pt]{\rotatebox[origin=c]{270}{CS}} \\[-2ex]

  &
  
  \subfloat{\includegraphics[width=0.15\textwidth]{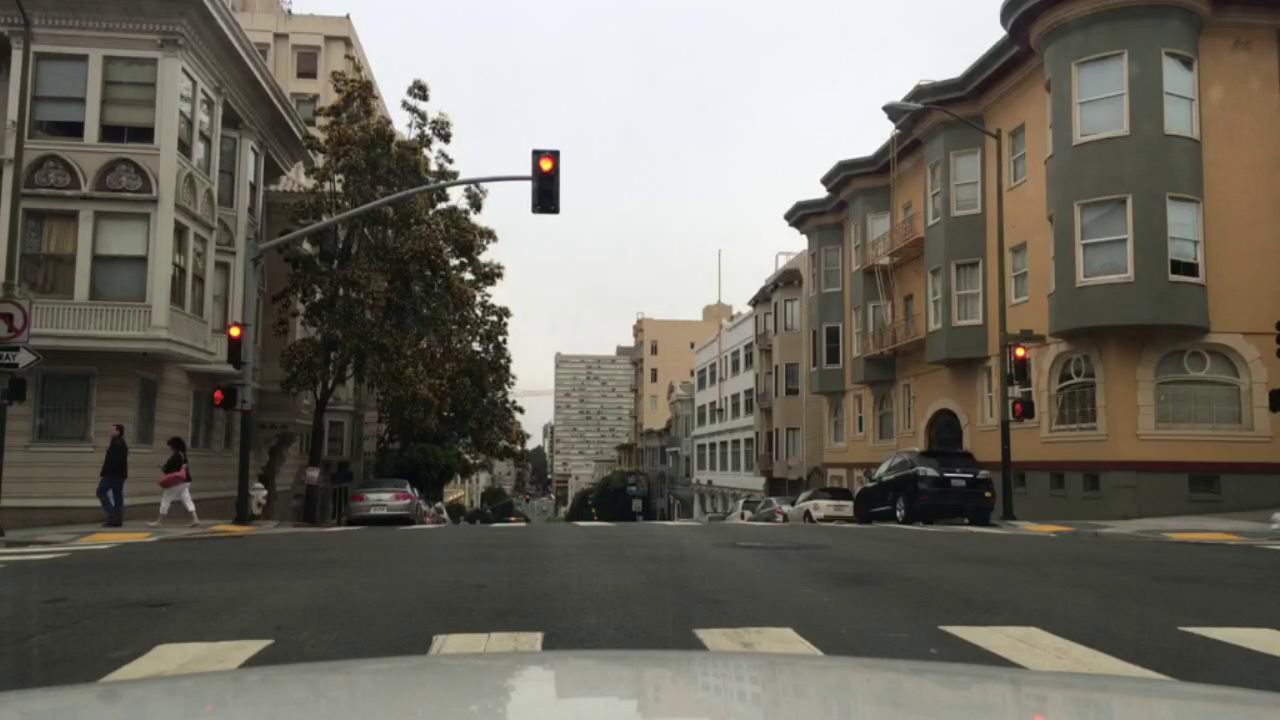}} &
  \subfloat{\includegraphics[width=0.15\textwidth]{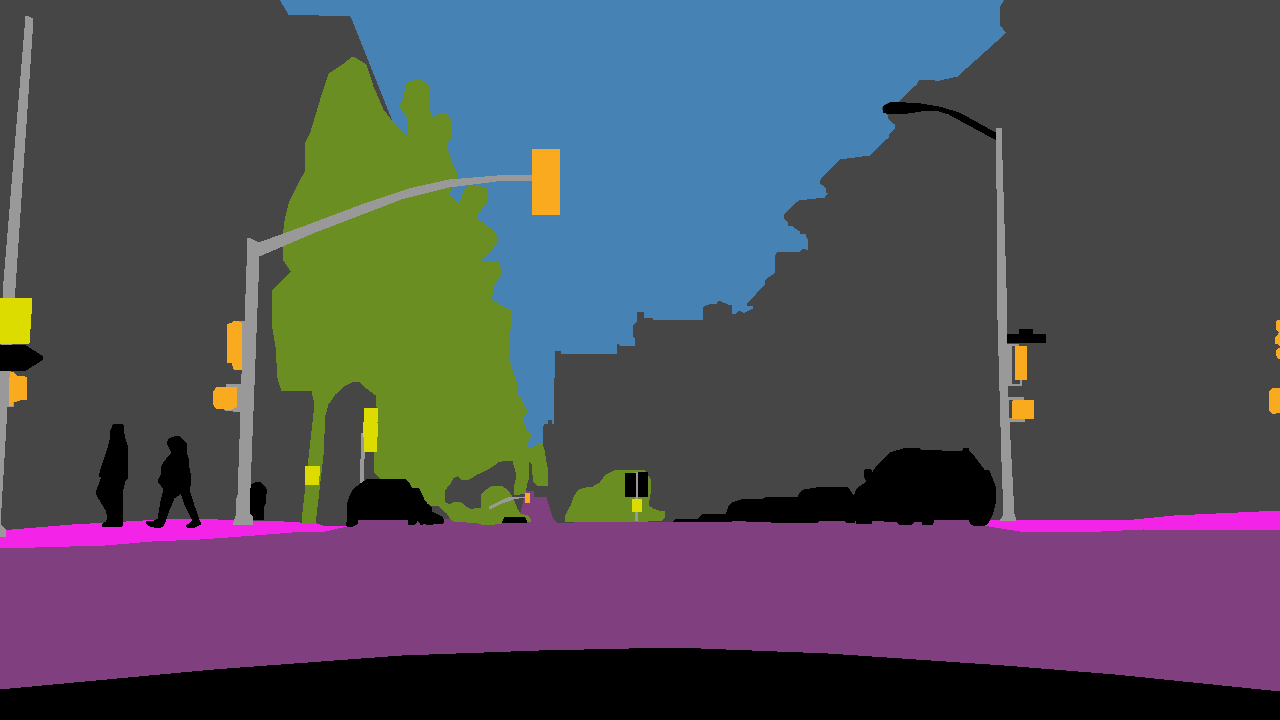}} &
  \subfloat{\includegraphics[width=0.15\textwidth]{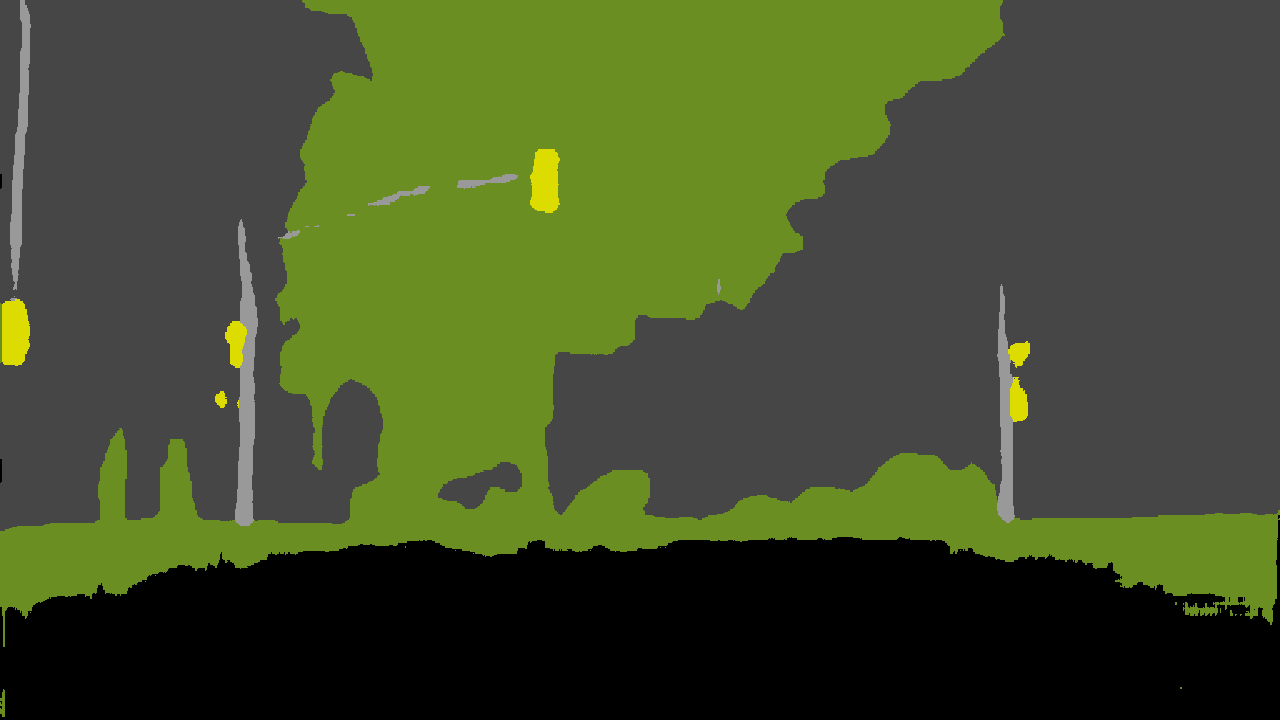}} &
  \subfloat{\includegraphics[width=0.15\textwidth]{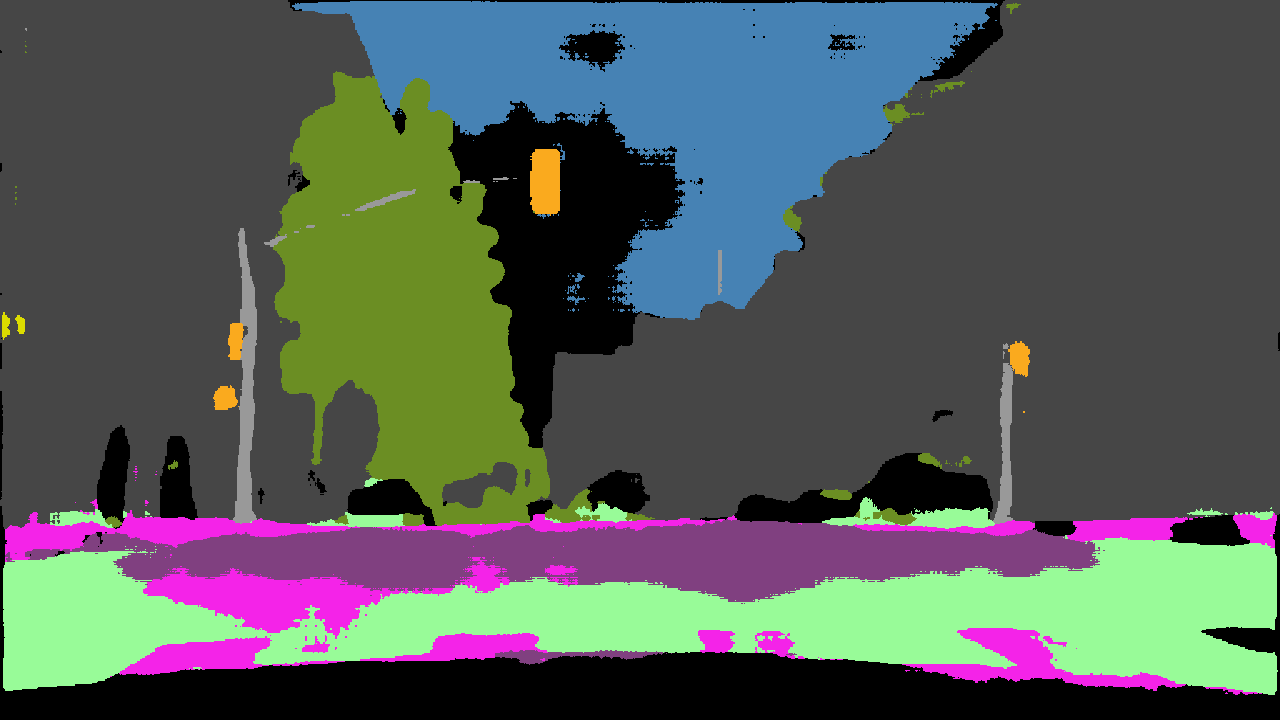}} &
  \subfloat{\includegraphics[width=0.15\textwidth]{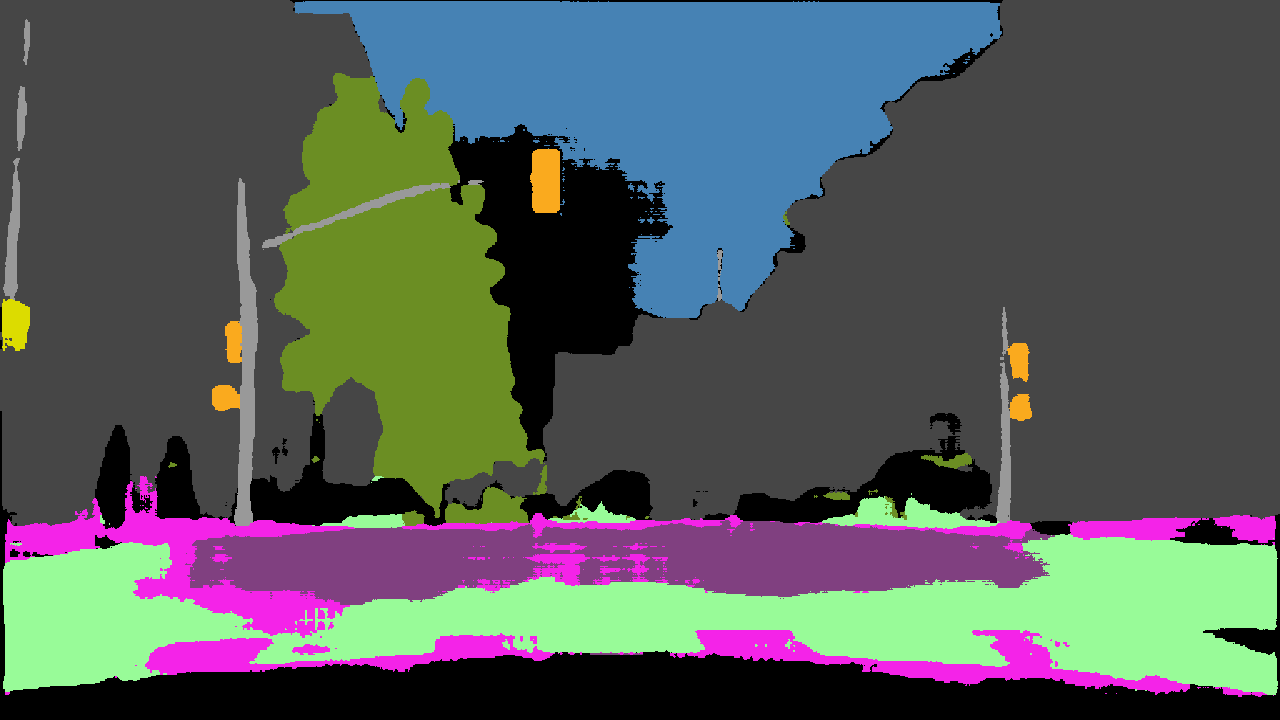}} &
  \subfloat{\includegraphics[width=0.15\textwidth]{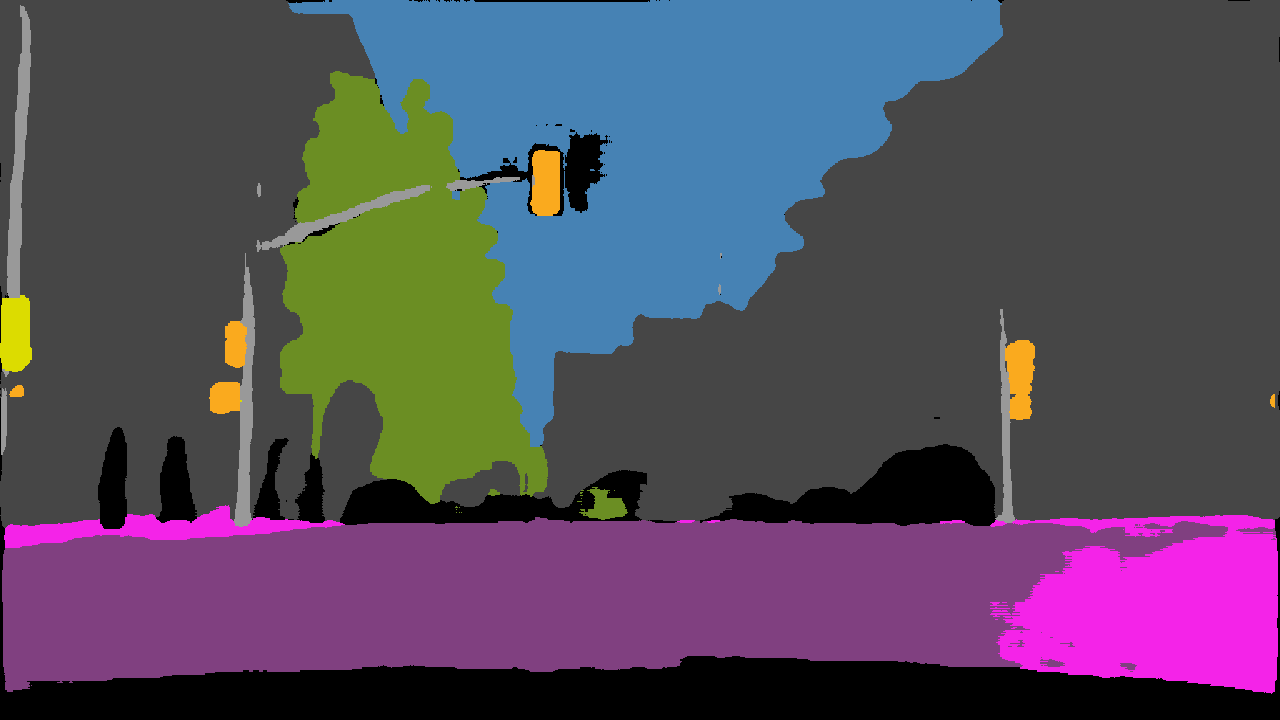}} &
  \raisebox{1.8\normalbaselineskip}[0pt][0pt]{\rotatebox[origin=c]{270}{\textbf{BDD}}} \\[0.5ex]

  \hline & \\[-2.25ex]

  \raisebox{-3\normalbaselineskip}[0pt][0pt]{\rotatebox[origin=c]{90}{Step 2}} &
  
  \subfloat{\includegraphics[width=0.15\textwidth]{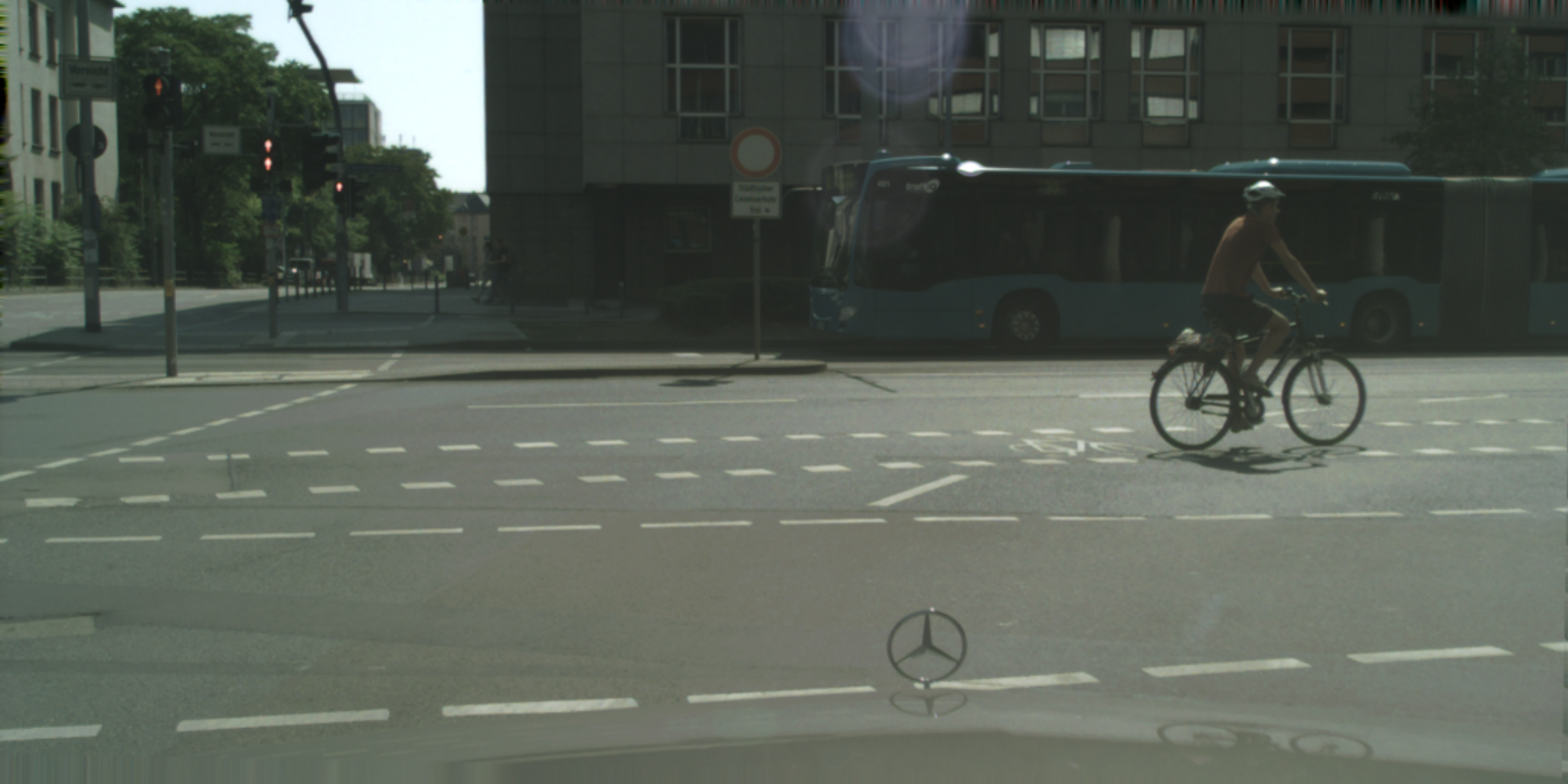}} &
  \subfloat{\includegraphics[width=0.15\textwidth]{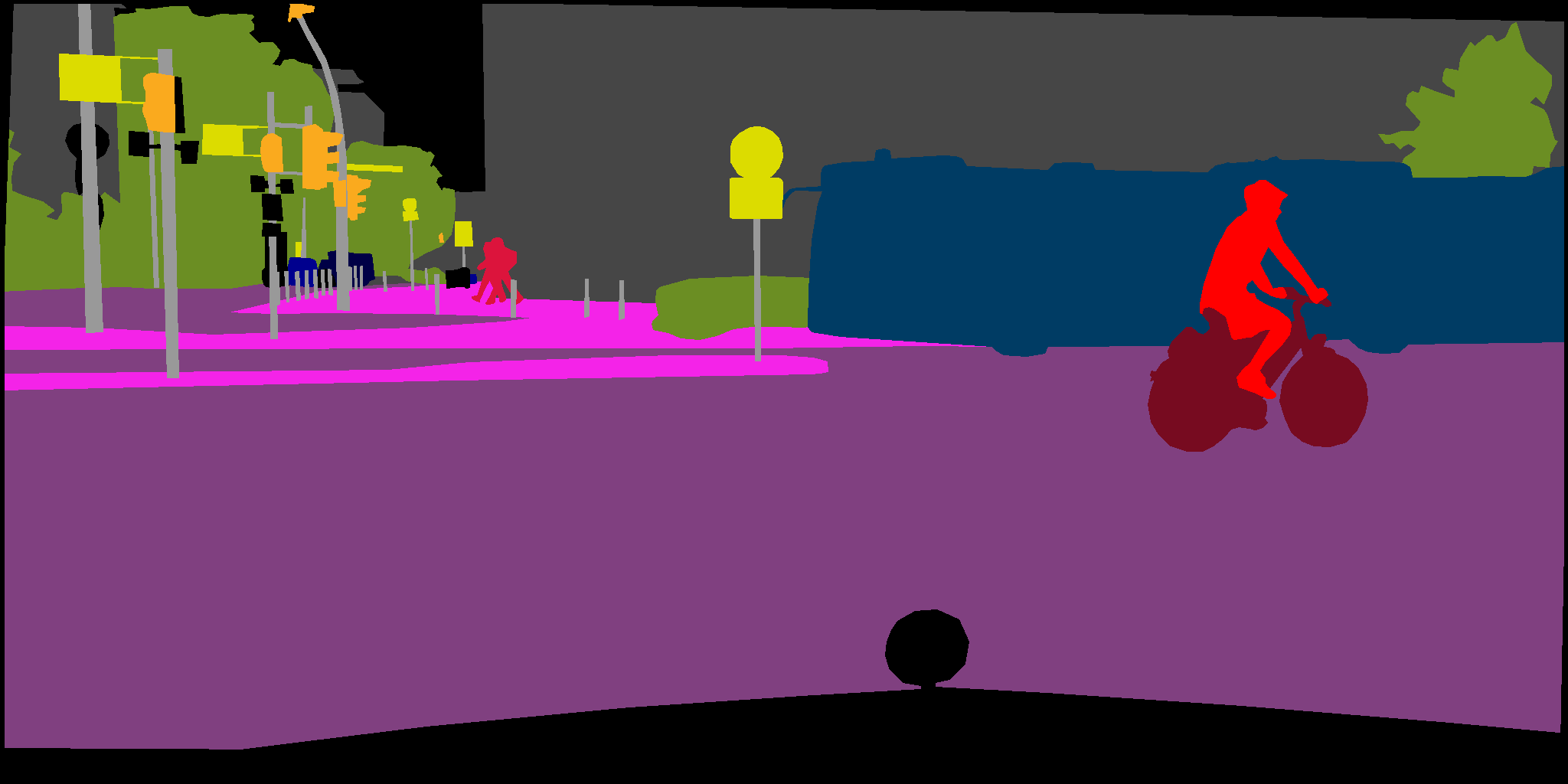}} &
  \subfloat{\includegraphics[width=0.15\textwidth]{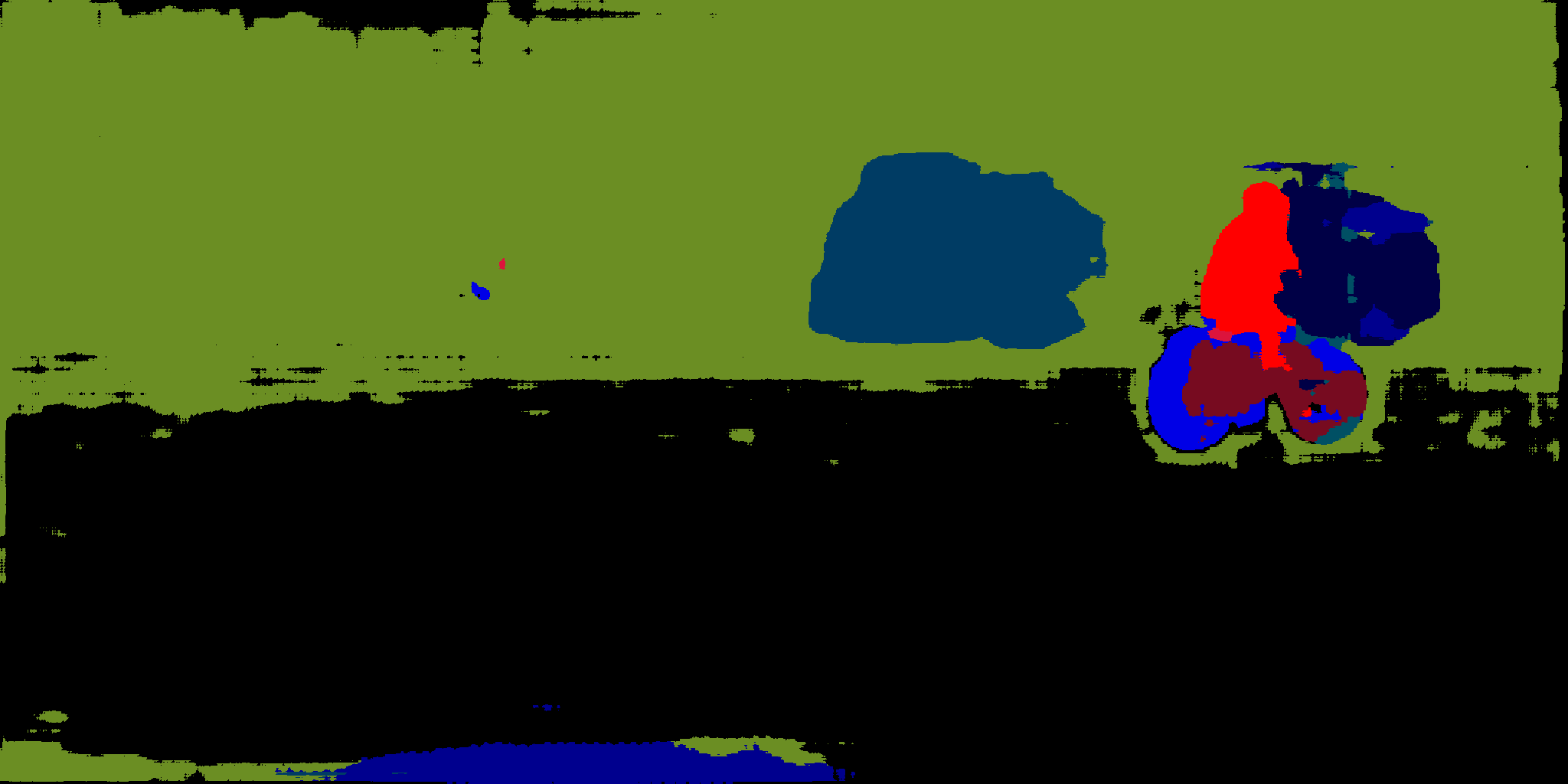}} &
  \subfloat{\includegraphics[width=0.15\textwidth]{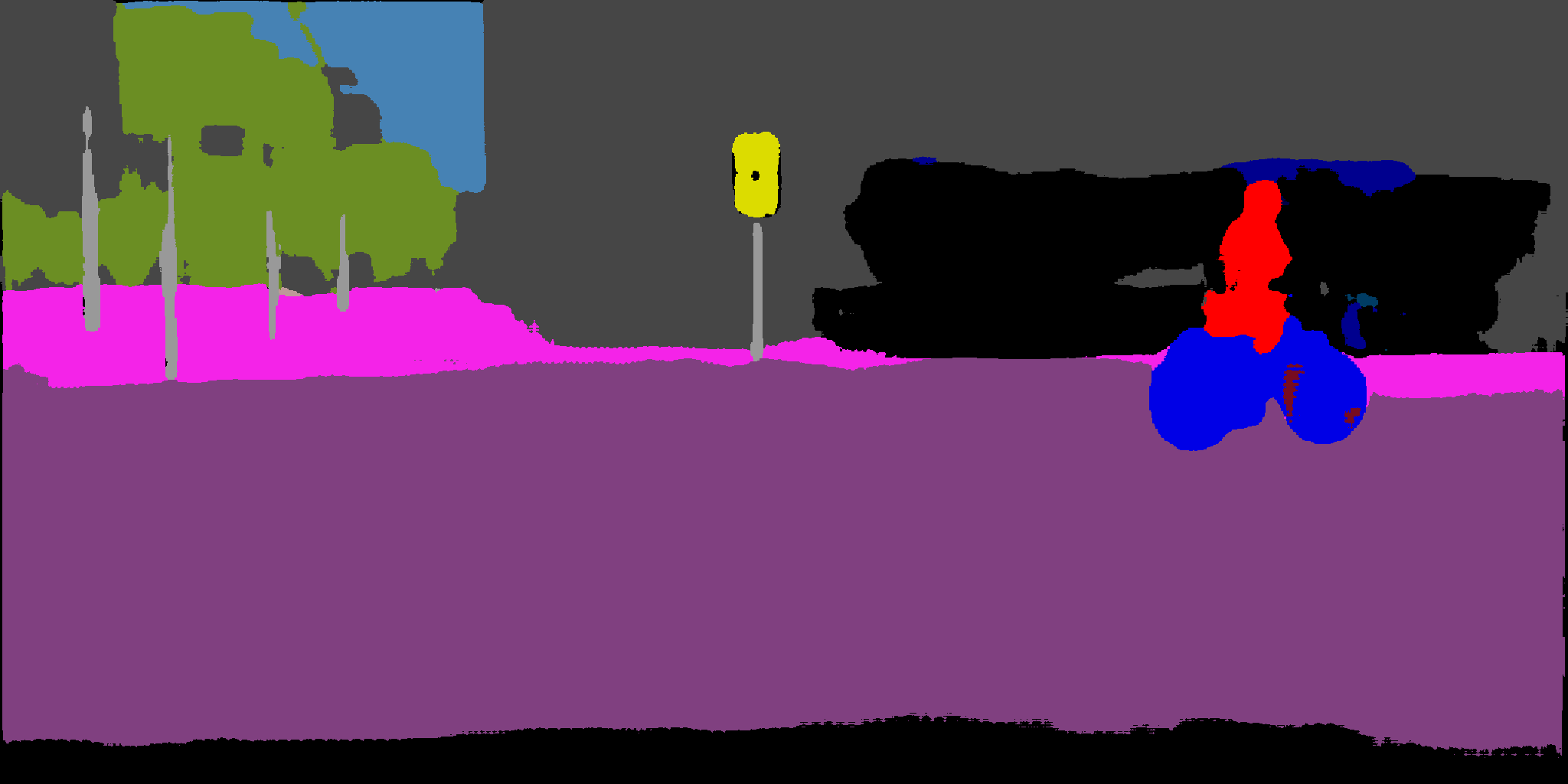}} &
  \subfloat{\includegraphics[width=0.15\textwidth]{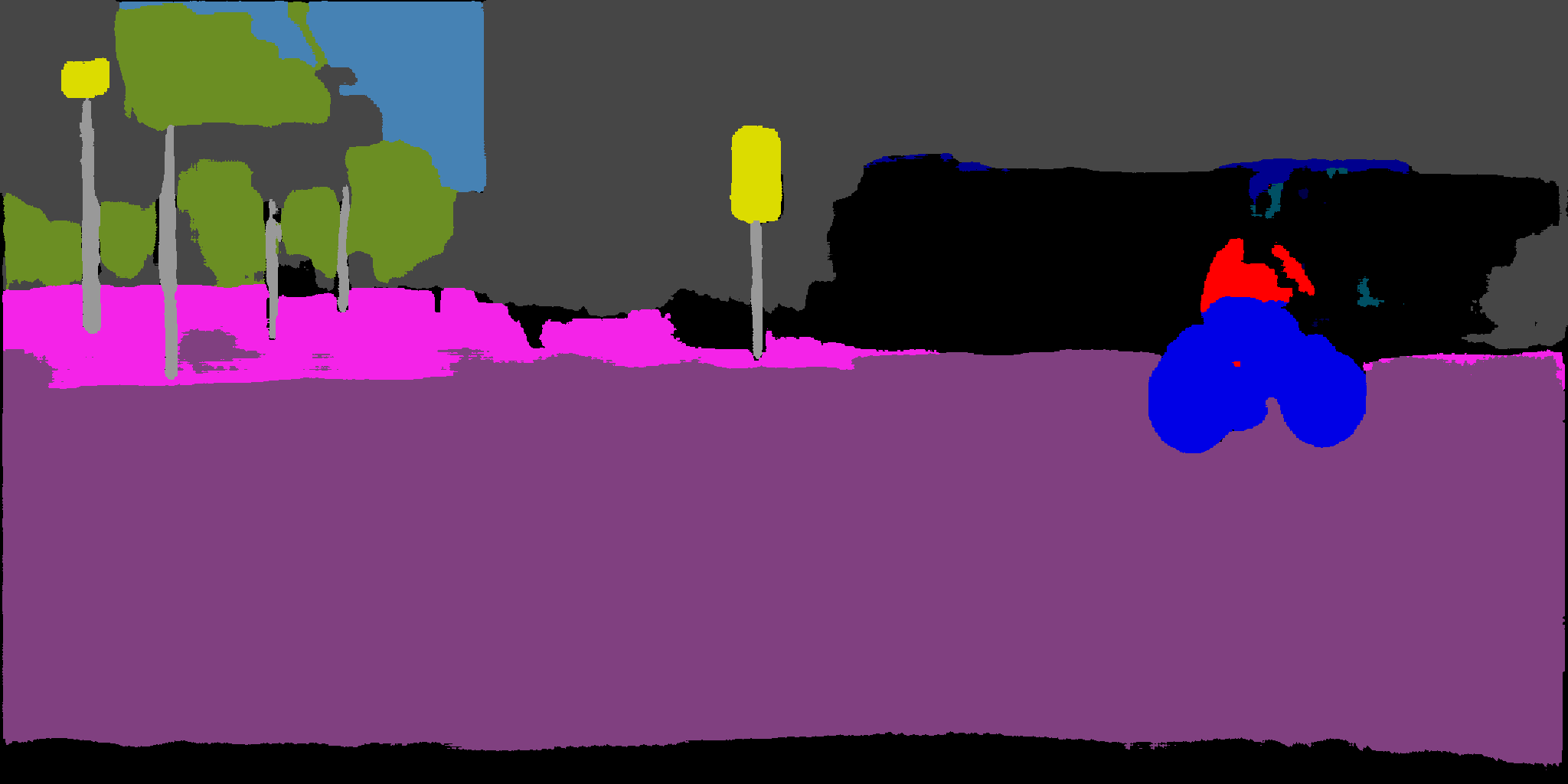}} &
  \subfloat{\includegraphics[width=0.15\textwidth]{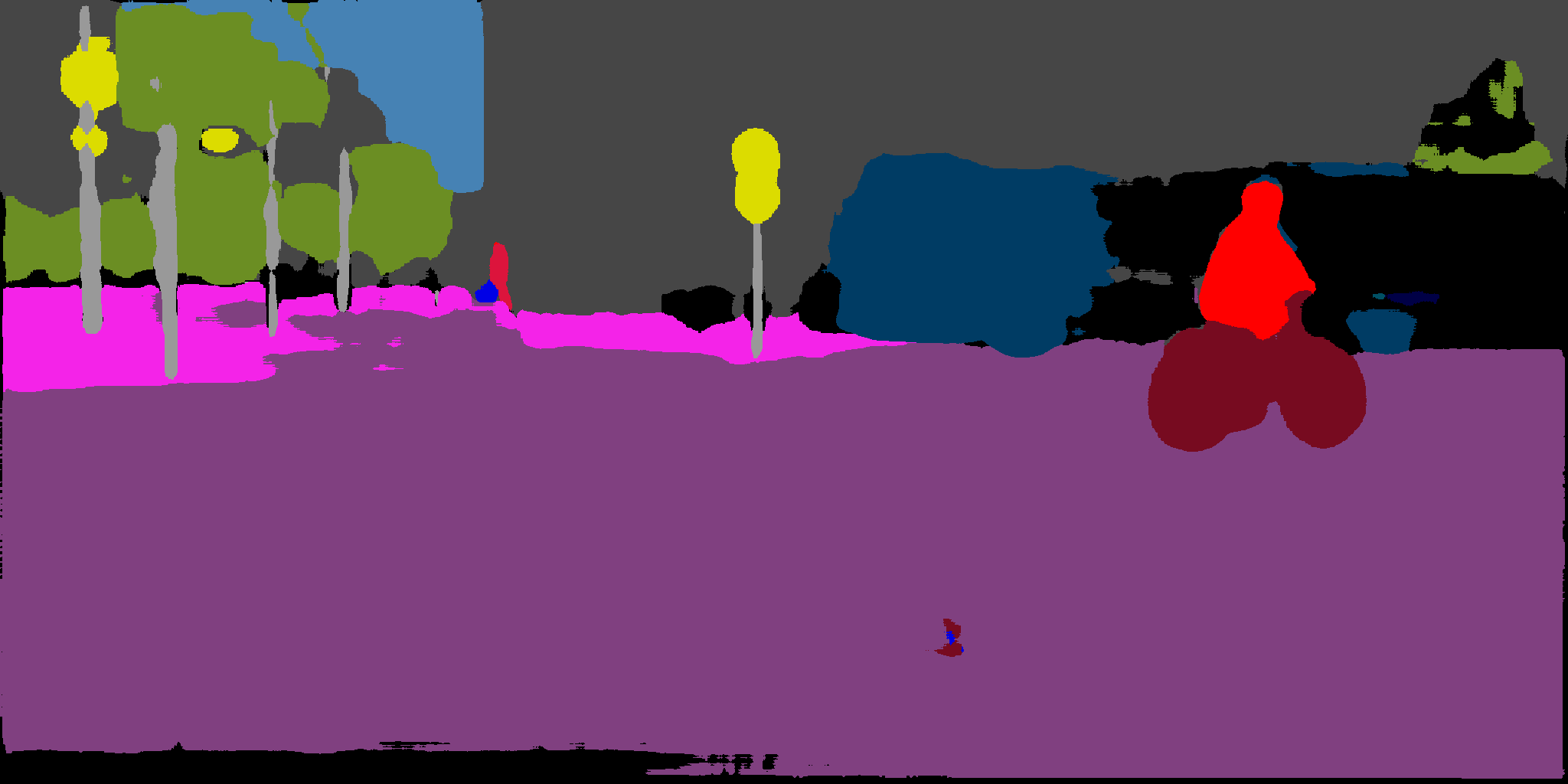}} &
  \raisebox{1.8\normalbaselineskip}[0pt][0pt]{\rotatebox[origin=c]{270}{CS}} \\[-2ex]
  
&
  
  \subfloat{\includegraphics[width=0.15\textwidth]{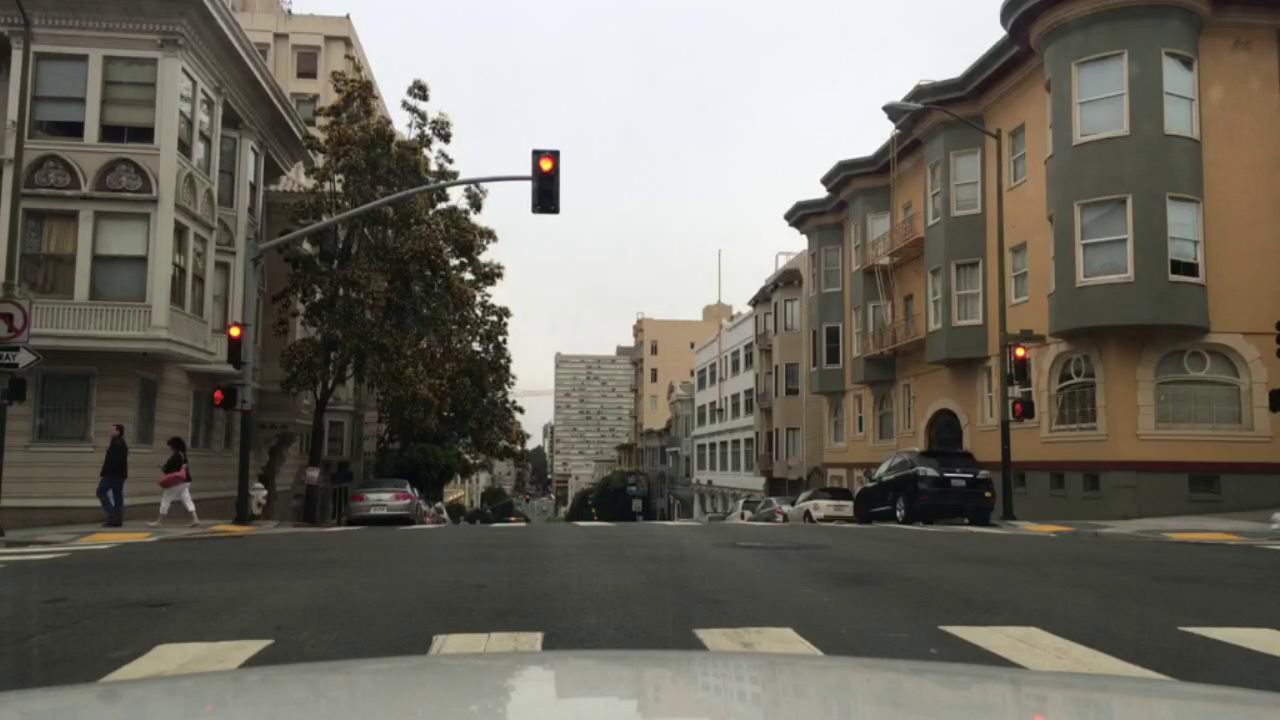}} &
  \subfloat{\includegraphics[width=0.15\textwidth]{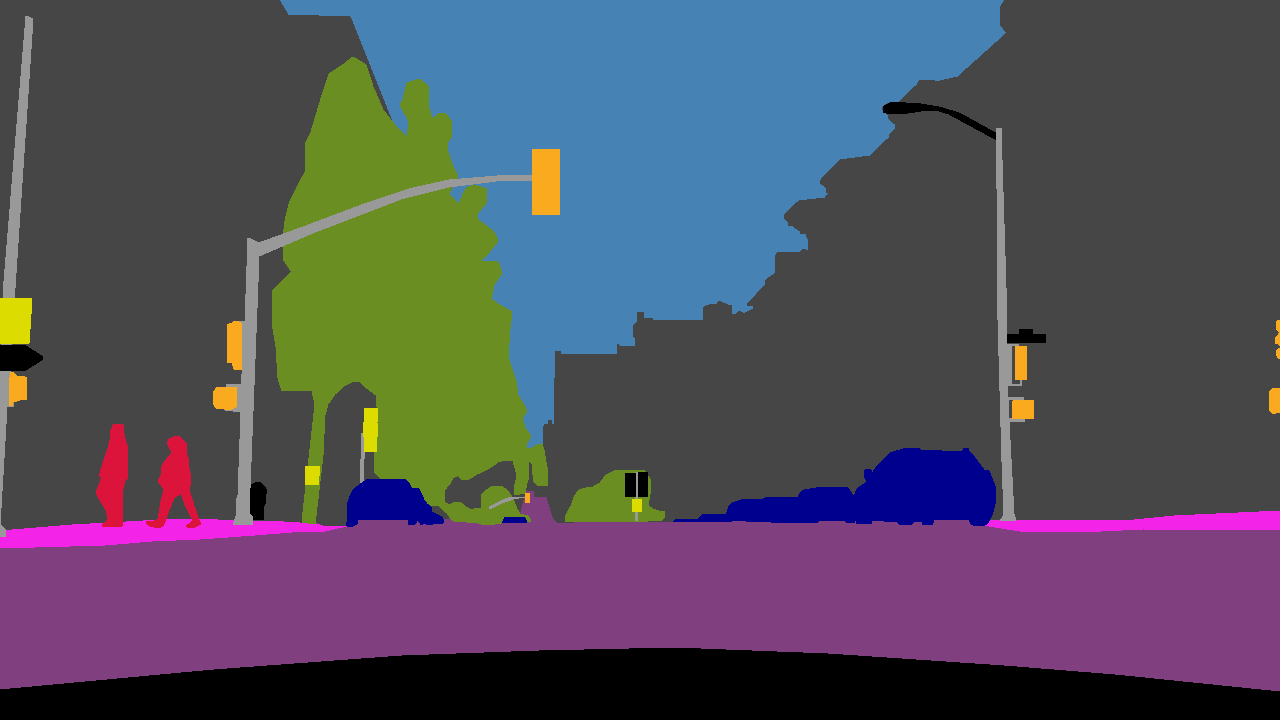}} &
  \subfloat{\includegraphics[width=0.15\textwidth]{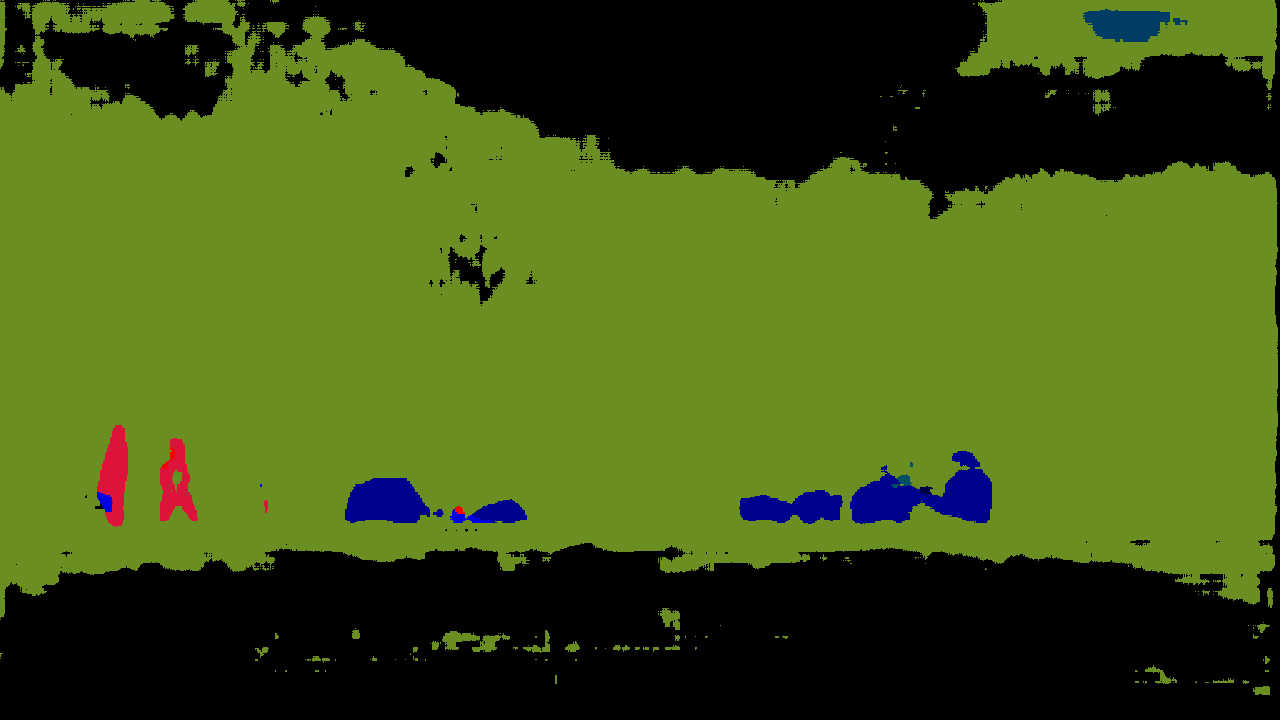}} &
  \subfloat{\includegraphics[width=0.15\textwidth]{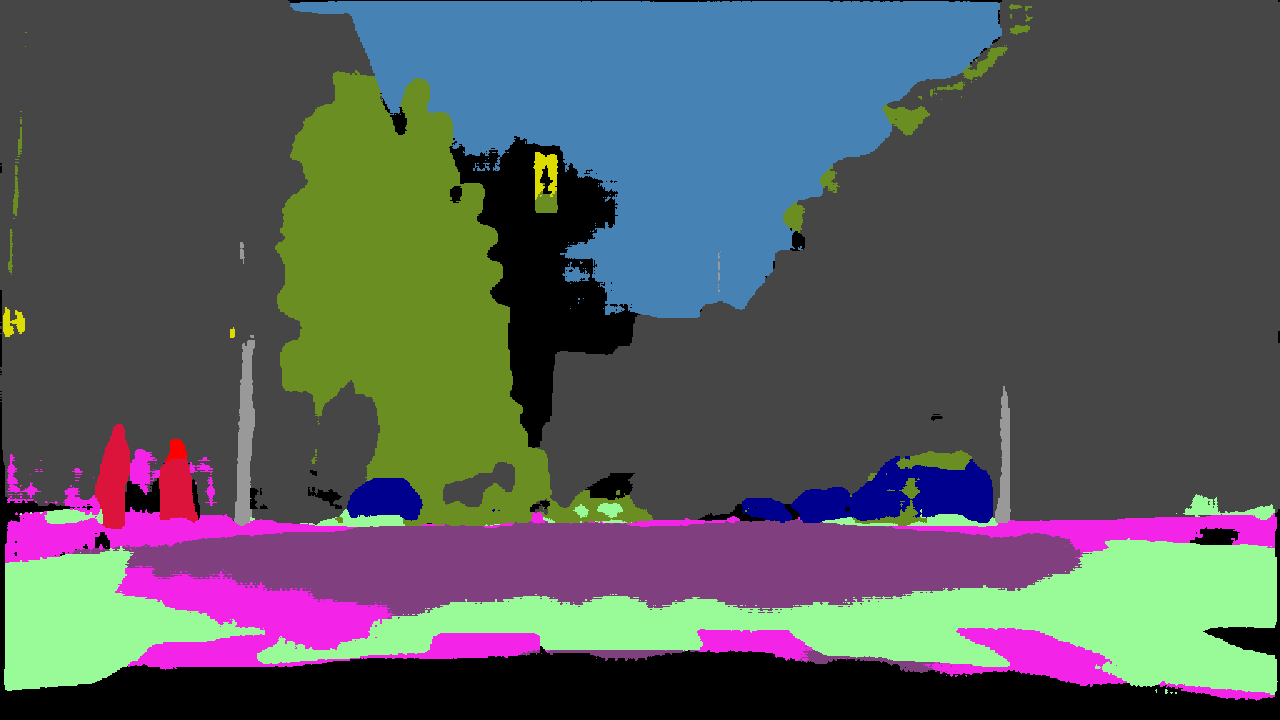}} &
  \subfloat{\includegraphics[width=0.15\textwidth]{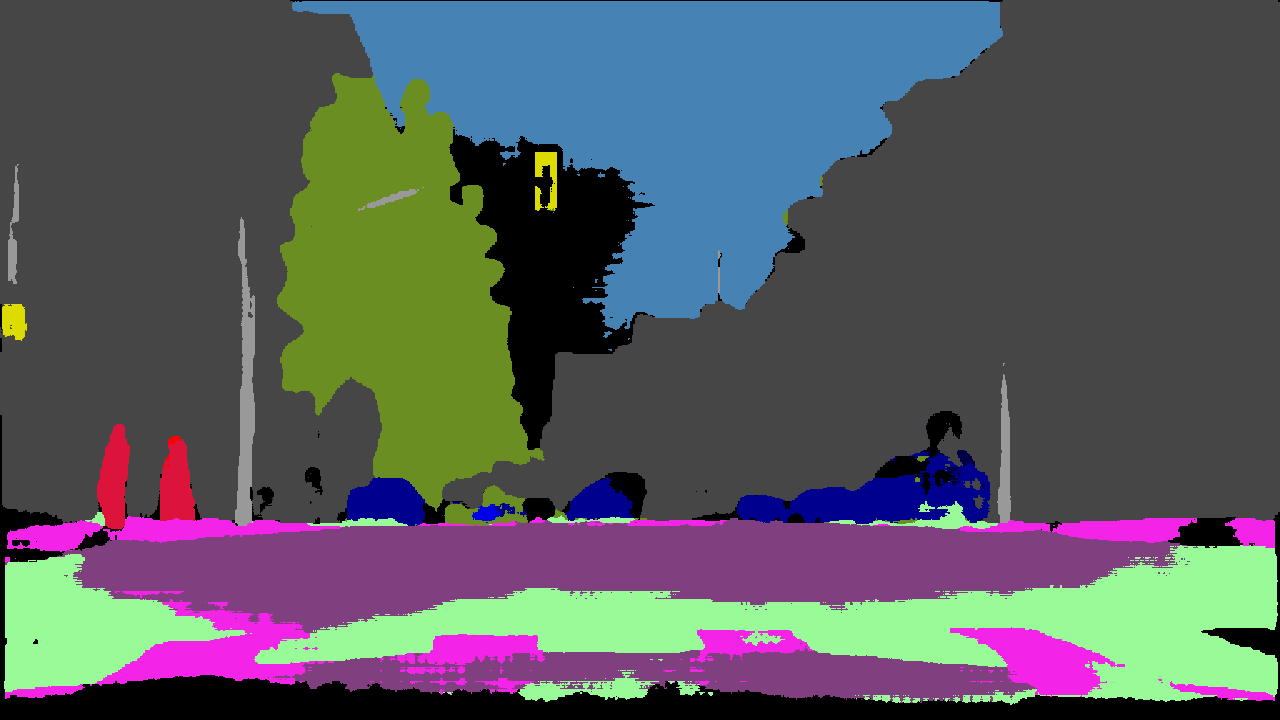}} &
  \subfloat{\includegraphics[width=0.15\textwidth]{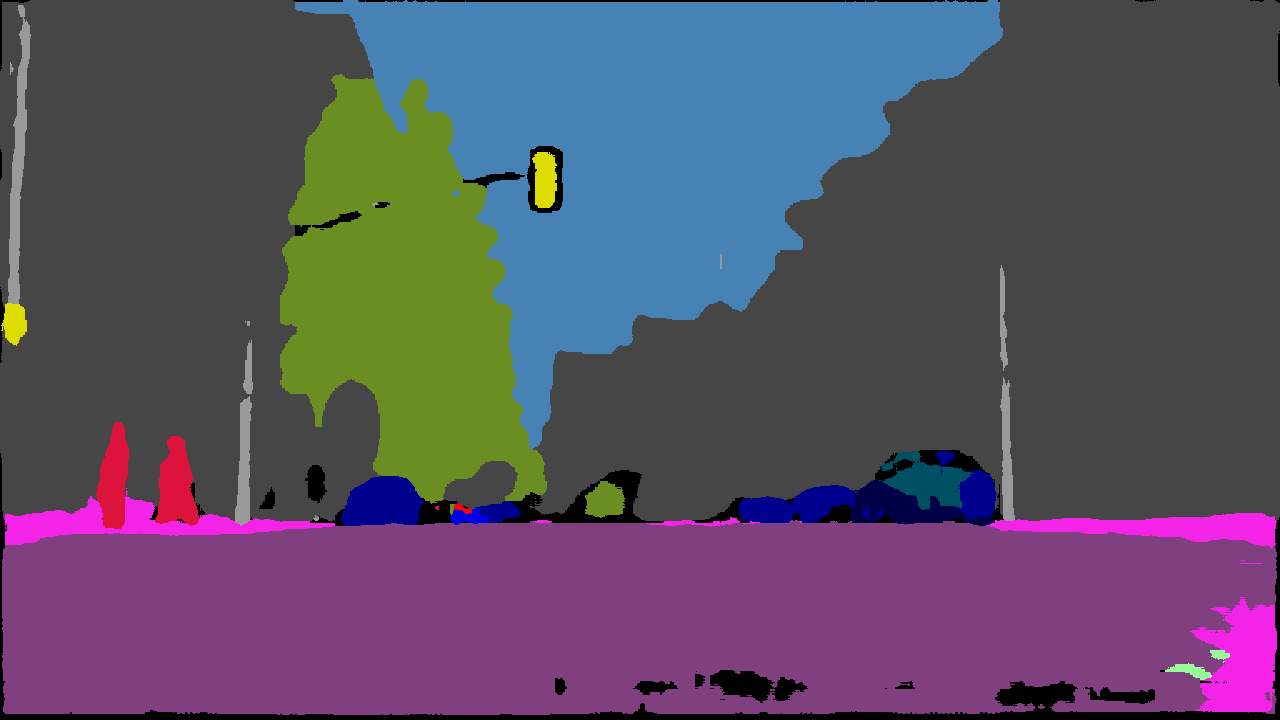}} &
  \raisebox{1.8\normalbaselineskip}[0pt][0pt]{\rotatebox[origin=c]{270}{BDD}} \\[-2ex]
  
&
  
  \subfloat{\includegraphics[width=0.15\textwidth]{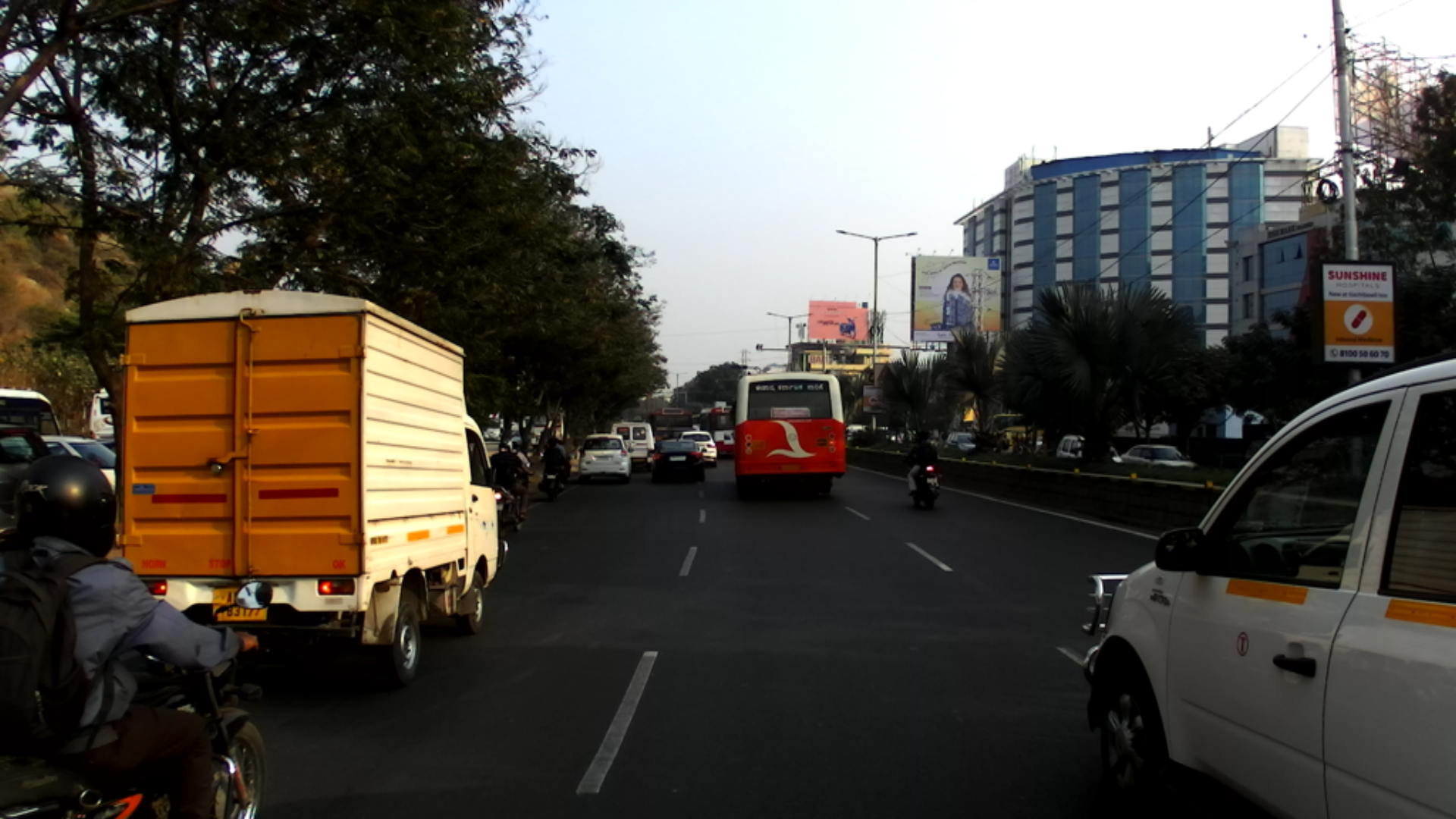}} &
  \subfloat{\includegraphics[width=0.15\textwidth]{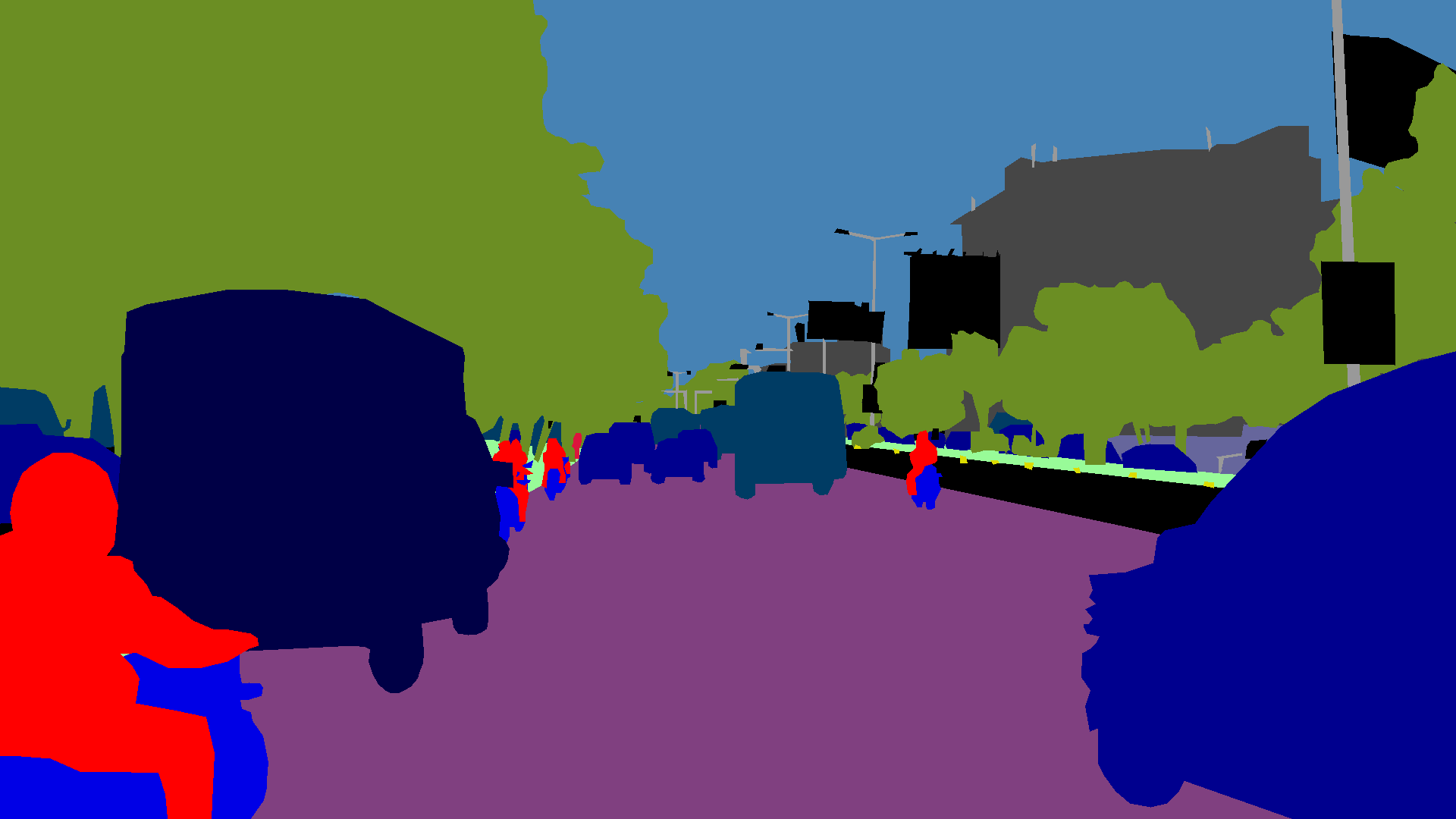}} &
  \subfloat{\includegraphics[width=0.15\textwidth]{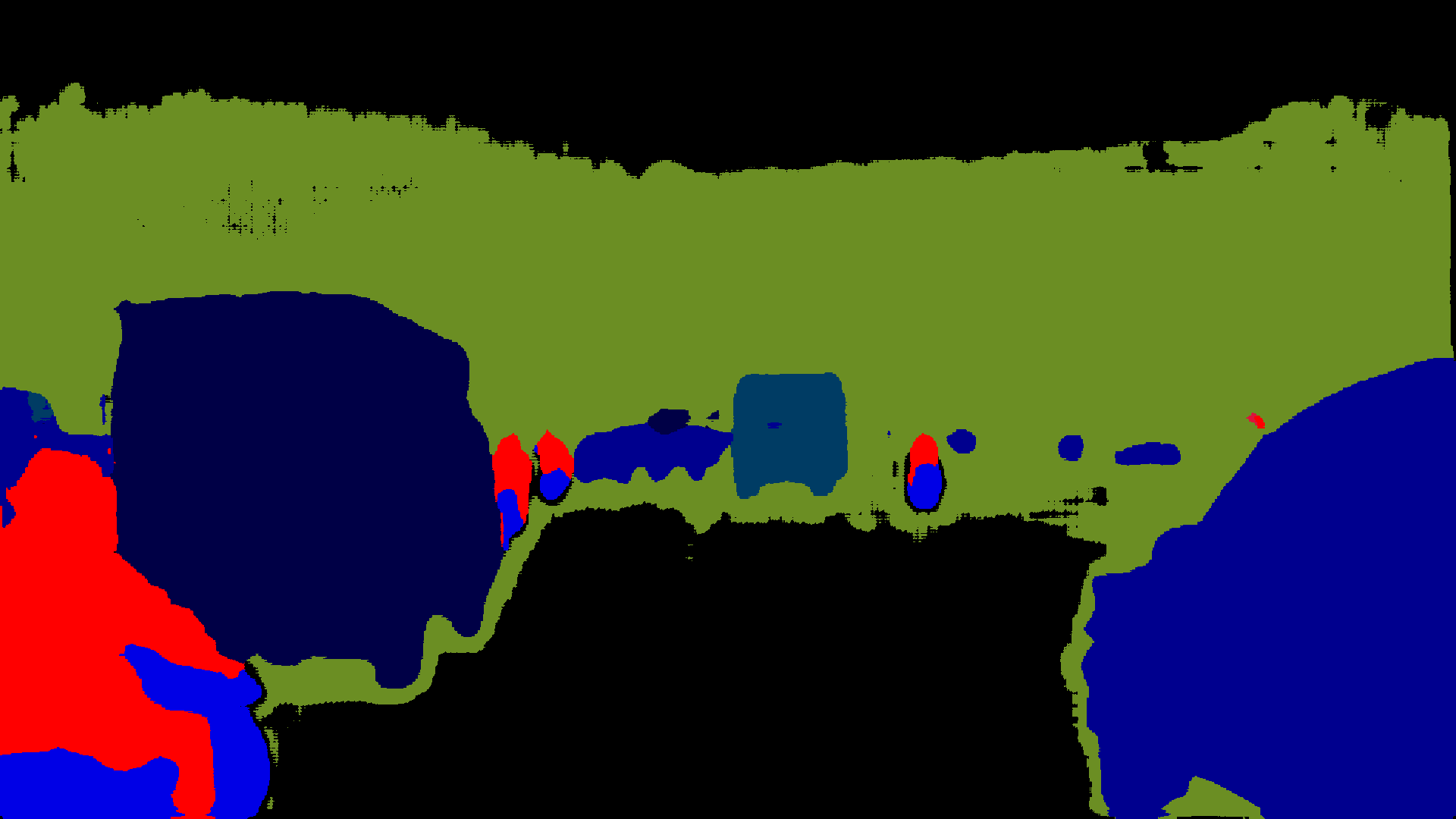}} &
  \subfloat{\includegraphics[width=0.15\textwidth]{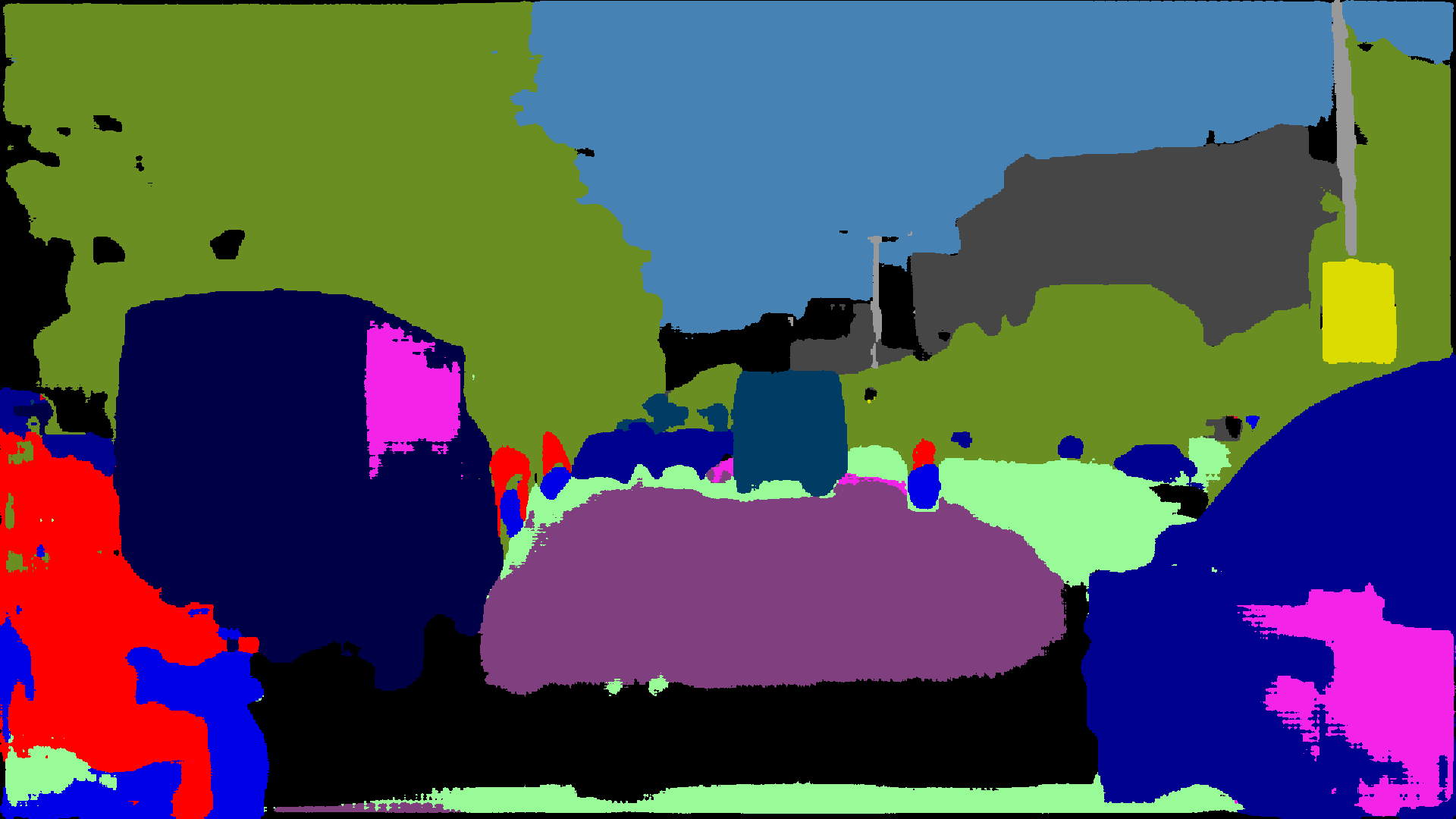}} &
  \subfloat{\includegraphics[width=0.15\textwidth]{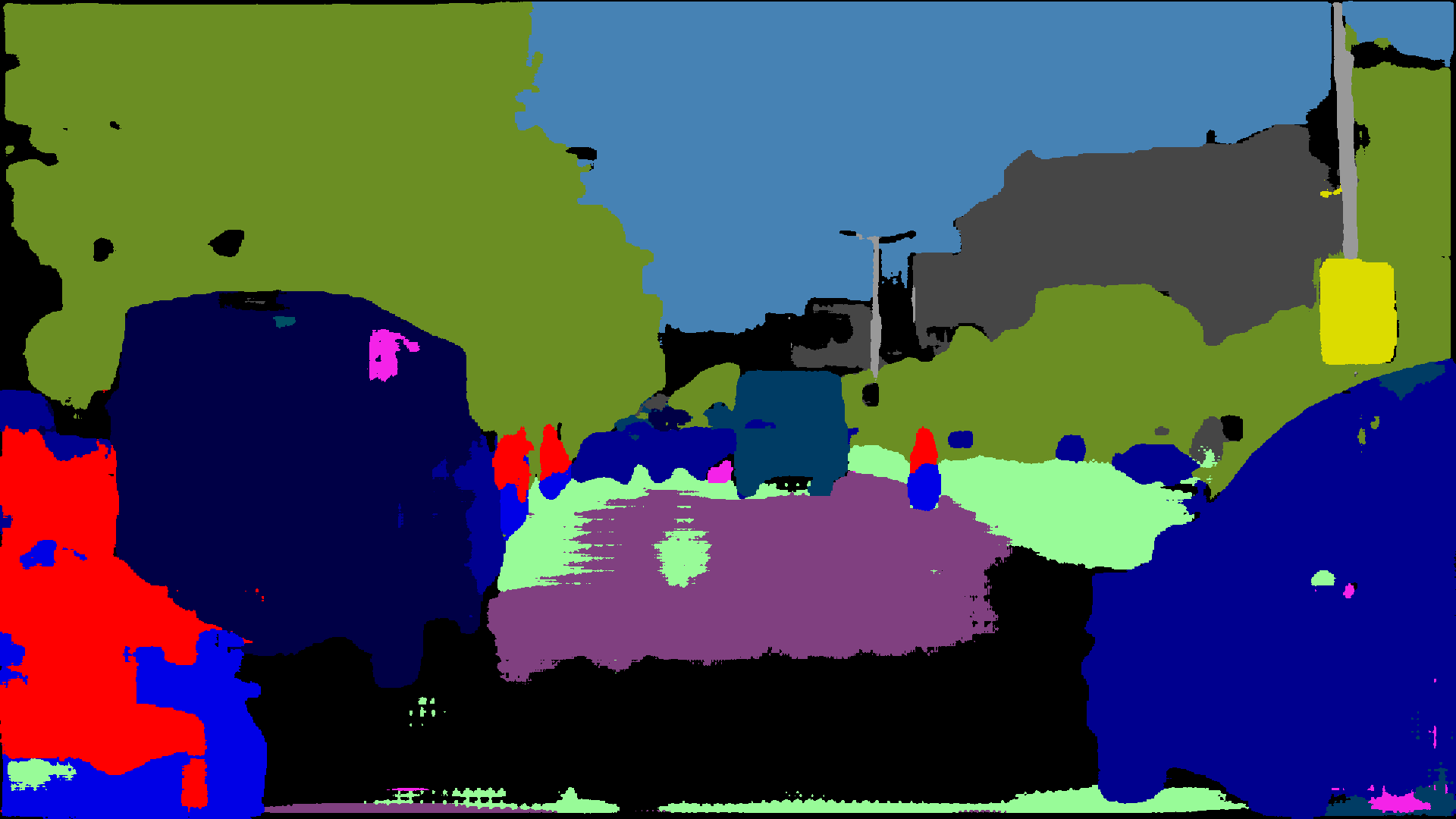}} &
  \subfloat{\includegraphics[width=0.15\textwidth]{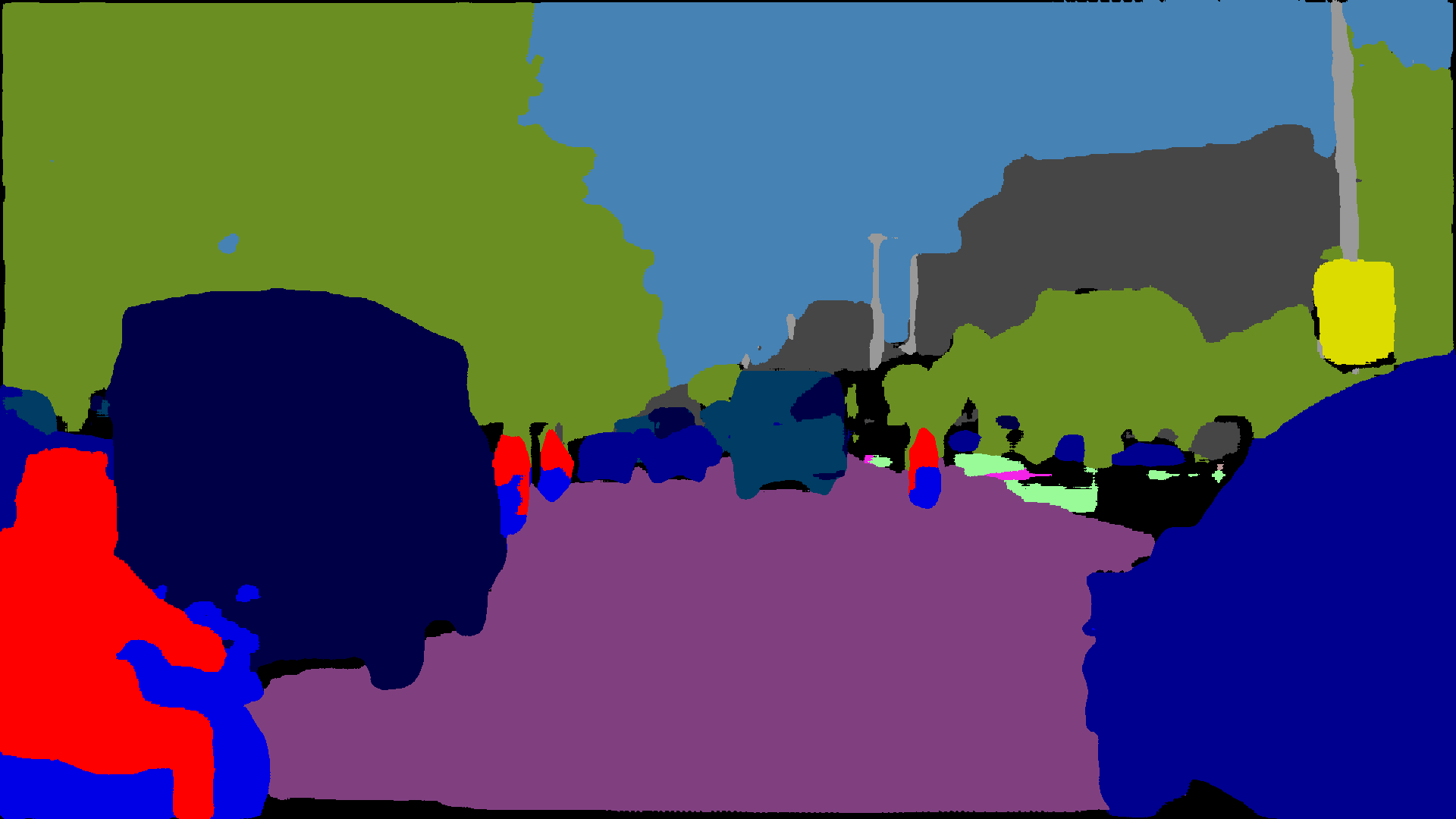}} &
  \raisebox{1.8\normalbaselineskip}[0pt][0pt]{\rotatebox[origin=c]{270}{\textbf{IDD}}} \\

  \end{tabular}
  \captionof{figure}{Qualitative results on $\text{CS} \protect\rightarrow  \text{BDD} \protect\rightarrow  \text{IDD}$ domain setup and $\mathcal{C}_{bgr} \protect\rightarrow  \mathcal{C}_{stat} \protect\rightarrow  \mathcal{C}_{mov}$ class setup.}
  \label{fig:qualitative}
  \end{minipage}
  
\end{table*}

Finally, qualitative results in the form of segmentation maps are provided in Fig.~\ref{fig:qualitative}.
We stress how the proposed approach yields better \textbf{backward} and \textbf{forward transfer} throughout the incremental learning.
In particular, moving classes like \textit{bicycle} and \textit{bus} appear to be recognized more effectively by our method on the Cityscapes (CS) dataset at the end of the incremental training, even though CS was experienced only along with \marco{background-class} supervision during the first step.
On the other hand, MiB and PLOP fail to provide satisfactory \textbf{backward transfer} of those classes to the past CS domain.
A similar reasoning can be done regarding the \textbf{forward transfer} aptitude.
Our approach is able to deliver good segmentation accuracy on the \textit{road} and \textit{sidewalk} background classes even on BDD and IDD datasets, despite them being experienced when $\mathcal{C}_{bgr}$ supervision is no longer available.
Contrarily, MiB and PLOP suffer from the domain statistical gap across learning steps, struggling to maintain satisfactory segmentation accuracy \pietro{on} first-step classes by forward transferring knowledge to future steps.
\tocheck{Additional \um{analyses} will be provided in Sec.~\ref{sec:abl_transfer}.}

\begin{table}[t!]
 \scriptsize
 \centering
  \setlength{\tabcolsep}{3.pt}
  \renewcommand{\arraystretch}{1.}
 \caption{\marco{Experimental results with DeeplabV3-ResNet101.}
 }

  \begin{tabu*}{c @{\hspace*{0.3cm}} cc ccccccc}
  \toprule
  \multicolumn{3}{c}{$\text{CS} \veryshortarrow \text{BDD} \veryshortarrow \text{IDD}$}  & \multicolumn{7}{c}{Method} \\
  \multicolumn{3}{c}{$\mathcal{C}_{bgr} \veryshortarrow \mathcal{C}_{stat} \veryshortarrow \mathcal{C}_{mov}$} 
  & $\mathcal{L}_{\scaleto{ce\mathstrut}{4pt}}^{\scaleto{n\mathstrut}{4pt}}$ 
  & $\mathcal{L}_{\scaleto{ce\mathstrut}{4pt}}^{\scaleto{\tilde{n}\mathstrut}{4pt}}$ 
  & MiB %
  & PLOP 
  & UCD
  & \name 
  & Oracle \\
  \midrule
  \multirow{2}{*}{Step 0} & \multirow{1}{*}{mIoU\textsubscript{0} $ \! \uparrow$} & \textbf{CS}  & 78.1 & 77.13 & 78.1 & 78.1 & 78.1 & 77.13 & \oracle{83.63}{84.30} \\
  
                          \arrayrulecolor{black!50}
                         \cmidrule{2-10}
                         \arrayrulecolor{black!100}
                         
                          & \multicolumn{2}{c}{$\bar{\Delta}_{0} \! \downarrow$}               & \oracle{6.61}{\tblbold{7.35}} & \oracle{7.77}{8.50} & \oracle{6.61}{\tblbold{7.35}} & \oracle{6.61}{\tblbold{7.35}} & \oracle{6.61}{\tblbold{7.35}} & \oracle{7.77}{8.50} & - \\
 \midrule                          
 \multirow{3}{*}{Step 1} & \multirow{2}{*}{mIoU\textsubscript{1} $ \! \uparrow$} & \textbf{BDD}   & 31.97 & 29.44 & 28.38 & 29.07 & 30.84 & 50.36 & \oracle{64.02}{64.60} \\
                         & & CS  & 31.82 & 55.13 & 45.10 & 45.15 & 45.51 & 54.75 & \oracle{69.73}{71.24} \\
                         
                       \arrayrulecolor{black!50}
                        \cmidrule{2-10}
                        \arrayrulecolor{black!100}
                         
                         & \multicolumn{2}{c}{$\bar{\Delta}_{1} \! \downarrow$}      & \oracle{52.22}{52.92} & \oracle{55.53}{56.19} & \oracle{45.50}{46.38} & \oracle{44.92}{45.81} & \oracle{43.28}{44.19} & \oracle{21.41}{\tblbold{22.60}} & - \\
 \midrule 
 \multirow{4}{*}{Step 2} & \multirow{3}{*}{mIoU\textsubscript{2} $ \! \uparrow$} & \textbf{IDD}  & 30.68 & 30.29 & 35.93 & 33.98 & 38.24 & 51.92 & \oracle{70.64}{70.94}\\
                         & & BDD  & 17.48 & 17.27 & 24.18 & 23.57 & 26.14 & 44.98 & \oracle{59.22}{61.48} \\
                         & & CS   & 19.46 & 17.92 & 33.22 & 34.38 & 34.93 & 47.08 & \oracle{70.64}{69.17} \\
                         
                       \arrayrulecolor{black!50}
                        \cmidrule{2-10}
                        \arrayrulecolor{black!100}
                         
                         & \multicolumn{2}{c}{$\bar{\Delta}_{2} \! \downarrow$}      & \oracle{66.10}{66.73} & \oracle{67.16}{67.77} & \oracle{53.06}{53.99} & \oracle{53.76}{54.68} & \oracle{50.03}{51.02} & \oracle{26.99}{\tblbold{28.53}} & -\\

  \bottomrule
 \end{tabu*}
 \label{tab:CBI_resnet}
\end{table}

\subsubsection{
\marco{Study on Class Ordering}
}
We further investigate the impact of \pietro{a permutation of  the} class incremental arrangement.
Table \ref{tab:CBI_rCIL} reports experimental results with \marco{the} $\text{CS} \rightarrow \text{BDD} \rightarrow \text{IDD}$ \marco{progression}, \marco{but} a modified class order with moving categories $\mathcal{C}_{mov}$ experienced before static ones $\mathcal{C}_{stat}$.
We notice a similar trend to that observed in Table \ref{tab:CBI} (\ie, same domain order, but different class order), with baselines and MDIL \cite{garg2022multi} performing poorly, and the improved accuracy achieved by CIL methods still being largely outperformed by the proposed approach.
\\
\marco{In addition}, we observe that the absolute results are decreased by applying the new class order.
The \um{performance} of our approach, in fact, \marco{drops} from $31.28\%$ to $39.29\%$ of $\bar \Delta_{2}$.
This discrepancy might be due to class sets observed on domains where it is harder to learn them, and, at the same time, to generalize to the other domains.
For instance, we note that IDD provides a lower overall percentage of pixels of $\mathcal{C}_{stat}$ w.r.t.\ the BDD ($11\%$ vs $17\%$), while for $\mathcal{C}_{mov}$ numbers are similar between them (both around $10\%$ of total pixels). %
\marco{Still}, the performance \marco{loss} is similar for CIL methods, with \marco{the} gap w.r.t.\ the best competitor  rising from $12$ to $13$ points \um{of $\bar\Delta_{2}$} \um{(compared to the previous class order)}.

\begin{table}[t!]
 \scriptsize
 \centering
  \setlength{\tabcolsep}{3.pt}
  \renewcommand{\arraystretch}{0.9}
  \caption{\marco{Experimental results on the Mapillary dataset.}}

  \begin{tabu}{cccccccccc}
  \toprule
  \multicolumn{3}{c}{
  $\text{EU} \!\veryshortarrow\! \text{NA} \!\veryshortarrow\! \text{AS} \!\veryshortarrow\! \text{OC} \!\veryshortarrow\! \text{AF} \!\veryshortarrow\! \text{SA}$
  } 
  & \multicolumn{7}{c}{Method} \\
  
  \multicolumn{3}{c}{$\mathcal{C}^{0 \!\rightarrow\! 1}_{bgr} \veryshortarrow \mathcal{C}^{0 \!\rightarrow\! 1}_{stat} \veryshortarrow \mathcal{C}^{0 \!\rightarrow\! 1}_{mov}$}
  & $\mathcal{L}_{\scaleto{ce\mathstrut}{4pt}}^{\scaleto{n\mathstrut}{4pt}}$ 
  & $\mathcal{L}_{\scaleto{ce\mathstrut}{4pt}}^{\scaleto{\tilde{n}\mathstrut}{4pt}}$ 
  & MiB %
  & PLOP 
  & UCD
  & \name 
  & Oracle\\
  \midrule
  \multirow{2}{*}{Step 0} & \multirow{1}{*}{mIoU\textsubscript{0} $ \! \uparrow$} & \textbf{EU}  & 73.12 & 73.07 & 73.12 & 73.12 & 73.12 & 73.07 & \oracle{80.08}{79.53} \\
  
                         \arrayrulecolor{black!50}
                         \cmidrule{2-10}
                         \arrayrulecolor{black!100}
  
                          & \multicolumn{2}{c}{$\bar{\Delta}_{0} \! \downarrow$}               &  \oracle{8.69}{\tblbold{8.06}} &  \oracle{8.75}{8.13} & \oracle{8.69}{\tblbold{8.06}} & \oracle{8.69}{\tblbold{8.06}} & \oracle{8.69}{\tblbold{8.06}} &  \oracle{8.75}{8.13} & - \\
 \midrule                          
 \multirow{3}{*}{Step 1} & \multirow{2}{*}{mIoU\textsubscript{1} $ \! \uparrow$} & \textbf{NA}   & 51.80 & 51.63 & 81.28 & 80.82 & 81.70 & 81.85 & \oracle{86.81}{87.51} \\
                         & & EU  & 47.67 & 47.52 & 76.05 & 75.76 & 75.26 & 74.80 & \oracle{82.05}{82.34} \\
                         \arrayrulecolor{black!50}
                         \cmidrule{2-10}
                         \arrayrulecolor{black!100}
                         & \multicolumn{2}{c}{$\bar{\Delta}_{1} \! \downarrow$}      & \oracle{41.11}{41.46} & \oracle{41.30}{41.65} & \oracle{6.84}{\tblbold{7.38}} & \oracle{7.28}{7.82} & \oracle{7.08}{7.62} & \oracle{7.27}{7.82} & - \\
 \midrule 
 \multirow{4}{*}{Step 2} & \multirow{3}{*}{mIoU\textsubscript{2} $ \! \uparrow$} & \textbf{AS}  & 25.18 & 26.09 & 65.40 & 65.98 & 65.70 & 65.36 & \oracle{73.09}{74.70} \\
                         & & NA  & 23.61 & 23.82 & 69.28 & 69.63 & 68.66 & 69.77 & \oracle{77.62}{79.40} \\
                         & & EU  & 23.98 & 24.10 & 66.66 & 66.86 & 65.79 & 65.87 & \oracle{76.04}{76.62} \\
                         \arrayrulecolor{black!50}
                         \cmidrule{2-10}
                         \arrayrulecolor{black!100}
                         & \multicolumn{2}{c}{$\bar{\Delta}_{2} \! \downarrow$}      & \oracle{67.87}{68.42} & \oracle{67.31}{67.87} & \oracle{11.20}{12.73} & \oracle{10.70}{\tblbold{12.24}} & \oracle{11.71}{13.23} & \oracle{11.36}{12.89} & - \\
 \midrule 
 \multirow{5}{*}{Step 3} & \multirow{4}{*}{mIoU\textsubscript{3} $ \! \uparrow$} & \textbf{OC}  & 16.53 & 16.74 & 61.29 & 60.58 & 60.53 & 63.07 & \oracle{70.42}{76.46} \\
                         & & AS  & 14.31 & 14.22 & 57.95 & 57.60 & 57.61 & 58.13 & \oracle{68.85}{70.96} \\
                         & & NA  & 17.10 & 17.20 & 62.41 & 63.29 & 61.91 & 64.04 & \oracle{74.64}{75.77} \\
                         & & EU  & 14.94 & 14.94 & 59.78 & 60.01 & 59.34 & 61.15 & \oracle{72.69}{72.97} \\
                         \arrayrulecolor{black!50}
                         \cmidrule{2-10}
                         \arrayrulecolor{black!100}
                         & \multicolumn{2}{c}{$\bar{\Delta}_{3} \! \downarrow$}      & \oracle{78.07}{78.79} & \oracle{77.99}{78.72} & \oracle{15.74}{18.47} & \oracle{15.74}{18.46} & \oracle{16.45}{19.15} & \oracle{14.02}{\tblbold{16.82}} & - \\
 \midrule 
 \multirow{6}{*}{Step 4} & \multirow{4}{*}{mIoU\textsubscript{4} $ \! \uparrow$} & \textbf{AF}  & 8.98 & 7.77 & 38.48 & 39.97 & 40.54 & 43.93 & \oracle{54.31}{66.54} \\
                         & & OC  & 6.03 & 5.95 & 40.17 & 43.52 & 42.15 & 47.43 & \oracle{63.74}{72.30} \\
                         & & AS  & 7.23 & 7.31 & 39.15 & 41.09 & 42.03 & 46.13 & \oracle{67.11}{69.87} \\
                         & & NA  & 7.78 & 7.07 & 43.10 & 45.12 & 45.00 & 50.07 & \oracle{71.99}{74.22} \\
                         & & EU  & 5.45 & 5.41 & 38.99 & 41.52 & 41.28 & 46.37 & \oracle{69.57}{70.22} \\
                         \arrayrulecolor{black!50}
                         \cmidrule{2-10}
                         \arrayrulecolor{black!100}
                         & \multicolumn{2}{c}{$\bar{\Delta}_{4} \! \downarrow$}      & \oracle{88.92}{89.91} & \oracle{89.57}{90.48} & \oracle{38.38}{43.40} & \oracle{34.91}{40.20} & \oracle{34.95}{40.24} & \oracle{27.95}{\tblbold{33.77}} & - \\
 \midrule 
 \multirow{7}{*}{Step 5} & \multirow{5}{*}{mIoU\textsubscript{5} $ \! \uparrow$} & \textbf{SA}  & 9.61 & 9.18 & 39.64 & 41.79 & 41.48 & 45.36 & \oracle{59.45}{64.45} \\
                         & & AF  & 11.35 & 9.63 & 40.76 & 41.98 & 41.44 & 45.25 & \oracle{50.03}{63.03} \\
                         & & OC  & 6.78 & 7.42 & 36.52 & 39.08 & 37.21 & 41.76 & \oracle{50.36}{60.82} \\
                         & & AS  & 8.97 & 8.17 & 37.90 & 40.15 & 38.76 & 43.63 & \oracle{60.67}{64.74} \\
                         & & NA  & 8.97 & 9.54 & 41.40 & 43.45 & 43.05 & 47.08 & \oracle{62.56}{66.88} \\
                         & & EU  & 8.18 & 7.63 & 38.37 & 40.53 & 38.93 & 43.51 & \oracle{64.39}{64.04} \\
                         \arrayrulecolor{black!50}
                         \cmidrule{2-10}
                         \arrayrulecolor{black!100}
                         & \multicolumn{2}{c}{$\bar{\Delta}_{5} \! \downarrow$}      & \oracle{84.31}{85.98} & \oracle{85.00}{86.58} & \oracle{31.84}{38.90} & \oracle{28.27}{35.67} & \oracle{30.05}{37.28} & \oracle{22.60}{\tblbold{30.57}} & - \\

  \bottomrule
 \end{tabu}
 \label{tab:map_cont}
\end{table}

\subsubsection{Study on Model Architecture}

\marco{We \tocheck{finally} evaluate the considered methods when a more \pietro{complex} segmentation network is used, moving from the lightweight ErfNet to the heavier DeeplabV3 with ResNet101 backbone.}
\marco{For comparison purposes,} the setup analyzed is again that involving $\text{CS} \rightarrow \text{BDD} \rightarrow \text{IDD}$ and $\mathcal{C}_{bgr} \rightarrow \mathcal{C}_{stat} \rightarrow \mathcal{C}_{mov}$ orders \marco{(Table~\ref{tab:CBI_resnet})}.
\marco{For what concerns our approach, we observe an improved relative performance, raising from $31.28\%$ to $28.53\%$ in terms of $\bar\Delta_{2}$.
We emphasize that the $\bar\Delta$ measure already takes into account the better oracle results; the accuracy boost, then, shows that our method is able to capitalize the increased capacity offered by the segmentation model.
}

On the other hand, the CIL competitors are unable to take advantage of the \marco{growth} in network capacity, which could indicate a tendency to overfit on the currently observed domain distribution.
\marco{
The best competitor (\ie, UCD), in fact, is significantly outperformed by more than 20\% in terms of $\bar\Delta$ at both steps 1 and 2.
}
We remark that no additional parameter tuning is performed in this experimental setup concerning method-specific parameters.

\subsection{
Evaluation with Larger Geographic Diversity
}
The second experimental \marco{class and domain incremental} setup we explore is derived from the Mapillary dataset.
Domain shift is once more induced by \marco{the variable} geographic \marco{origin} of image samples \marco{collected worldwide}, \ie, we identify data partitions associated to 6 different continents, corresponding to 6 incremental steps.
However, the Mapillary dataset contains variegate data distribution, even considering intra-continent samples, providing a more robust support for training segmentation models. %
Data richness in turns promotes generalization across steps, in fact lessening the domain gap between different domains.
We report experimental results in Table \ref{tab:map_cont}. %
\pietro{In the first steps,  when the domain shift is small (\eg, between Europe\um{, EU,} and North America\um{, NA}), the different methods achieve similar performance.} 
Nonetheless, when progressing to the last steps and experiencing increased statistical gap (\eg, when introducing Africa's images\um{, AF}), we note that our approach outperforms CIL competitors by a \marco{considerable} margin, which is of $5$ points of $\bar\Delta$ w.r.t.\ the  best competitor (PLOP) at the end of incremental training.
\marco{Also, superior performance in later steps is attained \pietro{on}  both new and old domains, confirming the better plasticity-stability trade-off provided by our method.}
\marco{Overall, the improved results 
\um{\name\ reaches w.r.t.\ state-of-the-art CIL}~competitors, even when training data is collected to ensure some statistical diversity (as in the experimental setup just considered), further suggests that CIL methods are likely to be inadequate to deal with distribution shift in the input space.}

\begin{table}[t!]
 \scriptsize
 \centering
  \setlength{\tabcolsep}{2.9pt}
  \renewcommand{\arraystretch}{1.1}
 \caption{
 \marco{Experimental results on the Shift dataset.}
 }
  
  \newcommand\cw{6.4mm}
  \newcommand\cwd{6.mm}
  \begin{tabu}{l | >{\centering}m{\cw}>{\centering}m{\cwd}| >{\centering}m{\cw}>{\centering}m{\cwd}| >{\centering}m{\cw}>{\centering}m{\cwd}| >{\centering}m{\cwd}}
  \toprule
  \multicolumn{1}{c|}{$\text{Daytime} \! \veryshortarrow \! \text{Twilight} \! \veryshortarrow \! \text{Night}$}
  & \multicolumn{2}{c}{\textbf{Night}} & \multicolumn{2}{c}{Twilight} & \multicolumn{2}{c}{Daytime} & \\
  
  \multicolumn{1}{c|}{$\mathcal{C}_{bgr} \veryshortarrow \mathcal{C}_{stat} \veryshortarrow \mathcal{C}_{mov}$} 
  & \scalebox{1.}{$\text{mIoU}_{2}^{2}$ $ \!\! \uparrow$} & $\Delta_{2}^{2} \!\! \downarrow$ 
  & \scalebox{1.}{$\text{mIoU}_{2}^{1}$ $ \!\! \uparrow$} & $\Delta_{2}^{1} \!\! \downarrow$ 
  & \scalebox{1.}{$\text{mIoU}_{2}^{0}$ $ \!\! \uparrow$} & $\Delta_{2}^{0} \!\! \downarrow$ 
  & $\bar{\Delta}_{2} \! \downarrow$ \\

\midrule

FT ($\mathcal{L}_{ce}^{n}$)  
  & 10.54 & \oracle{-}{85.82} & 9.62 & \oracle{-}{87.21} & 4.61 & \oracle{-}{94.06} & \oracle{-}{89.03} \\
 
FT w/ self-style ($\mathcal{L}_{ce}^{\tilde{n}}$) 
 & 10.12 & \oracle{-}{86.39} & 8.50 & \oracle{-}{88.70} & 7.56 & \oracle{-}{90.26} & \oracle{-}{88.45} \\
  
MiB
  & 48.07 & \oracle{-}{35.35} & 52.71 & \oracle{-}{29.92} & 48.29 & \oracle{-}{37.77} & \oracle{-}{34.34} \\
  
PLOP
  & 48.58 & \oracle{-}{34.67} & 53.66 & \oracle{-}{28.66} & 51.11 & \oracle{-}{34.13} & \oracle{-}{32.48} \\
  
\name
  & 60.27 & \oracle{-}{18.94} & 62.57 & \oracle{-}{16.81} & 59.78 & \oracle{-}{22.97} & \oracle{-}{\tblbold{19.57}} \\
  \midrule
  Oracle
    & 74.35 & \oracle{-}{-} & 75.21 & \oracle{-}{-} & 77.60 & \oracle{-}{-} & \oracle{-}{-} \\
  
  \bottomrule
 \end{tabu}
 \label{tab:shift}
 \vspace{-0mm}
\end{table}

\subsection{Evaluation with Variable Environmental Conditions}

We evaluate the proposed method when incremental domain shift is due to changing environmental \marco{factors}, \ie, \marco{variable light conditions experienced at different times during the day. In this setting, we employed the Shift synthetic benchmark.}
\tocheck{We consider the $\textit{Daytime} \rightarrow\allowbreak \textit{Twilight} \rightarrow\allowbreak \textit{Night}$ domain sequence.}
\marco{Class incremental scheduling follows the $\mathcal{C}_{bgr} \rightarrow \allowbreak\mathcal{C}_{stat} \rightarrow\allowbreak \mathcal{C}_{mov}$ arrangement of \cite{klingner2020class}, with the only difference from \cite{klingner2020class} being that the starting class pool to be split corresponds to the 22 Shift's categories in place of the 19 Cityscape's ones.}
Results are reported in Table \ref{tab:shift}, where we \marco{compare} with MiB and PLOP \um{as} CIL competitors, along with fine-tuning baselines.
We verify the superiority of our approach in jointly handling class and domain incremental training, as we surpass PLOP by $13$ points of \um{$\bar\Delta_{2}$}.
We once more point out the better stability-plasticity \marco{balance} reached by our method, which achieves improved performance simultaneously over novel and \marco{former domains}.
\marco{
Overall, results show that the proposed method is effective
\marco{under domain shifts of different nature.}
On the other hand, CIL methods prove to be greatly penalized just from the variable 
scene illumination in
different tasks.
We argue that in many real-world applications, such as autonomous driving, it is unrealistic to assume that a continual learner will not experience any sort of alteration in input data distribution, making our \pietro{continual learning} approach much more applicable.}

\section{Ablation Studies}
\label{sec:ablation}

\marco{
In this section\um{,} we provide extensive ablation studies 
to investigate key features of  our approach. 
We will consider the \textit{urban} experimental setup, with $\text{CS} \rightarrow  \text{BDD} \rightarrow  \text{IDD}$ domain and $\mathcal{C}_{bgr} \rightarrow  \mathcal{C}_{stat} \rightarrow  \mathcal{C}_{mov}$ class orders, unless otherwise stated.}

\begin{table}[t!]
 \scriptsize
 \centering
  \renewcommand{\arraystretch}{1.2}
 \caption{Ablation study on the contribution of loss components. The \tocheck{$\mathcal{L}_{\lossname}^{n}$} notation here implies that pseudo-labels are generated leveraging new-domain input samples.}

  \setlength{\tabcolsep}{2.7pt}
  \newcommand\cw{6.8mm}
  \newcommand\cwd{6.4mm}
  
  \begin{tabu}{l | >{\centering}m{\cw}>{\centering}m{\cwd}| >{\centering}m{\cw}>{\centering}m{\cwd}| >{\centering}m{\cw}>{\centering}m{\cwd}| >{\centering}m{\cwd}}
  \toprule
  \multicolumn{1}{c|}{$\text{CS} \veryshortarrow \text{BDD} \veryshortarrow \text{IDD}$}
  & \multicolumn{2}{c}{\textbf{IDD}} & \multicolumn{2}{c}{BDD} & \multicolumn{2}{c}{CS} & \\
  
  \multicolumn{1}{c|}{$\mathcal{C}_{bgr} \veryshortarrow \mathcal{C}_{stat} \veryshortarrow \mathcal{C}_{mov}$} 
  & \scalebox{1.}{$\text{mIoU}_{2}^{2}$ $ \!\! \uparrow$} & $\Delta_{2}^{2} \!\! \downarrow$ 
  & \scalebox{1.}{$\text{mIoU}_{2}^{1}$ $ \!\! \uparrow$} & $\Delta_{2}^{1} \!\! \downarrow$ 
  & \scalebox{1.}{$\text{mIoU}_{2}^{0}$ $ \!\! \uparrow$} & $\Delta_{2}^{0} \!\! \downarrow$ 
  & $\bar{\Delta}_{2} \! \downarrow$ \\
  
  \midrule
  
  $\mathcal{L}_{ce}^{n}$   
  & 26.27 & \oracle{61.45}{61.48} & 10.47 & \oracle{78.90}{81.72} & 12.10 & \oracle{82.18}{81.18} & \oracle{74.18}{74.79} \\
  
  $\mathcal{L}_{ce}^{\tilde{n}}$
  & 27.12 & \oracle{60.20}{60.24} & 11.51 & \oracle{76.80}{79.91} & 13.68 & \oracle{79.86}{78.72} & \oracle{\underline{72.29}}{\underline{72.95}} \\
  
  \midrule
  
  $\mathcal{L}_{ce}^{\tilde{n}} \!+\! 
  \mathcal{L}_{ce}^{\tilde{o}}$
  & 28.09 & \oracle{58.78}{58.81} & 13.32  & \oracle{73.16}{76.75} & 16.32 & \oracle{75.97}{74.61} & \oracle{69.30}{70.06} \\
  
  $\mathcal{L}_{ce}^{\tilde{n}} \!+\! 
  \mathcal{L}_{kd}^{\tilde{o}}$
  & 40.63 & \oracle{40.37}{40.43} & 24.95  & \oracle{49.72}{56.44} & 34.14 & \oracle{49.73}{46.90} & \oracle{46.61}{47.92} \\
  
  $\mathcal{L}_{ce}^{n} \!+\! 
  \tocheck{\mathcal{L}_{\lossname}^{n}}$
  & 43.33 & \oracle{36.41}{36.47} & 26.62 & \oracle{46.35}{53.53} & 37.36 & \oracle{44.99}{41.89} & \oracle{42.58}{43.96} \\
  
  $\mathcal{L}_{ce}^{\tilde{n}} \!+\!
  \mathcal{L}_{\lossname}^{\tilde{n}}$
  & 48.12 & \oracle{29.38}{29.45} & 32.40  & \oracle{34.70}{43.44} & 40.57 & \oracle{40.26}{36.89} & \oracle{\underline{34.78}}{\underline{36.59}} \\
  
  \midrule
  
  $\mathcal{L}_{ce}^{\tilde{n}} \!+\! 
  \mathcal{L}_{\lossname}^{\tilde{n}} \!+\! 
  \mathcal{L}_{kd}^{\tilde{o}}$
  & 19.23 & \oracle{71.78}{71.80} & 17.68  & \oracle{64.37}{69.13} & 24.15 & \oracle{64.44}{62.44} & \oracle{66.86}{67.79} \\
    
  $\mathcal{L}_{ce}^{\tilde{n}} \!+\! 
  \mathcal{L}_{ce}^{\tilde{o}} \!+\! 
  \mathcal{L}_{kd}^{\tilde{o}}$
  & 50.08 & \oracle{26.50}{26.57} & 34.16  & \oracle{31.16}{40.36} & 42.86 & \oracle{36.89}{33.33} & \oracle{31.52}{33.42} \\
  
$\mathcal{L}_{ce}^{\tilde{n}} \!+\! 
  \mathcal{L}_{\lossname}^{\tilde{n}} \!+\! 
  \mathcal{L}_{ce}^{\tilde{o}}$
  & 50.70 & \oracle{25.59}{25.66} & 34.86 & \oracle{29.75}{39.14} & 43.04 & \oracle{36.62}{33.05} & \oracle{\underline{30.65}}{\underline{32.62}} \\

\midrule

  $\mathcal{L}_{ce}^{n} \!+\! 
  \tocheck{\mathcal{L}_{\lossname}^{n}} \!+\!  %
  \mathcal{L}_{ce}^{\tilde{o}} \!+\! 
  \mathcal{L}_{kd}^{\tilde{o}}$
  & 46.59 & \oracle{31.63}{31.69} & 30.51 & \oracle{38.51}{46.74} & 40.44 & \oracle{40.45}{37.10} & \oracle{36.86}{38.51} \\
  
  $\mathcal{L}_{ce}^{\tilde{n}} \!+\! 
  \mathcal{L}_{\lossname}^{\tilde{n}} \!+\! 
  \mathcal{L}_{ce}^{\tilde{o}} \!+\! 
  \mathcal{L}_{kd}^{\tilde{o}}$ 
  & 51.20 & \oracle{24.86}{24.93} & 35.73 & \oracle{27.99}{37.62} & 44.17 & \oracle{34.96}{31.29} & \oracle{\underline{\textbf{29.27}}}{\underline{\textbf{31.28}}} \\
  
    \midrule
  Oracle
    & 68.20 & \oracle{-}{-} & 57.28 & \oracle{-}{-} & 64.29 & \oracle{-}{-} & \oracle{-}{-} \\
  
  \bottomrule
 \end{tabu}
 \label{tab:loss_comp}
 \vspace{-0mm}
\end{table}

\subsection{Contribution of Individual Optimization Objectives}
We investigate the impact of each of the proposed learning objectives in the overall optimization framework \pietro{
 in Table} \ref{tab:loss_comp}.
Just leveraging the currently available training data by fine-tuning (first two rows) yields unsatisfactory results (even with self-stylization), leading to catastrophic forgetting of class and domain knowledge.
Yet, $\mathcal{L}_{ce}^{n}$ (or $\mathcal{L}_{ce}^{\tilde{n}}$) is essential to learn new tasks, so it will be kept in the following analyses to test multi-term objectives.

\marco{By} adding a second term in the overall objective (second block of rows) we improve 
\marco{results,}
especially if the supplemental objective is focused on retaining old-class knowledge.
We reach, in fact, the best performance with a 2-term %
configuration
when $\mathcal{L}^{\tilde{n}}_{\lossname}$ is introduced.
This suggests that old-class knowledge preservation is effective even when applied on the new domain, which is directly experienced by means of the available training data. 
At the same time, the $\mathcal{L}^{\tilde{n}}_{\lossname}$ objective allows to retain good accuracy w.r.t. past domains, thanks to the improved generalization aptitude promoted by the \tocheck{stylization} %
mechanism, without which (\ie, third row of the block) multiple accuracy points are lost.

When analyzing 3-term objectives (third block of rows), we see noticeable gain with different combinations, except for $\mathcal{L}^{\tilde{n}}_{\lossname}$ and $\mathcal{L}^{\tilde{o}}_{kd}$ jointly active, where the excessive focus on past-class knowledge preservation generates training instability.
\marco{In the last row of the block, we clearly see that, by adding the $\mathcal{L}^{\tilde{o}}_{ce}$ loss on top of the best two-term configuration, the incremental learning becomes more robust, with improved final results on all domains.}

Finally, we remark that the \marco{full} framework \marco{(last block)} yields the best overall performance, with \tocheck{stylization} %
once more playing a substantial role.
The overall performance is, in fact, strongly degraded if 
\tocheck{stylization}
is turned off, as showed in the second last row.

\subsection{Pseudo-label Generation}
\label{sec:abl_pseudo}

\begin{table}[t!]
 \scriptsize
 \centering
  \setlength{\tabcolsep}{3.pt}
  \renewcommand{\arraystretch}{1.2}
 \caption{
 \marco{Ablation study on pseudo-labeling schemes. \tocheck{We added %
 $\{n,o\}$ %
 to the loss notation to indicate if pseudo-labels are generated leveraging new-domain 
 ($\mathcal{L}^{d}_{\marco{kd},n}$)
 or oldly-stylized 
 ($\mathcal{L}^{d}_{\marco{kd},o})$ %
 input samples.}}
 }

  \newcommand\cw{6.4mm}
  \newcommand\cwd{5.9mm}
  \begin{tabu}{l | >{\centering}m{\cw}>{\centering}m{\cwd}| >{\centering}m{\cw}>{\centering}m{\cwd}| >{\centering}m{\cw}>{\centering}m{\cwd}| >{\centering}m{\cwd}}
  
  \toprule
  \multicolumn{1}{c|}{$\text{CS} \veryshortarrow \text{BDD} \veryshortarrow \text{IDD}$}
  & \multicolumn{2}{c}{\textbf{IDD}} & \multicolumn{2}{c}{BDD} & \multicolumn{2}{c}{CS} & \\
  
  \multicolumn{1}{c|}{$\mathcal{C}_{bgr} \veryshortarrow \mathcal{C}_{stat} \veryshortarrow \mathcal{C}_{mov}$} 
  & \scalebox{1.}{$\text{mIoU}_{2}^{2}$ $ \!\! \uparrow$} & $\Delta_{2}^{2} \!\! \downarrow$ 
  & \scalebox{1.}{$\text{mIoU}_{2}^{1}$ $ \!\! \uparrow$} & $\Delta_{2}^{1} \!\! \downarrow$ 
  & \scalebox{1.}{$\text{mIoU}_{2}^{0}$ $ \!\! \uparrow$} & $\Delta_{2}^{0} \!\! \downarrow$ 
  & $\bar{\Delta}_{2} \! \downarrow$ \\

\midrule

  $\mathcal{L}_{ce}^{n} \! + \! 
  \mathcal{L}_{\marco{kd}, n}^{n} \! + \!   %
  \mathcal{L}_{ce}^{\tilde{o}} \! + \!  
  \mathcal{L}_{kd}^{\tilde{o}}$
  & 46.59 & \oracle{31.63}{31.69} & 30.51 & \oracle{38.51}{46.74} & 40.44 & \oracle{40.45}{37.10} & \oracle{36.86}{38.51} \\
 
  $\mathcal{L}_{ce}^{n} \! + \! 
  \mathcal{L}_{\marco{kd}, o}^{n} \! + \! %
  \mathcal{L}_{ce}^{\tilde{o}} \! + \! 
  \mathcal{L}_{kd}^{\tilde{o}}$
  & 40.09 & \oracle{41.17}{41.22} & 25.55  & \oracle{48.51}{55.40} & 34.90 & \oracle{48.61}{45.71}  & \oracle{46.09}{47.44} \\
  
    $\mathcal{L}_{ce}^{\tilde{n}} \! + \! 
  \mathcal{L}_{\marco{kd}, n}^{\tilde{n}} \! + \! 
  \mathcal{L}_{ce}^{\tilde{o}} \! + \! 
  \mathcal{L}_{kd}^{\tilde{o}}$ 
  & 51.11 & \oracle{24.99}{25.06} & 34.01 & \oracle{31.46}{40.63} & 43.96 & \oracle{35.27}{31.62} & \oracle{30.57}{32.44} \\
  
  $\mathcal{L}_{ce}^{\tilde{n}} \! + \!
  \mathcal{L}_{\marco{kd}, o}^{\tilde{n}} \! + \!
  \mathcal{L}_{ce}^{\tilde{o}} \! + \!
  \mathcal{L}_{kd}^{\tilde{o}}$ 
  & 51.20 & \oracle{24.86}{24.93} & 35.73 & \oracle{27.99}{37.62} & 44.17 & \oracle{34.96}{31.29} & \oracle{29.27}{\tblbold{31.28}} \\ 
  
  \midrule
  Oracle & 68.20 & \oracle{-}{-} & 57.28 & \oracle{-}{-} & 64.29 & \oracle{-}{-} & \oracle{-}{-} \\
  
  \bottomrule
 \end{tabu}
 \label{tab:pseudo}
\end{table}

\begin{table}[t]
  \centering
  \setlength{\tabcolsep}{1pt}
  \small
  \begin{minipage}{0.99\linewidth}\centering
  \begin{tabular}{ ccc }

  Image & GT & $ \fullhatt[-0.3ex]{\mathbf{Y}}^{\lower.75em\hbox{\fontsize{6}{6}\selectfont \{2\}}}_{t\shortminus1} $ \\[-2.2ex]
  
  \subfloat{\includegraphics[width=0.31\linewidth]{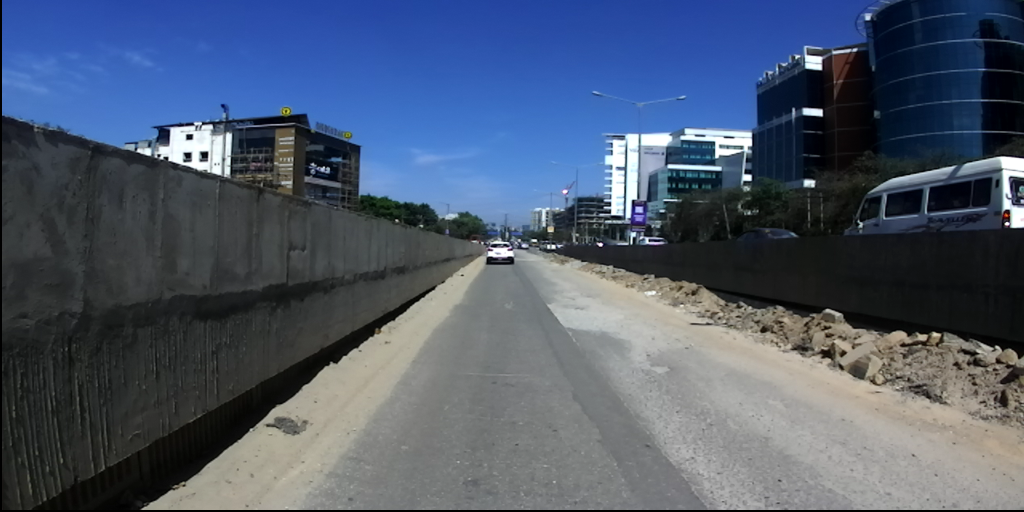}} &
  \subfloat{\includegraphics[width=0.31\linewidth]{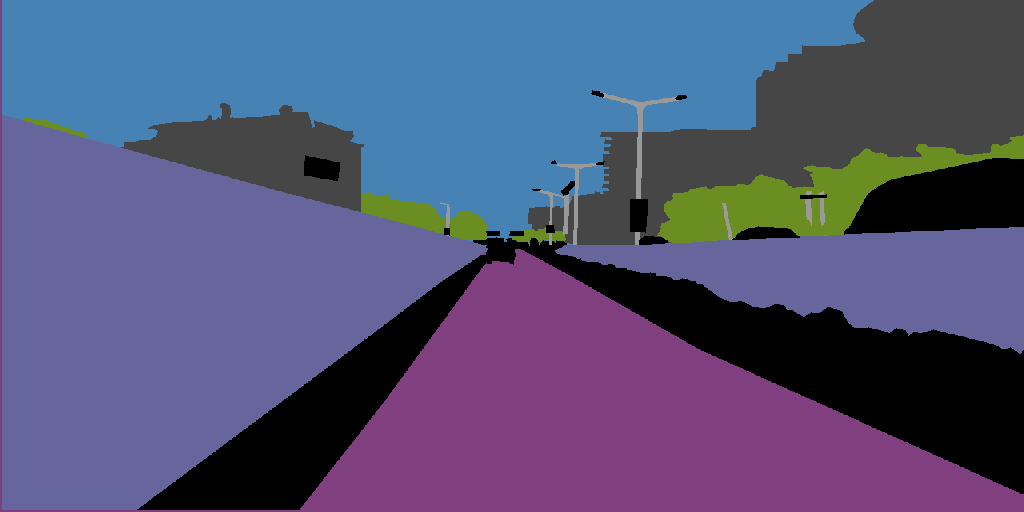}} &
  \subfloat{\includegraphics[width=0.31\linewidth]{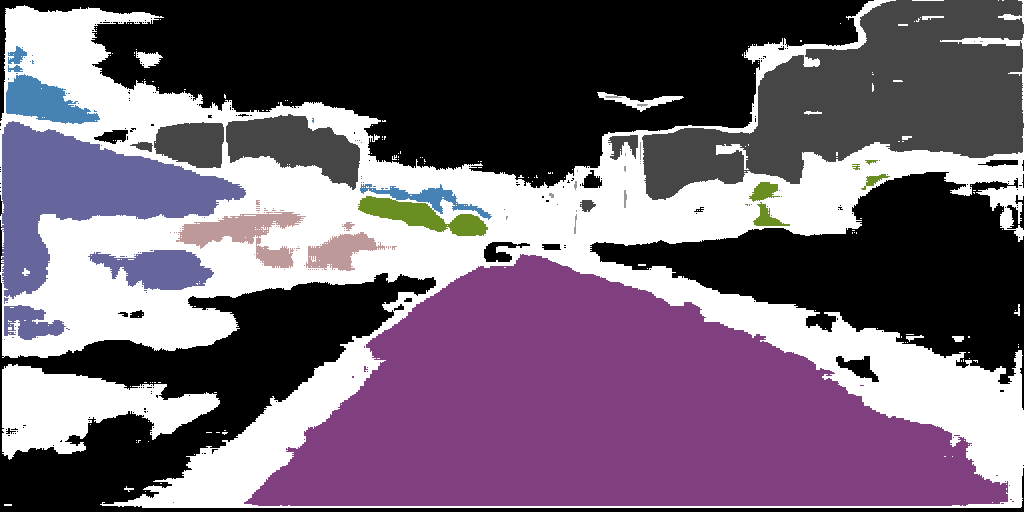}} \\[-2.8ex]
  
  \subfloat{\includegraphics[width=0.31\linewidth]{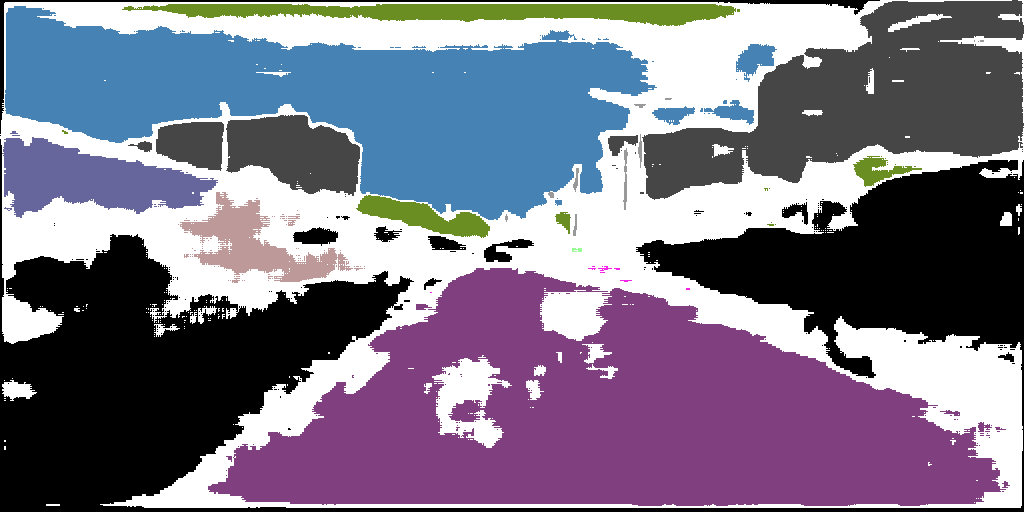}} &
  \subfloat{\includegraphics[width=0.31\linewidth]{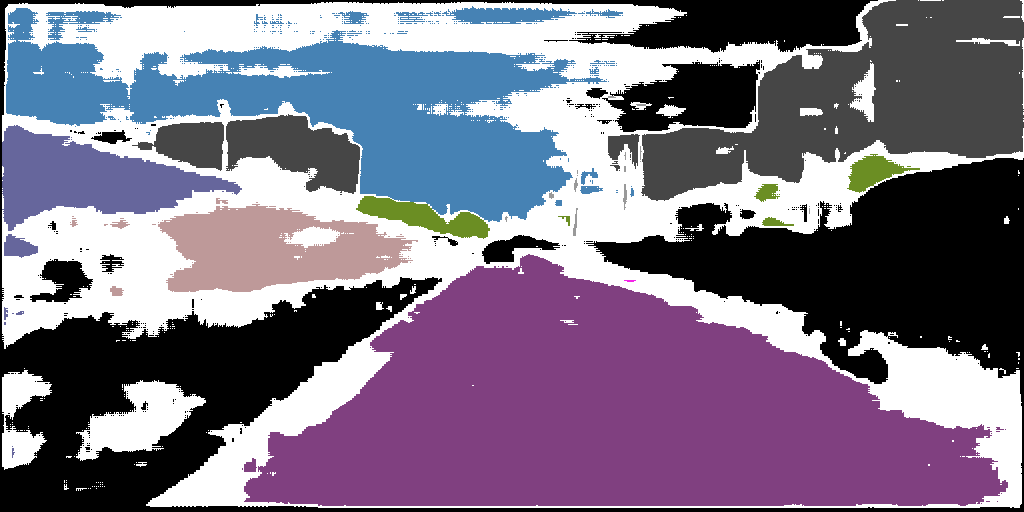}} &
  \subfloat{\includegraphics[width=0.31\linewidth]{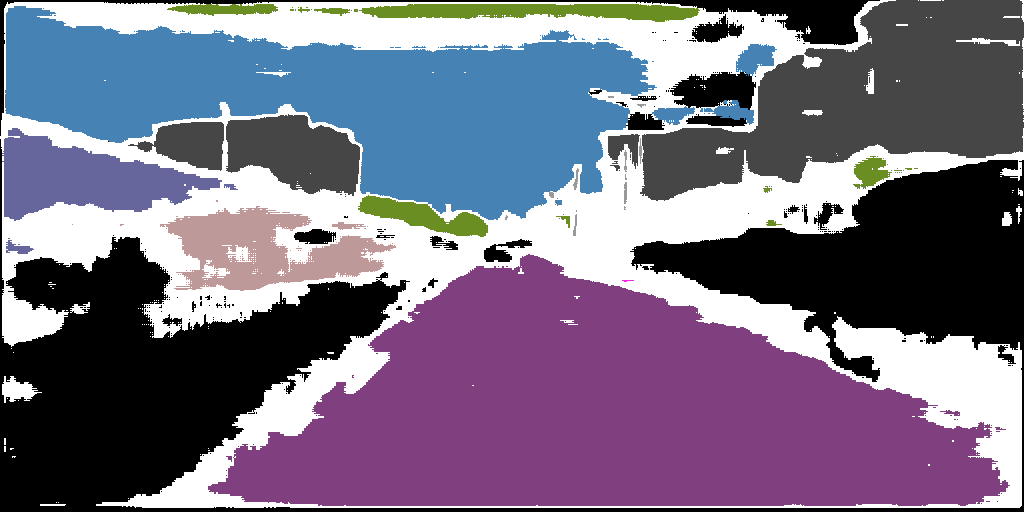}} \\[-0.5ex]
  
  $\fullhatt[-0.3ex]{\mathbf{Y}}^{\lower.75em\hbox{\fontsize{6}{6}\selectfont $\{0\}$}}_{t\shortminus1}$ & 
  $\fullhatt[-0.3ex]{\mathbf{Y}}^{\lower.75em\hbox{\fontsize{6}{6}\selectfont $\{1\}$}}_{t\shortminus1}$ & 
  $\fullhatt[-0.3ex]{\mathbf{Y}}^{\lower.75em\hbox{\smalless \fontsize{6}{6}\selectfont $t$}}_{t\shortminus1} =
  \fullhatt[-0.3ex]{\mathbf{Y}}^{\lower.75em\hbox{\fontsize{6}{6}\selectfont $\{0, \!1\}$}}_{t\shortminus1}$ 
  \\

  \end{tabular}
  \captionof{figure}{Different ways of pseudo-labeling ($t\!=\!2$). White regions correspond to the \textit{ignore} label.}
  \label{fig:pseudo}
  \end{minipage}
  \vspace{-0mm}
  
\end{table}

We \marco{further} analyze the influence exerted by pseudo-labeling \pietro{in Table \ref{tab:pseudo}}. %
We remark that the proposed enhanced labeling mechanism (described in Sec.~\ref{sec:newD_oldC}) exploits oldly-stylized images to mitigate the domain shift endured by the frozen segmentation model distilling knowledge from the past.

\pietro{
We notice that when self-stylization is disabled (first two rows) the efficacy of our method is reduced, %
while %
the beneficial effect offered by \marco{the self-stylizing module} can be appreciated in the last two rows. 
This occurs because self-stylization better prepares the segmentation model for future steps, in which the stylizing mechanism leverages old-domain styles to inject old-domain knowledge into the ongoing learning step.
\tocheck{In other words, when self-\marco{stylizing} images, what will be experienced as an \textit{old} style will have already been experienced \marco{as a \textit{new}} style before.
Therefore, the undesired visual artifacts generated by style transfer are experienced by the network from the very first step in which each domain is introduced.
This, in turn, ensures greater robustness over the incremental learning process.}
Furthermore, in setups with self-stylization, as opposed to what occurs without \marco{it}, %
pseudo-labeling performed on top of oldly-stylized images yields the best overall performance, if compared to the same labeling process executed over image samples with new-domain style.
This happens because the  network (frozen from the past step) used to \um{generate pseudo-labels} is better equipped to face input distributions of previously experienced old domains, while, instead, it may suffer from domain shift when presented with new unseen input distributions.}

In Fig.~\ref{fig:pseudo} we report pseudo-labels generated according to different criteria, to provide visual confirmation of the improved pseudo-supervision achieved \marco{on top of the \um{oldly stylization}}.
The considered setup involves $\text{CS} \rightarrow \text{BDD} \rightarrow \text{IDD}$ and $\mathcal{C}_{bgr} \rightarrow \mathcal{C}_{stat} \rightarrow \mathcal{C}_{mov}$ progressions, and maps are retrieved at the last step (\ie, $t\!=\!2$).
We observe that the segmentation model taken from
\marco{step $t \!-\! 1$ (\ie, second last step)}
is not detecting the sky region of the new-domain image, \ie, $\fullhatt[-0.3ex]{\mathbf{Y}}^{\lower.3em\hbox{\fontsize{6}{6}\selectfont $\{2\}$}}_{t\shortminus1}$ provides unreliable supervision by labeling the top portion of the picture as \marco{\textit{unknown}} %
(when the true \textit{sky} class is among those already seen).
\marco{On the other hand, when}
leveraging oldly-stylized images to generate pseudo-supervision ($\fullhatt[-0.3ex]{\mathbf{Y}}^{\lower.3em\hbox{\smalless \fontsize{6}{6}\selectfont $t$}}_{t\shortminus1}$), 
\marco{more reliable} old-domain \tocheck{guidance} %
($\fullhatt[-0.3ex]{\mathbf{Y}}^{\lower.3em\hbox{\fontsize{6}{6}\selectfont $\{0\}$}}_{t\shortminus1}$ and $\fullhatt[-0.3ex]{\mathbf{Y}}^{\lower.3em\hbox{\fontsize{6}{6}\selectfont $\{1\}$}}_{t\shortminus1}$) 
\tocheck{is exploited}, 
with individual positive contributions successfully merged in the final map (\eg, in \textit{sky} and \textit{road} regions).
\marco{
Thus, we end up with 
$\fullhatt[-0.3ex]{\mathbf{Y}}^{\lower.3em\hbox{\smalless \fontsize{6}{6}\selectfont $t$}}_{t\shortminus1}$
being more accurate than each domain-specific alternative 
$\fullhatt[-0.3ex]{\mathbf{Y}}^{\lower.3em\hbox{\fontsize{6}{6}\selectfont $\{k\}$}}_{t\shortminus1}, \, k \leq t$.}

\begin{table}[t!]
 \scriptsize
 \centering
  \setlength{\tabcolsep}{4.5pt}
  \renewcommand{\arraystretch}{1.}
 \caption{
 \marco{Ablation study on stylization ($\beta \!=\! 0.01$ corresponds to the default configuration).}
 }

  \begin{tabu}{ccccccc}
  \toprule
  \multicolumn{3}{c}{$\text{CS} \veryshortarrow \text{BDD} \veryshortarrow \text{IDD}$} &
  \multicolumn{1}{c}{No} &
  \multicolumn{3}{c}{$\beta$}
  \\
  \multicolumn{3}{c}{$\mathcal{C}_{bgr} \veryshortarrow \mathcal{C}_{stat} \veryshortarrow \mathcal{C}_{mov}$}
  &  stylization
  & $0.001$ 
  & $0.01$ 
  & $0.1$ \\
  \midrule
  \multirow{2}{*}{Step 0} & \multirow{1}{*}{mIoU\textsubscript{0} $ \! \uparrow$} & \textbf{CS}  & 79.67 & 79.8 & 79.19 & 78.54 \\
  
                         \arrayrulecolor{black!50}
                         \cmidrule{2-7}
                         \arrayrulecolor{black!100}
                         
                          & \multicolumn{2}{c}{$\bar{\Delta}_{0}$  $ \! \downarrow$}               &  \oracle{6.44}{5.32} & \oracle{6.29}{\tblbold{5.17}} & \oracle{7.0}{5.89} & \oracle{7.76}{6.66} \\
 \midrule                          
 \multirow{3}{*}{Step 1} & \multirow{2}{*}{mIoU\textsubscript{1}  $ \! \uparrow$} & \textbf{BDD}   & 33.67 & 35.06 & 44.47 & 44.79 \\
                         & & CS  & 49.20 & 43.75 & 53.31 & 50.45 \\
                         
                         \arrayrulecolor{black!50}
                         \cmidrule{2-7}
                         \arrayrulecolor{black!100}
                         
                         & \multicolumn{2}{c}{$\bar{\Delta}_{1}$ $ \! \downarrow$}      & \oracle{39.71}{38.08} & \oracle{42.35}{40.88} & \tblbold{\oracle{28.37}{26.58}} & \oracle{30.08}{28.37} \\
 \midrule 
 \multirow{4}{*}{Step 2} & \multirow{3}{*}{mIoU\textsubscript{2}  $ \! \uparrow$} & \textbf{IDD}  & 43.33 & 48.60 & 51.20 & 50.03 \\
                         & & BDD  & 26.62 & 27.77 & 35.73 & 34.84 \\
                         & & CS  & 37.36 & 37.61 & 44.17 & 43.01 \\
                         
                         \arrayrulecolor{black!50}
                         \cmidrule{2-7}
                         \arrayrulecolor{black!100}
                         
                         & \multicolumn{2}{c}{$\bar{\Delta}_{2}$  $ \! \downarrow$}      & \oracle{42.58}{43.96} & \oracle{39.11}{40.59} & \tblbold{\oracle{29.27}{31.28}} & \oracle{31.01}{32.97} \\

  \bottomrule
 \end{tabu}
 \label{tab:style}
\end{table}

\subsection{Degree of Stylization}
\label{sec:abl_style}

We propose an additional analysis on the stylization mechanism.
Table \ref{tab:style} shows the results of the \marco{our} method 
\marco{(complete with all objectives)}
under different degrees of stylization, which are \marco{determined} by the $\beta$ parameter (\marco{see} Sec.~\ref{sec:domain_style}). 
We notice that disabling stylization or \marco{operating} it in a more conservative manner (\marco{\ie, with} $\beta \!=\! 0.001$) yields low results, with the latter configuration still outperforming the no stylization approach, \marco{as} the statistical properties captured and transferred are not sufficient to successfully retain old-domain information.
On the other hand, if the stylization is \marco{raised} to an excessive extent (\marco{\ie, with} $\beta \!=\! 0.1$), we observe performance degradation on the overall $\bar\Delta_{2}$ score. 
In this scenario, artifacts are more likely to be introduced on oldly-stylized images, thus hindering the segmentation task.

\subsection{Knowledge Transfer Across Tasks and Domains}
\label{sec:abl_transfer}

\begin{figure}[!t]
  \centering
  \includegraphics[width=0.9\linewidth,trim={2.cm 1.7cm 1.8cm 1.2cm},clip]{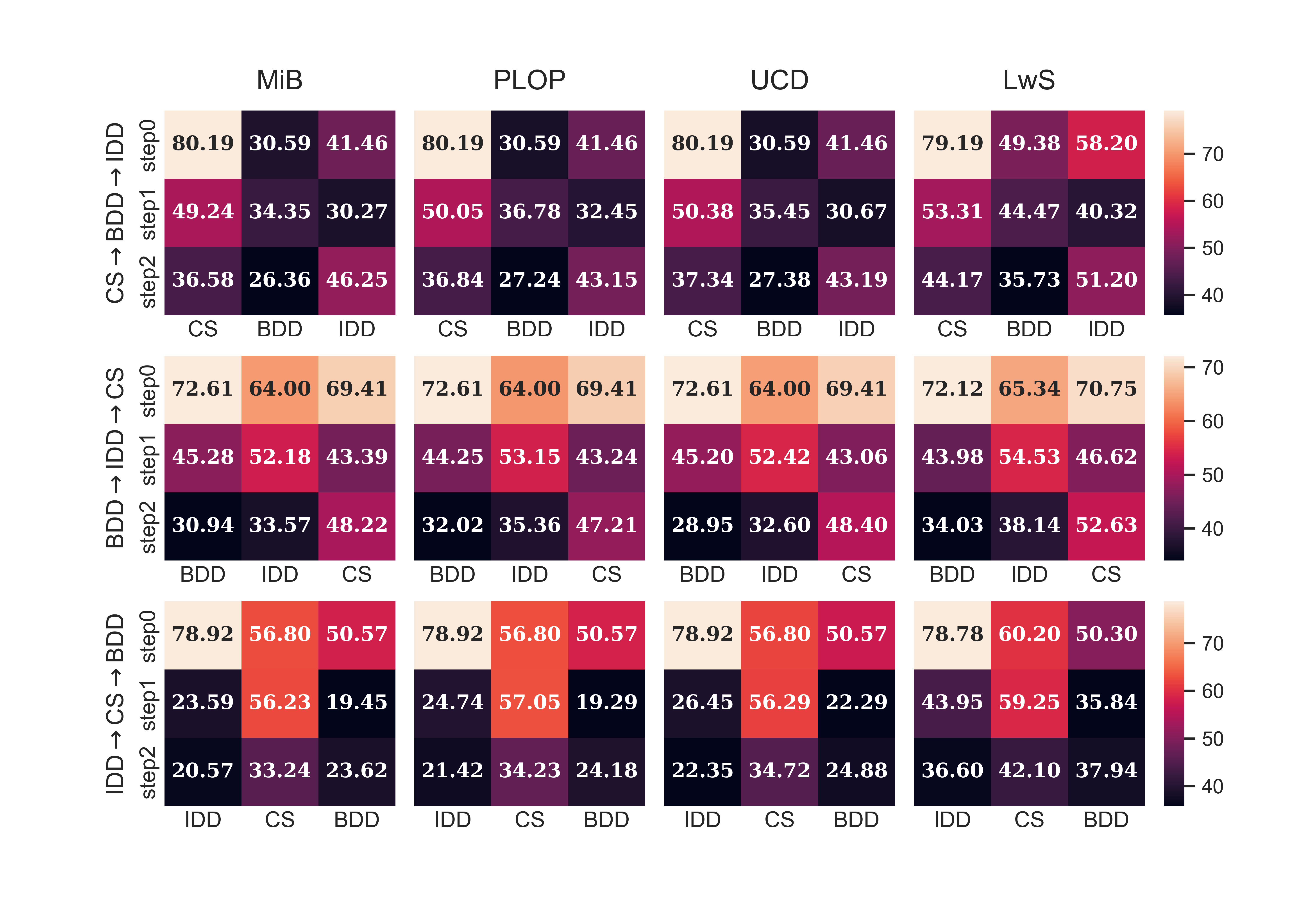}
  \captionof{figure}{Domain-knowledge transfer (mIoU $\! \uparrow$ (\%)).}
  \label{fig:domain_tf}
\end{figure}

We \marco{propose} further ablation studies to evaluate the knowledge transfer aptitude of \marco{our} method, both \marco{under} task and domain perspectives.
Fig.~\ref{fig:domain_tf} \marco{presents} a comparative of multiple CIL competitors in terms of predisposition towards \textit{domain-knowledge transfer;}
we report the mIoU achieved on individual domains \pietro{only on classes experienced so far} across multiple steps in matrix form. %
\marco{We consider multiple incremental setups, with urban datasets and variable domain order.}
We observe that our approach, right from the first learning step, achieves better \textit{forward transfer} to future domains, 
\marco{as indicated by per-domain mIoU values in the top triangular sections}, 
regardless of the setup considered.
\marco{At the same time, this translates into superior performance on current domains (represented by diagonal mIoU values), as they benefit from a better forward-adaptability acquired before.}
Plus, improved \textit{backward transfer} to \marco{former} domains is testified by higher \marco{mIoU} values in the bottom triangular part of matrices.

\begin{figure}[!t]
  \centering
  \includegraphics[width=0.9\linewidth,trim={2cm 1.7cm 1.8cm 1.2cm},clip]{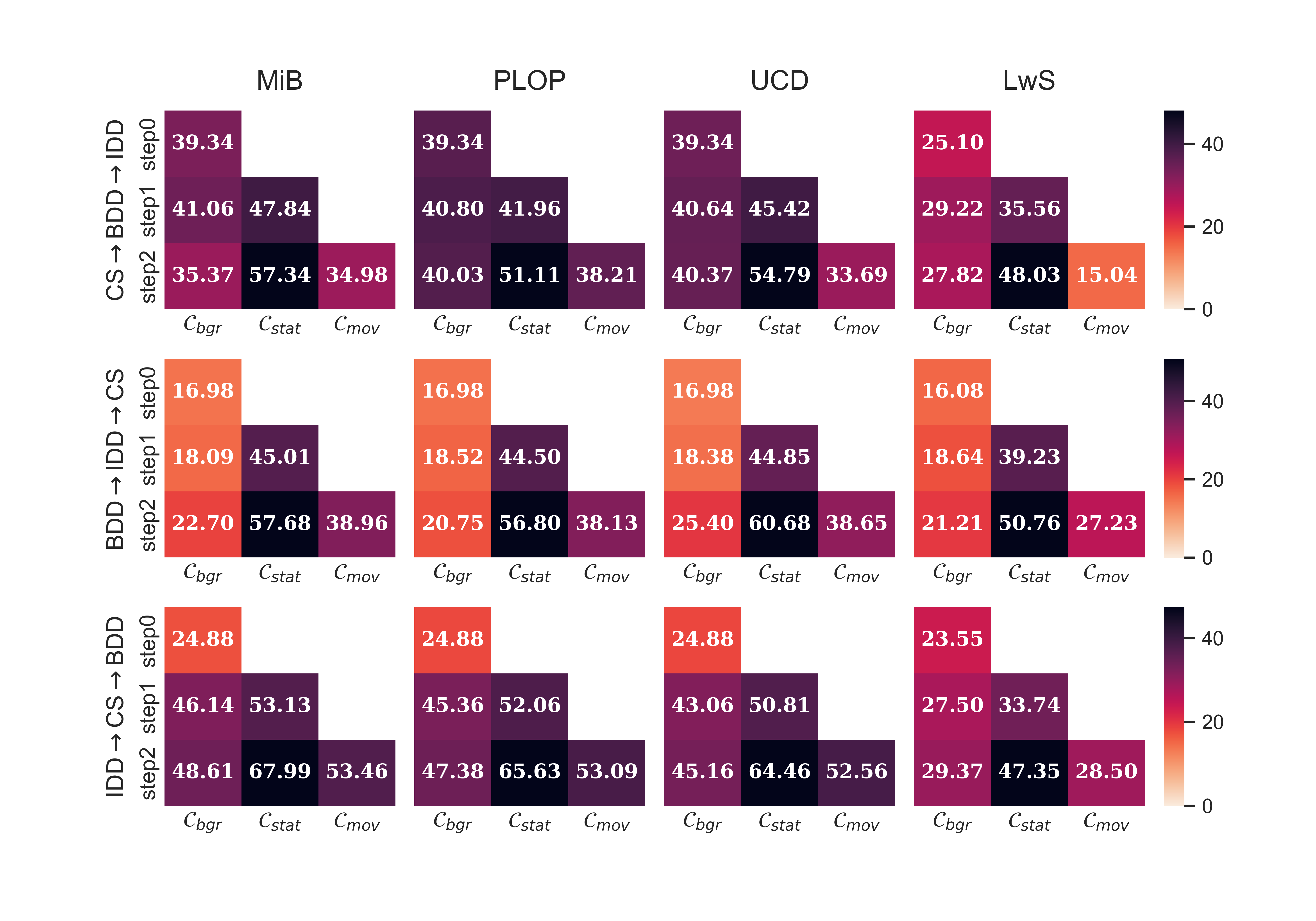}
  \captionof{figure}{Task-knowledge transfer ($\bar\Delta \! \downarrow$).}
  \label{fig:task_tf}
\end{figure}

\marco{To provide an insight on \textit{task-knowledge transfer} proneness of different incremental methods, in Fig.~\ref{fig:task_tf} we report a comparative in terms of $\bar\Delta$ results at multiple learning steps; values are computed on \textit{single} incremental sets of classes and represent an average score across all domains (both experienced and future ones).%
}
The experimental setups are the same considered \marco{when studying} domain transfer\marco{, and results are arranged in matrix form}.
We observe that our $\bar\Delta$ scores in the bottom triangular part of matrices are lower than competitors, \marco{suggesting that our method yields} better \textit{backward transfer} in terms of task knowledge.
\marco{At the same time, the smaller $\bar\Delta$ diagonal elements indicate improved performance on current tasks, confirming the better stability-plasticity compromise offered by our approach.}

\section{Conclusions}
In this paper, we formalized a general setting for continual learning, where both domains and tasks to be learned incrementally change over time.
We addressed this under-explored learning setting 
targeting the semantic segmentation task 
by breaking it down into underlying sub-problems, each tackled with a specific learning objective.
Leveraging a stylization mechanism, domain knowledge is replayed over time, whereas a robust distillation mechanism allows to retain and adapt old-task information.
Overall, the proposed learning framework enables learning new tasks, while preserving \pietro{performance on old ones and spreading task knowledge across all the encountered domains}.
We achieved significant results outperforming state-of-the-art competitors on multiple challenging benchmarks. %
Further research will  tackle even more application-oriented settings, \ie, where task and domain shifts happen in a \pietro{continuous} 
fashion rather than in discrete steps and distinct overlapping sets of classes are \marco{introduced} in different domains.

\bibliographystyle{IEEEtran}
\bibliography{strings_reduced,refs}

\begin{IEEEbiography}[{\includegraphics[width=0.9\textwidth, trim=2cm 4cm 1.8cm 0, clip]{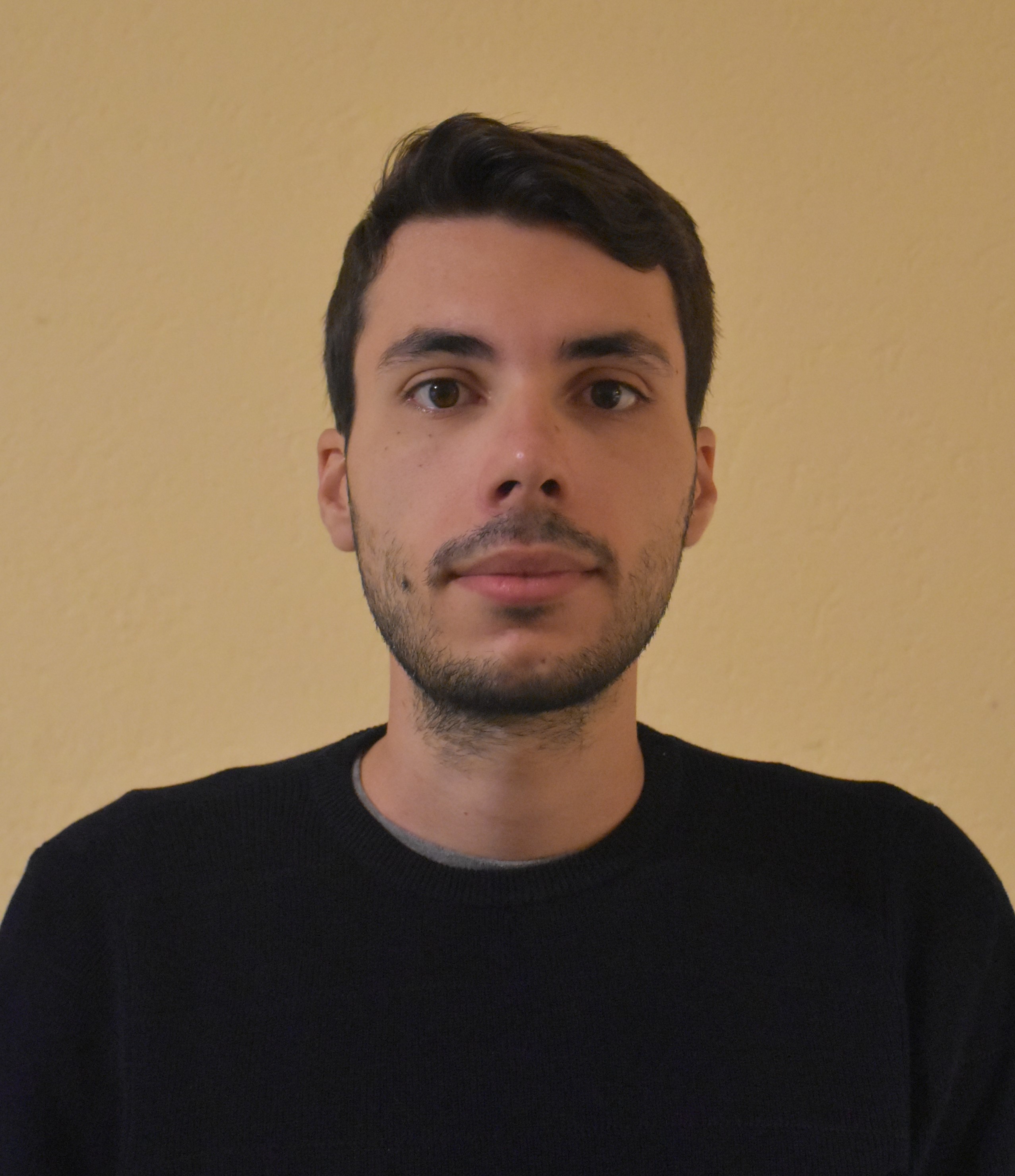}}]{Marco Toldo}
received the M.Sc.\ degree in ICT for Internet and Multimedia in 2019 at the University of Padova. At present, he is doing his Ph.D. at the Department of Information Engineering of the same university. In 2021, he did an internship as Research Engineer at Samsung Research UK. His research involves domain adaptation and continual learning applied to computer vision.
\end{IEEEbiography}
\vspace{-1cm}
\begin{IEEEbiography}[{\includegraphics[width=1in,height=1in,clip,keepaspectratio]{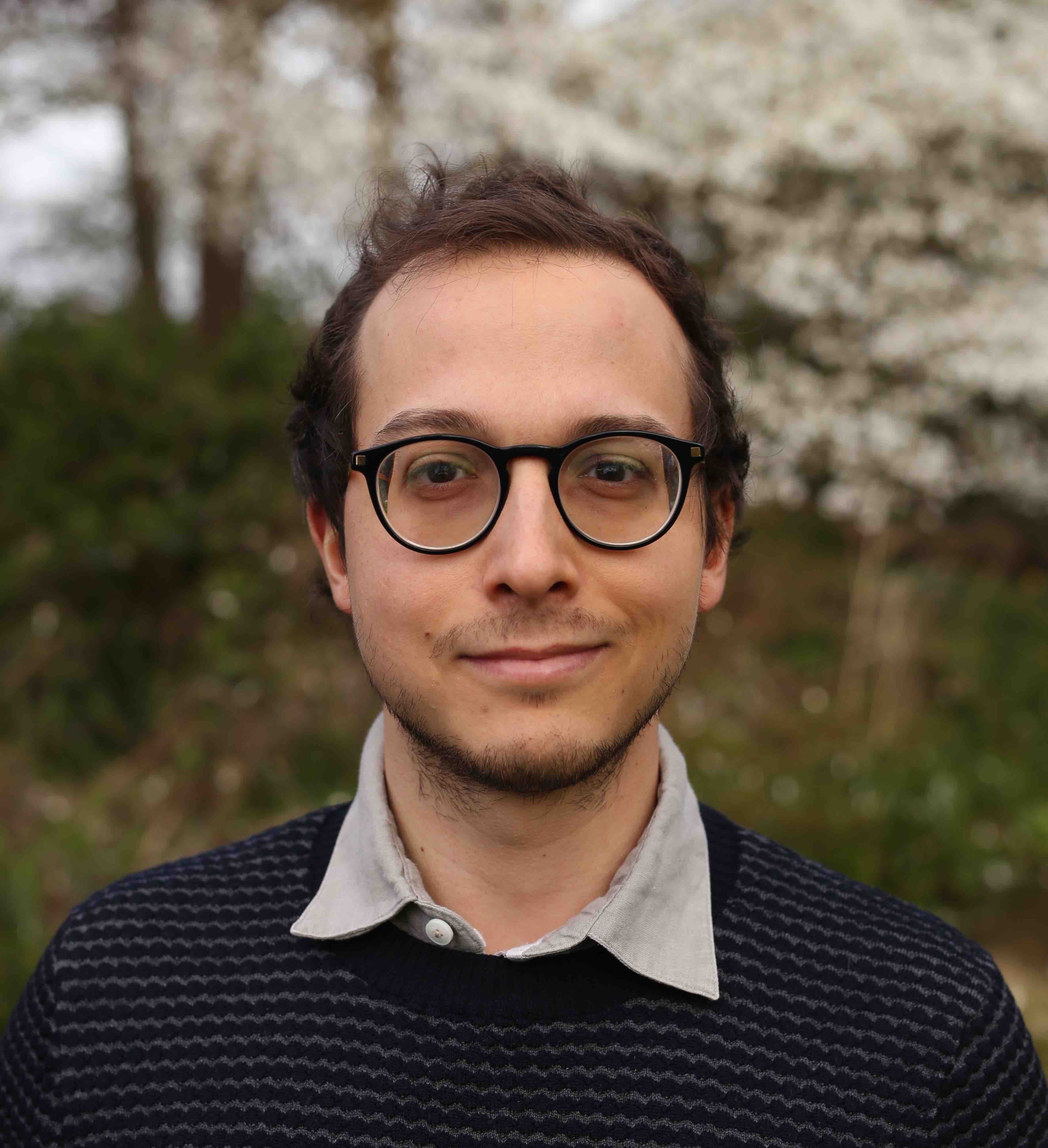}}]{Umberto Michieli}
received his Ph.D.\ in Information Engineering from the University of Padova in 2021. Currently, he is a Postoctoral Researcher and Adjunct Professor at the same University. He spent research periods at Technische Universit\"at Dresden and Samsung Research UK. His research lies at the intersection of foundation AI problems applied to semantic understanding. In particular, he focuses on domain adaptation, continual learning, coarse-to-fine learning and federated learning.
\end{IEEEbiography}
\vspace{-1cm}
\begin{IEEEbiography}[{\includegraphics[width=1in,height=1in,clip,keepaspectratio]{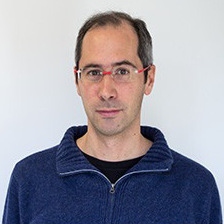}}]{Pietro Zanuttigh}  received the  the Ph.D. degree from the University of Padova, Italy 
in  2007. He is currently an Associate Professor at the Department of Information Engineering. His research interests are image and 3D data processing and analysis, with a special focus on domain adaptation and incremental learning in semantic segmentation, ToF sensors data processing and hand gesture recognition. 
\end{IEEEbiography}

\vfill

\end{document}